\documentclass[11pt]{article}
\pdfoutput=1
\usepackage{graphicx}
\usepackage{booktabs}
\usepackage{amsmath}
\usepackage{amssymb}
\usepackage{amsthm,url}
\usepackage{multicol,multirow}
\usepackage{algorithm,algorithmic}
\usepackage{bbm}
\usepackage{natbib}

\theoremstyle{plain}
\newtheorem{theorem}{Theorem}
\newtheorem{corollary}[theorem]{Corollary}
\allowdisplaybreaks[4]

\newcommand{\diff}{\mathrm{d}}

\newcommand{\argmax}{\mathop{\rm arg~max}\limits}

\newcommand{\figimage}[4]{\begin{figure}[t]\centering\includegraphics[width=#4\textwidth]{#1}\caption{#2}\label{#3}\end{figure}}
\newcommand{\figimagep}[4]{\begin{figure}[p]\centering\includegraphics[width=#4\textwidth]{#1}\caption{#2}\label{#3}\end{figure}}
\newcommand{\figimagetwo}[6]{\begin{figure}[t]\centering\includegraphics[width=#5\textwidth]{#1}\qquad\includegraphics[width=#6\textwidth]{#2}\caption{#3}\label{#4}\end{figure}}
\newcommand{\figimagethree}[8]{\begin{figure}[tbp]\centering\includegraphics[width=#6\textwidth]{#1}\includegraphics[width=#7\textwidth]{#2}\includegraphics[width=#8\textwidth]{#3}\caption{#4}\label{#5}\end{figure}}
\newcommand{\figimages}[8]{\begin{figure}[tbp]\begin{tabular}{cc}\begin{minipage}{0.5 \textwidth}\begin{center}\centering\includegraphics[width=#4 \textwidth]{#1}\caption{#2}\label{#3}\end{center}\end{minipage}\begin{minipage}{0.5 \textwidth}\centering\includegraphics[width=#8 \textwidth]{#5}\caption{#6}\label{#7}\end{minipage}\end{tabular}\end{figure}}

\begin{document}

\title{Ensemble Kalman Variational Objective: \\Nonlinear Latent Time-Series Model Inference by a Hybrid of Variational Inference and Ensemble Kalman Filter}
\author{Tsuyoshi Ishizone$^1$, Tomoyuki Higuchi$^2$, and Kazuyuki Nakamura$^1$}
\date{1: Meiji University, 2: Chuo University}

%

\maketitle

\begin{abstract}
Variational inference (VI) combined with Bayesian nonlinear filtering produces state-of-the-art results for latent time-series modeling.
A body of recent work has focused on sequential Monte Carlo (SMC) and its variants, e.g., forward filtering backward simulation (FFBSi).
Although these studies have succeeded, serious problems remain in particle degeneracy and biased gradient estimators.
In this paper, we propose {\it Ensemble Kalman Variational Objective} (EnKO), a hybrid method of VI and the ensemble Kalman filter (EnKF), to infer state space models (SSMs).
Our proposed method can efficiently identify latent dynamics because of its particle diversity and unbiased gradient estimators.
We demonstrate that our EnKO outperforms SMC-based methods in terms of predictive ability and particle efficiency for three benchmark nonlinear system identification tasks.
\end{abstract}


\section{Introduction}
\label{sec:intro}

Model inference of time series is essential for prediction, control, and reanalysis.
For example, epidemiologists use models to predict the spread of infectious diseases, economists use models to make transactions, and meteorologists use models to reanalyze weather fields.
If the model is already established, there is no need for inference, but in many cases it is not, and data-driven model inference is necessary.

Time-series model inference has been studied for many years in the statistics and control communities.
In the statistics community, ARIMA models, state space models \citep{shu17}, and Gaussian process dynamical models \citep{wan08} have been proposed.
In the control community, time series model inference has been studied as a form of system identification; proper orthogonal decomposition \citep{hol12} and dynamic mode decomposition \citep{kut16} are aimed at identifying low-dimensional linear dynamics from high-dimensional data.
However, it is difficult to reconstruct complex models in these statistical and control system methods due to the lack of model representation capability.

In order to improve the representational capability of models, time-series model inference techniques using deep learning have been proposed \citep{chu15,fra16,bay15,goy17,mor19s}.
These techniques, called Sequential Variational AutoEncoders (SVAEs), use Variational AutoEncoder (VAE) \citep{kin14v} to capture intrinsic latent states from observed data, and recurrent networks such as GRU \citep{cho14} and LSTM \citep{hoc97} to capture latent dynamic structures.
VAE is a kind of stochastic generative model that aims to obtain a process $p_\theta(x|z)$ where the observed data $x$ is generated from an unknown latent variable $z$ (Figure \ref{fig:svae}).
For practical use, it simultaneously learns the parameters of encoder $q_\varphi(z|x)$ which infers $z$ and decoder $p_\theta(x|z)$ which corresponds to the generative process.
SVAEs can be viewed as an extension of VAEs to dynamic data, where the encoder $q_\varphi(z_t|x_{1:T})$, decoder $p(x_t|z_t)$, and latent transition process $p_\theta(z_t|z_{1:t-1})$ are learned to obtain the generative process
\begin{equation}
p_\theta(x_{1:T},z_{1:T})=p_\theta(z_1)\prod_{t=2}^Tp_\theta(z_t|z_{1:t-1})\prod_{t=1}^Tp_\theta(x_t|z_t).
\end{equation}
Because of their high model representation capability, these techniques have been applied to a wide range of data, such as music data and mouse brain voltage data.

Technically, VAE and SVAEs use variational inference (VI) \citep{bea03,jor99} to learn the parameters.
VI is a technique that avoids calculating the intractable maximum of the marginal likelihood $p(x_{1:T})$ by maximizing the evidence lower bound, defined as the lower bound of the log marginal likelihood.
In order to achieve a tighter bound, the objective function is given by an ensemble of particles in IWAE \citep{bur15,dom18} and FIVO \citep{mad17}.
While IWAE gives the objective function by simple Monte Carlo integration, FIVO uses sequential Monte Carlo (SMC) to asymptotically give a tighter bound than IWAE.

However, FIVO has two drawbacks: particle degeneracy and biased gradient estimators.
Particle degeneracy means only one or a few particles occupy a large weight and are copied at the resampling step of SMC, resulting in low particle diversity \citep{duc11}.
Particle diversity enhances the ability to represent the probabilistic density function (PDF), and the low diversity results in poor model inferences.
Biased gradient estimators are used in the method due to the loss of the resampling gradient because the resampling gradient results in estimators with high variance and unstable learning.
These problems are inherent to the SMC.
In contrast, the ensemble Kalman filter (EnKF) \citep{eve94,eve03}, which is a nonlinear Bayesian filtering method used mainly in geophysics \citep{sak12,fox18}, does not suffer from these problems.

Thus, we propose a new method called {\it Ensemble Kalman Variational Objective} (EnKO), which combines SVAEs with the EnKF to infer the latent time-series model and obtain rich PDF representations.
The main idea of the proposed method is to take advantage of the characteristics of the EnKF.

\figimage{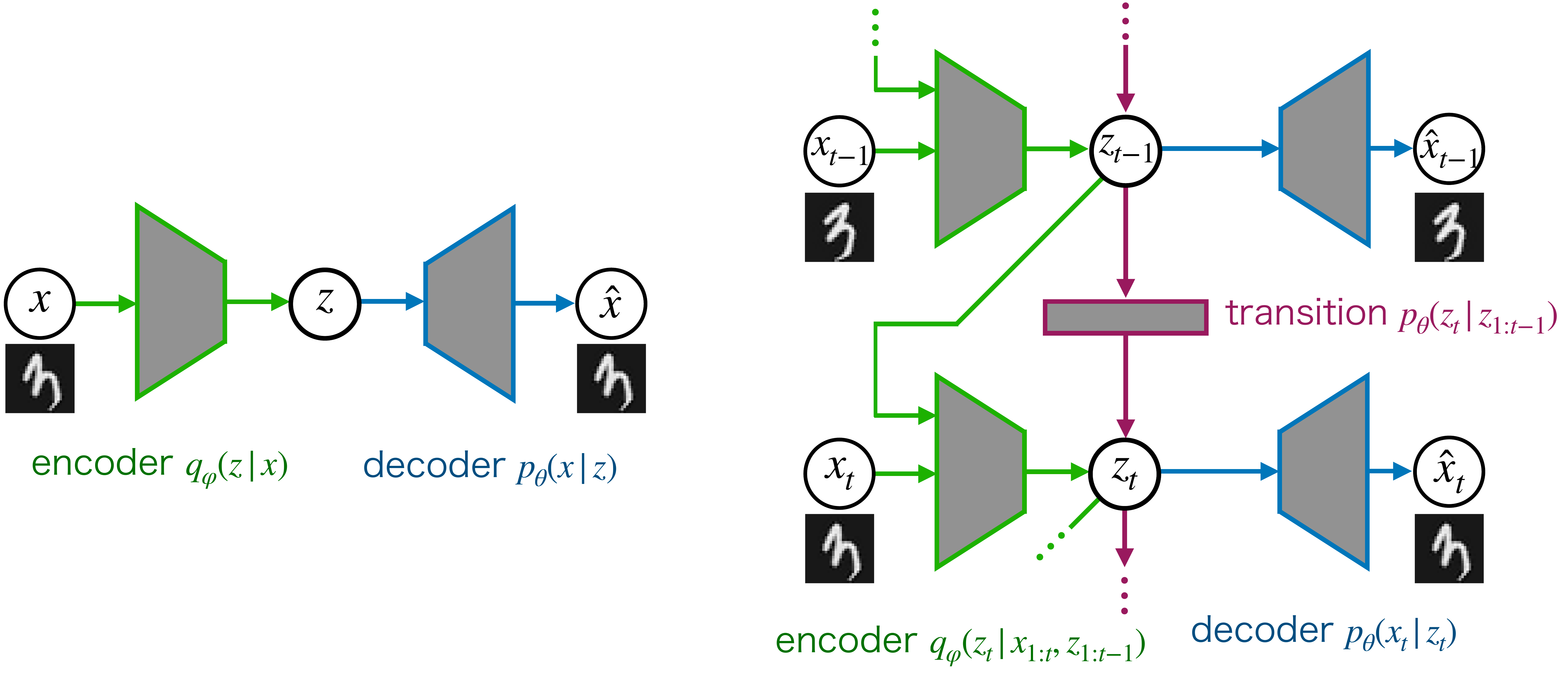}{Schematic image of VAE (left) and SVAE (right). Rectangles and circles represent neural networks and random variables, respectively.}{fig:svae}{1}

\subsection{Contributions}
In this paper, our main contributions are follows:
\begin{enumerate}
\item We first provide a new framework called EnKO, which uses the EnKF in the inference phase.
The proposed method has two main advantages over the FIVO framework: particle diversity and unbiased gradient estimators.
The diversity provides rich information of the latent states to predict observations and quantify their uncertainty.
The gradient estimators are unbiased because of no need to truncate all terms while the FIVO truncates the resampling term.
\item It is known that the EnKF often underestimates the sample covariance matrix of the latent states.
To alleviate the underestimating problem, we introduce a covariance inflation technique generally used in geophysics.
\item The computational complexity of EnKO is $O(d_x^2)$, which is less scalable than the computational complexity of FIVO $O(d_x)$, where $d_x$ represents the observation dimensions.
We provide a technique to reduce the computational complexity to $O(d_x)$ by incorporating the auxiliary variables to alleviate this problem.
\item We show that EnKO has been applied to multiple synthetic and real-world datasets to achieve experimentally superior prediction accuracy.
Furthermore, we report that the variance of the gradient estimator of EnKO is lower than that of FIVO and IWAE for toy examples.
\end{enumerate}

\subsection{Organization of the Paper}
In Section 2, we introduce previous studies.
Section 3 introduces VI, SVAEs, EnKF, and covariance inflation to prepare for the proposed method.
Section 4 introduces the proposed method, EnKO, and discusses its concrete algorithm and simple theoretical validity.
In Section 5, we conduct experiments on several synthetic and real-world datasets, compare the results with existing methods, and explain the results of ablation studies.
Finally, Section 6 summarizes this paper, and detailed theoretical and experimental results are given in the Appendix.

\section{Related Work}
\label{sec:work}

\paragraph{Variational Inference}
Variational inference (VI) is a method to learn parameters by maximizing the evidence lower bound (ELBO) of the log marginal likelihood; the maximization is equivalent to the minimization of Kullback-Leibler divergence between the true posterior and the variational posterior \citep{jor99,bea03}.
Regarding applications to neural networks, VAE \citep{kin14v} is important; this incorporates probabilistic models into an auto-encoder.
Some advanced VI methods have been proposed to achieve tighter bounds of the log marginal likelihood \citep{zha17}.
IWAE \citep{bur15,dom18} achieves tighter bounds by describing the ELBO by averages over multiple particles.
TVO \citep{mas19} bridges thermodynamic integration and VI.
SIVI \citep{yin18}, DSIVI \citep{mol19}, and UIVI \citep{mic19} expand the applicability of VI by defining expressive variational families.

\paragraph{Sequential VAEs}
Sequential VAE methods (SVAEs) expand VAE \citep{kin14v} to probabilistic time-series models and infer the model by VI.
DKF \citep{kri16}, DMM \citep{kri16s}, KVAE \citep{fra17}, E2C \citep{wat15}, and DynaNet \citep{che19d} directly construct state space model (SSM), a latent time-series model which satisfy Markov property, and infer the latent probabilistic models.
VRNN \citep{chu15} and STORN \citep{bay15} construct the transition model with RNNs such as GRU \citep{cho14} and LSTM \citep{hoc97} to capture long-term dependencies.
SRNN \citep{fra16}, Z-forcing \citep{goy17}, and SVO \citep{mor19s} introduce backward recursion to the forward model to capture future information. 

To obtain tighter bounds, FIVO \citep{mad17}, AESMC \citep{le18}, and VSMC \citep{nae18} proposed VI methods combined with sequential Monte Carlo (SMC).
These methods are ensemble systems applied to SVAEs to infer more accurate models.
Hitherto, a framework called MCO was generalized, including ELBO, IWAE, and FIVO, and it was revealed that the objectives achieved tighter bounds of the log marginal likelihood \citep{mad17}.
TVSMC \citep{law18} and SMC-Twist \citep{lin18} obtain tighter bounds by introducing twisted functions for capturing future information; however, experimental results revealed that these methods have lower log marginal likelihood compared to non-twisted methods.
These SMC-based methods obtain tighter bounds for sequential data; however, they have two key drawbacks: particle degeneracy and biased gradient estimators.

PSVO \citep{mor19p,mor19p2} proposed an inference framework using forward filtering backward simulation (FFBSi) \citep{god04,duc11}, a method to estimator the posterior conditioned on all observations in the SSM formulation, to achieve unbiased estimators and a tighter bound than SMC-based methods.
The method, however, has three key disadvantages: particle degeneracy, high calculation and memory cost, and high sensitivity of the hyper-parameters.
The first disadvantage also applies to SMC-based methods; the second one arises from long backward loss calculation and storage of all particles at the time forward calculation for the time backward calculation.
The last one means PSVO needs careful tuning of the hyper-parameters due to training instability.
In contrast, the proposed method does not suffer from the disadvantages of the SMC-based methods and the PSVO method.

\paragraph{Ensemble Kalman Filter}
The ensemble Kalman filter (EnKF) represents the PDF with an ensemble of particles and corrects the particles for explaining observations well at each time-step \citep{eve94,eve03}.
Combined with deep neural networks, Bayesian LSTM \citep{che18b} represents weights of LSTM by an ensemble of particles and propagates the particles using EnKF at each mini-batch which is considered as time-step.
A similar framework \citep{loh18} has been applied to time production prediction in natural gas wells, and robust estimators were obtained.
Bayesian neural networks with EnKF \citep{che19b} approximate the distribution of weights by particles and update the particles using EnKF at each mini-batch.
Although these methods use EnKF at the batch domain of neural network parameters, to the best of our knowledge, the proposed method is the first attempt to infer a latent time-series model using a hybrid of VI and the EnKF.

\section{Preliminaries}
\label{sec:pre}

\paragraph{Probabilistic Time-Series Models}
Let $\boldsymbol{x}_t\in\mathbb{R}^{d_x}$ denote the observation vector at a given time $t$, where $d_x$ represents the observation dimension.
Probabilistic time-series model (PTSM) describes the generating process of $X=\boldsymbol{x}_{1:T}=\{\boldsymbol{x}_1,\cdots,\boldsymbol{x}_T\}$ through $d_z$-dimensional unobserved latent sequence $Z=\boldsymbol{z}_{1:T}$.
Using these two variables, the models are represented by
\begin{subequations}
\begin{align}
\boldsymbol{z}_t&\sim f_{\boldsymbol{\theta}}(\boldsymbol{z}_t|\boldsymbol{z}_{1:t-1}),\\
\boldsymbol{x}_t&\sim g_{\boldsymbol{\theta}}(\boldsymbol{x}_t|\boldsymbol{z}_t).
\end{align}
\end{subequations}
The joint distribution is then factorized as
\begin{align}
p_{\boldsymbol{\theta}}(X,Z)=f_{\boldsymbol{\theta}}(\boldsymbol{z}_1)\prod_{t=2}^Tf_{\boldsymbol{\theta}}(\boldsymbol{z}_t|\boldsymbol{z}_{1:t-1})\prod_{t=1}^Tg_{\boldsymbol{\theta}}(\boldsymbol{x}_t|\boldsymbol{z}_t).
\end{align}

\paragraph{Variational Inference}
The maximization of the log marginal likelihood is the intractable optimization problem
\begin{equation}
\argmax_{\boldsymbol{\theta}}\log p_{\boldsymbol{\theta}}(X)=\argmax_{\boldsymbol{\theta}}\log\int p_{\boldsymbol{\theta}}(X,Z)\ \diff Z.
\end{equation}
Variational inference (VI) is an alternating method to maximize the evidence lower bound (ELBO) $\mathcal{L}_{\mathrm{ELBO}}$:
\begin{equation}
\log p_{\boldsymbol{\theta}}(X)\ge\mathbb{E}_{q_{\boldsymbol{\varphi}}(Z|X)}\left[\log\frac{p_{\boldsymbol{\theta}}(X,Z)}{q_{\boldsymbol{\varphi}}(Z|X)}\right]=:\mathcal{L}_{\mathrm{ELBO}}(\boldsymbol{\theta},\boldsymbol{\varphi},X),
\end{equation}
where $q_{\boldsymbol{\varphi}}(Z|X)$ is the tractable variational posterior distribution.
The maximization is equivalent to the minimization of Kullback-Leibler divergence between the true posterior $p_{\boldsymbol{\theta}}(Z|X)$ and the variational posterior $q_{\boldsymbol{\varphi}}(Z|X)$.

IWAE \citep{bur15,dom18} provides a tighter bound by an ensemble of particles:
\begin{equation}
\mathcal{L}^N_{\mathrm{IWAE}}(\boldsymbol{\theta},\boldsymbol{\varphi},X):=\mathbb{E}_{\prod_iq_{\boldsymbol{\varphi}}(Z^{(i)}|X)}\left[\log\left(\frac{1}{N}\sum_{i=1}^N\frac{p_{\boldsymbol{\theta}}(X,Z^{(i)})}{q_{\boldsymbol{\varphi}}(Z^{(i)}|X)}\right)\right],\label{eq:iwae}
\end{equation}
where $N$ denotes the number of particles and $\mathcal{L}_{\mathrm{IWAE}}^1=\mathcal{L}_{\mathrm{ELBO}}$.

\paragraph{Sequential VAEs}
\begin{table}[t]
\caption{Examples of SVAEs}
\label{tab:svae}
\vskip 0.05in
\begin{center}
\begin{small}
\begin{tabular}{lccccr}
\toprule
\multirow{2}{*}{Network} & VRNN & SRNN & SVO \\
& \citep{chu15} & \citep{fra16} & \citep{mor19s}\\
\midrule
encoder & $q_{\boldsymbol{\varphi}}(\boldsymbol{z}_t|\boldsymbol{z}_{1:t-1},\boldsymbol{x}_{1:t})$ & $q_{\boldsymbol{\varphi}}(\boldsymbol{z}_t|\boldsymbol{z}_{t-1},\boldsymbol{x}_{1:T})$ & $q_{\boldsymbol{\varphi}}(\boldsymbol{z}_t|\boldsymbol{z}_{t-1},\boldsymbol{x}_{1:T})$ \\
decoder & $g_{\boldsymbol{\theta}}(\boldsymbol{x}_t|\boldsymbol{z}_{1:t},\boldsymbol{x}_{1:t-1})$ & $g_{\boldsymbol{\theta}}(\boldsymbol{x}_t|\boldsymbol{z}_{t},\boldsymbol{x}_{1:t-1})$ & $g_{\boldsymbol{\theta}}(\boldsymbol{x}_t|\boldsymbol{z}_{t})$ \\
transition & $f_{\boldsymbol{\theta}}(\boldsymbol{z}_t|\boldsymbol{z}_{1:t-1},\boldsymbol{x}_{1:t-1})$ & $f_{\boldsymbol{\theta}}(\boldsymbol{z}_t|\boldsymbol{z}_{t-1},\boldsymbol{x}_{1:t-1})$ & $f_{\boldsymbol{\theta}}(\boldsymbol{z}_t|\boldsymbol{z}_{t-1})$ \\
graph & \begin{minipage}{0.25\textwidth}\centering\scalebox{0.2}{\includegraphics{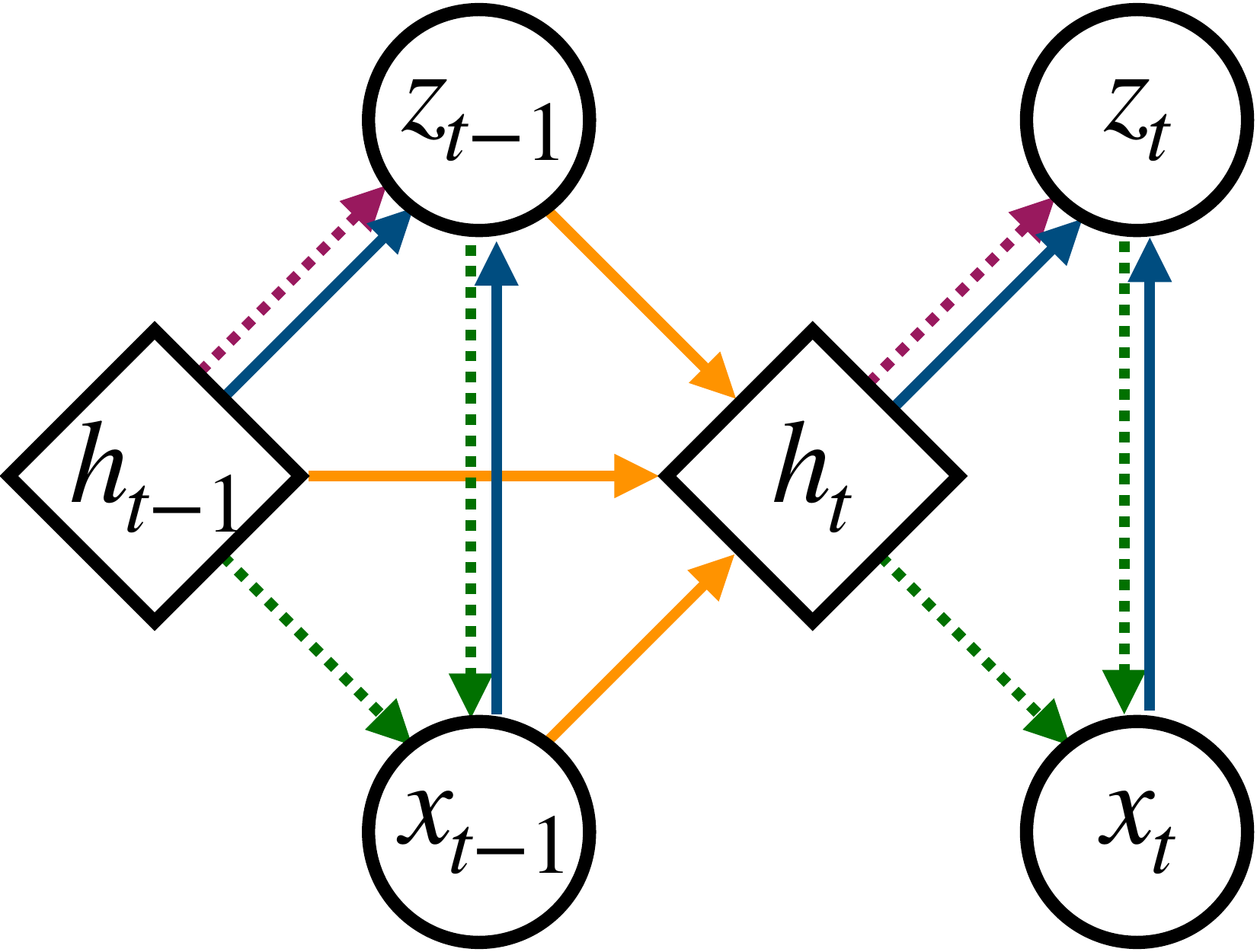}}\end{minipage} & \begin{minipage}{0.25\textwidth}\centering\scalebox{0.2}{\includegraphics{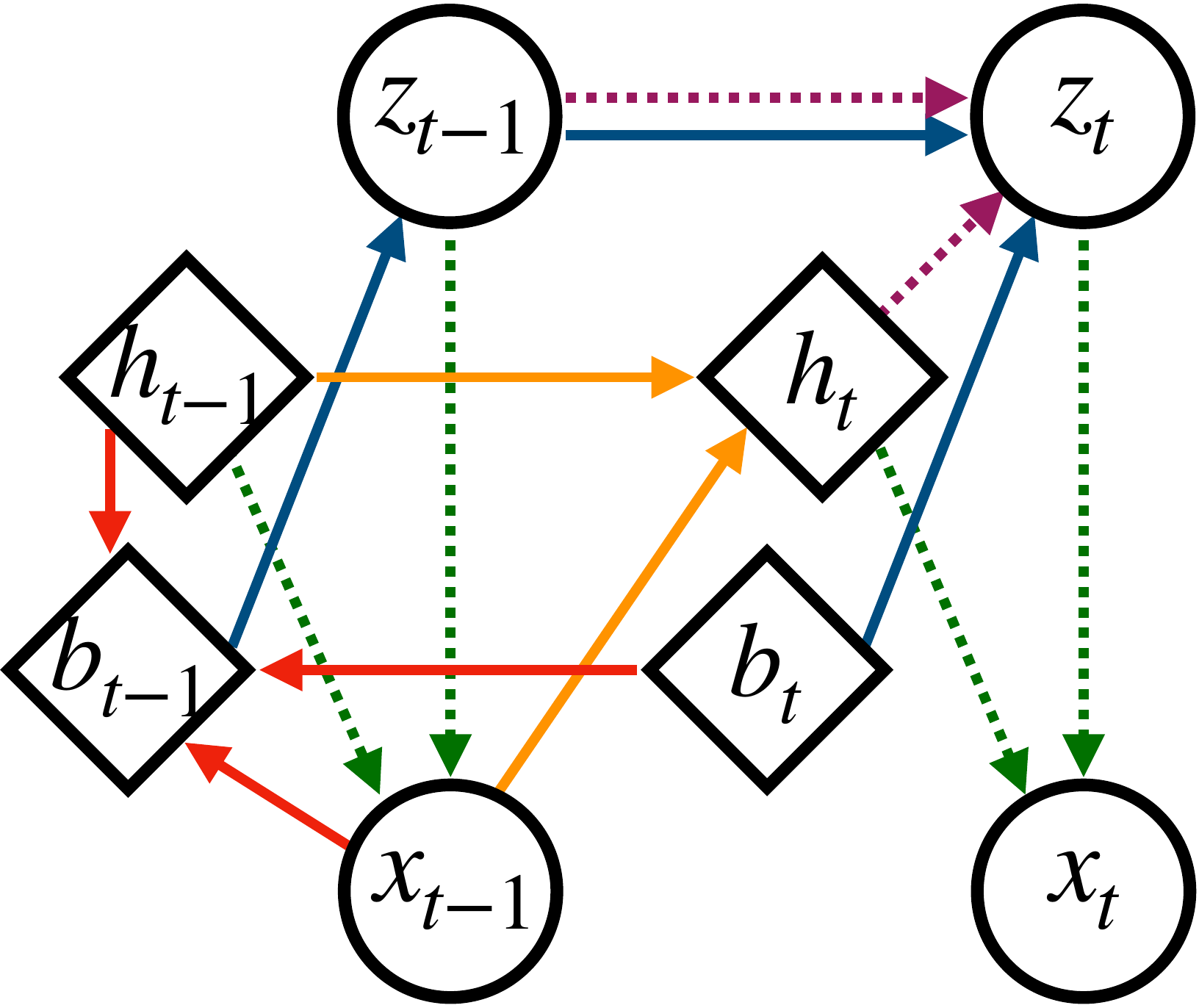}}\end{minipage} & \begin{minipage}{0.25\textwidth}\centering\scalebox{0.2}{\includegraphics{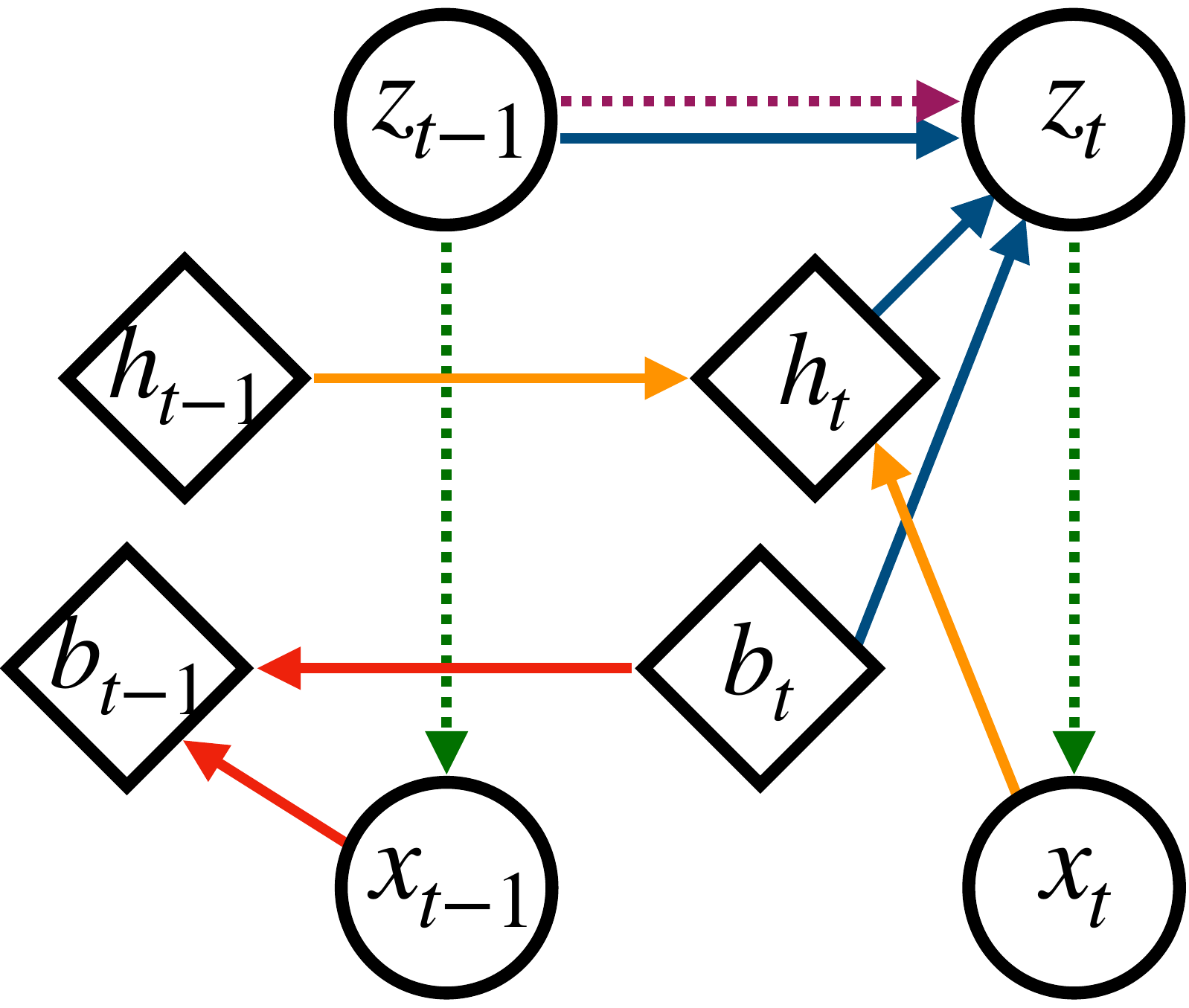}}\end{minipage} \\
\bottomrule
\end{tabular}
\end{small}
\end{center}
\vskip -0.1in
\end{table}

Sequential VAEs aim to learn the true PTSM by deep learning.
Table \ref{tab:svae} shows probablistic models and structures of VRNN \citep{chu15}, SRNN \citep{fra16}, and SVO \citep{mor19s}.
In the graph, the solid and dotted lines represent inferential and generative processes, and the circles and squares represent random and deterministic variables, respectively.
While VRNN infers the latent states from past and current observations, SRNN and SVO use future observations.
SVO restricts PTSM to SSM, which satisfies Markov property.
Although each structure is different, the FIVO and the proposed EnKO framework can be applied to all structures.

\paragraph{Ensemble Kalman Filter}
As per sequential Monte Carlo (SMC) methods used in \citep{le18,mad17,nae18,mor19s,mor19p}, the ensemble Kalman filter (EnKF) represents the latent states using $N$ particles.
The original one \citep{eve94} assumes linear observation model with additive noise
\begin{equation}
\boldsymbol{x}_t=H_t\boldsymbol{z}_t+\boldsymbol{w}_t,\ \boldsymbol{w}_t\sim p_{\boldsymbol{\theta},w}(\boldsymbol{w}_t).
\end{equation}
The EnKF updates the particles by
\begin{subequations}
\begin{align}
\boldsymbol{z}_t^{(i)}&\sim f_{\boldsymbol{\theta}}(\boldsymbol{z}_t|\boldsymbol{z}_{1:t-1}^{(i)}),\ \forall i\in\mathbb{N}_N\\
\bar{\boldsymbol{z}}_t&=\frac{1}{N}\sum_{i=1}^N\boldsymbol{z}_t^{(i)},\label{eq:enkf1}\\
\Sigma^z_t&=\frac{1}{N-1}\sum_{i=1}^N(\boldsymbol{z}_t^{(i)}-\bar{\boldsymbol{z}}_t)(\boldsymbol{z}_t^{(i)}-\bar{\boldsymbol{z}}_t)^T,\label{eq:enkf2}\\
\boldsymbol{w}_t^{(i)}&\sim p_{\boldsymbol{\theta},w}(\boldsymbol{w}_t),\ \forall i\in\mathbb{N}_N,\label{eq:enkf3}\\
\bar{\boldsymbol{w}}_t&=\frac{1}{N}\sum_{i=1}^N\boldsymbol{w}_t^{(i)},\label{eq:enkf4}\\
\Sigma^w_t&=\frac{1}{N-1}\sum_{i=1}^N(\boldsymbol{w}_t^{(i)}-\bar{\boldsymbol{w}}_t)(\boldsymbol{w}_t^{(i)}-\bar{\boldsymbol{w}}_t)^T,\label{eq:enkf5}\\
K_t&=\Sigma^z_tH_t^T(H_t\Sigma^z_tH_t^T+\Sigma^w_t)^{-1},\label{eq:enkf6}\\
\boldsymbol{z}_t^{f,(i)}&=\boldsymbol{z}_t^{(i)}+K_t(\boldsymbol{x}_t-H_t\boldsymbol{z}_t^{(i)}-\boldsymbol{w}_t^{(i)}),\ \forall i\in\mathbb{N}_N,\label{eq:enkf7}
\end{align}
\label{eq:enkf}
\end{subequations}
where $\mathbb{N}_N=\{1,\cdots,N\}$, superscript $T$ indicates vector or matrix transpose, and $\boldsymbol{z}_t^{f,(i)}$ denotes the $i$-th latent particle after filtering at time $t$.

The original EnKF is easily applied to nonlinear observation model
\begin{equation}
\boldsymbol{x}_t=h_{\boldsymbol{\theta}}(\boldsymbol{z}_t)+\boldsymbol{w}_t,\ \boldsymbol{w}_t\sim p_{\boldsymbol{\theta},w}(\boldsymbol{w}_t)
\end{equation}
by the augmented PTSM
\begin{subequations} 
\begin{align}
\tilde{\boldsymbol{z}}_t&=\begin{pmatrix}\boldsymbol{z}_t\\h_{\boldsymbol{\theta}}(\boldsymbol{z}_t)\end{pmatrix}\notag\\
&\sim\tilde f_{\boldsymbol{\theta}}(\tilde{\boldsymbol{z}}_t|\tilde{\boldsymbol{z}}_{1:t-1})=f_{\boldsymbol{\theta}}\left(\begin{pmatrix}I_{d_z}&O_{d_x}\end{pmatrix}\tilde{\boldsymbol{z}}_t|\begin{pmatrix}I_{d_z}&O_{d_x}\end{pmatrix}\tilde{\boldsymbol{z}}_{1:t-1}\right),\label{eq:assm1}\\
\boldsymbol{x}_t&=\begin{pmatrix}O_{d_z}&I_{d_x}\end{pmatrix}\tilde{\boldsymbol{z}}_t+\boldsymbol{w}_t=\tilde H_t\tilde{\boldsymbol{z}}_t+\boldsymbol{w}_t\notag\\
&\sim\tilde g_{\boldsymbol{\theta}}(\boldsymbol{x}_t|\tilde{\boldsymbol{z}}_t)=g_{\boldsymbol{\theta}}\left(\boldsymbol{x}_t|\begin{pmatrix}I_{d_z}&O_{d_x}\end{pmatrix}\tilde{\boldsymbol{z}}_t\right)\label{eq:assm2}
\end{align}
\end{subequations}
for the augmented latent states $\tilde{\boldsymbol{z}}_t\in\mathbb{R}^{d_z+d_x}$, where $I_d$ and $O_d$ are the $d$-dimensional square identity matrix and zero matrix, respectively.
By virtue of equation \eqref{eq:assm2}, the nonlinear emission is regarded as a linear representation; then, the augmented PTSM can be applied to \eqref{eq:enkf}.

Intuitively, the EnKF propagates particles for explaining observations well considering their sample covariance as shown in Figure \ref{fig:filter} (right).
The EnKF method provides particle diversity, whereas the SMC method is often faced with poor diversity due to its inherent properties \citep{duc11}.

\figimagetwo{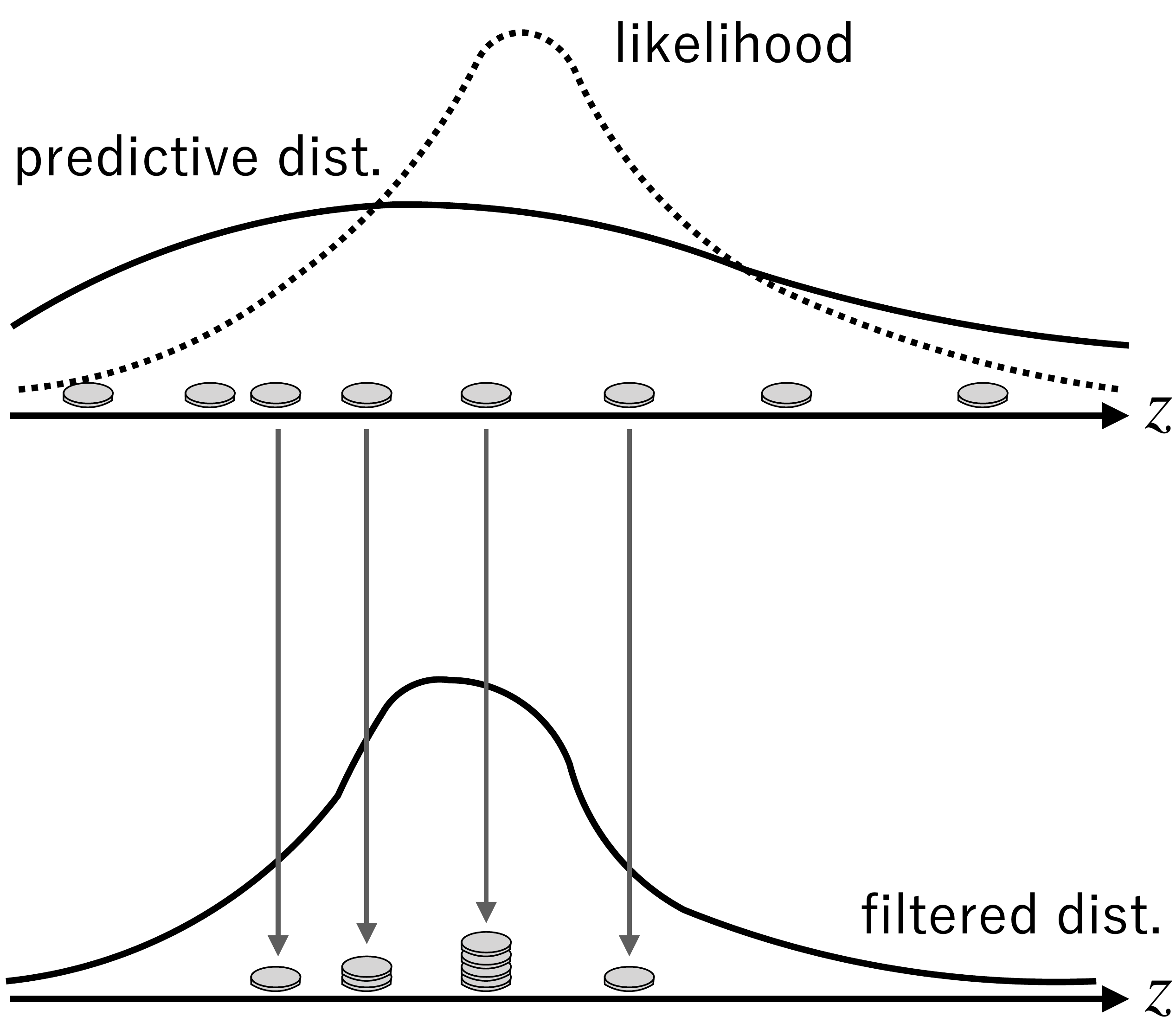}{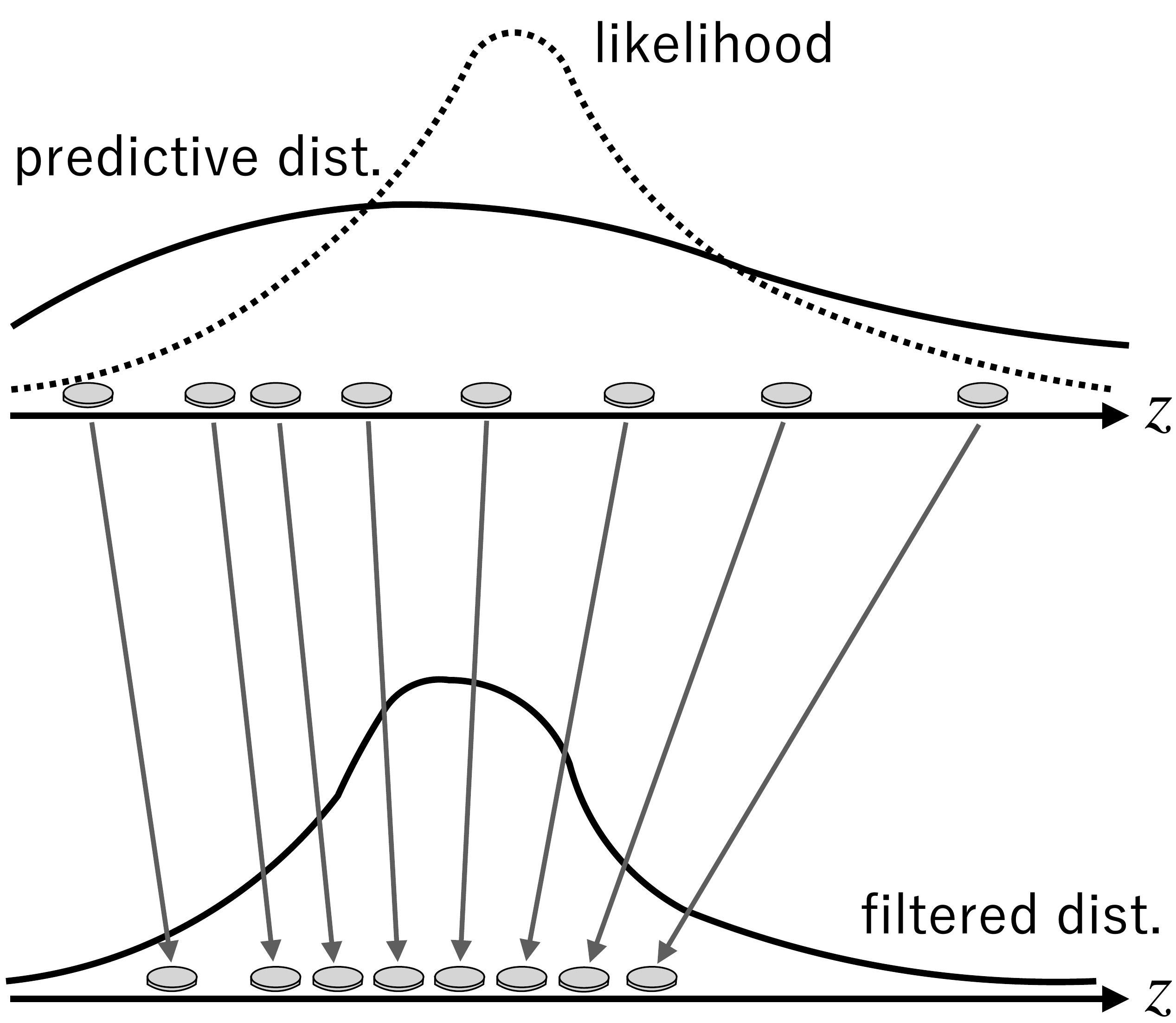}{Comparison of SMC and the EnKF in filtering step. (left) SMC resamples particles with their weights. (right) The EnKF propagates particles for explaining the observation well, considering their sample covariance.}{fig:filter}{0.45}{0.45}

\paragraph{Covariance Inflation}
It is generally known that the EnKF underestimates the state covariance matrix
\begin{equation}
\Sigma^{z^f}_t=\frac{1}{N-1}\sum_{i=1}^N(\boldsymbol{z}_t^{f,(i)}-\bar{\boldsymbol{z}}^f_t)(\boldsymbol{z}_t^{f,(i)}-\bar{\boldsymbol{z}}^f_t)^T
\end{equation}
due to several factors such as limited particle size and model error, where $\bar{\boldsymbol{z}}^f_t$ is the average of $\{\boldsymbol{z}_t^{f,(i)}\}_{i=1}^N$.
To overcome this problem, covariance inflation methods that inflate the covariance have been proposed \citep{and99,mit00,cor03,zha04,whi12}.
The representative methods are multiplicative inflation \citep{and99}, additive inflation \citep{mit00,cor03}, and relaxation to prior \citep{zha04,whi12}.
The relaxation to prior (RTP) methods relax the reduction of the spread of particles at the filtering step and are suitable for our proposed algorithm.
The relaxation to prior perturbation (RTPP) \citep{zha04} method blends the particles before and after filtering; the mathematical equation is
\begin{equation}
\boldsymbol{z}_t^{c,(i)}=\alpha\boldsymbol{z}_t^{(i)}+(1-\alpha)\boldsymbol{z}_t^{f,(i)},\label{eq:rtpp}
\end{equation}
where $\alpha\in[0,1]$ is known as the inflation factor and $\boldsymbol{z}_t^{c,(i)}$ denotes the $i$-th latent particle after covariance inflation.
The relaxation to prior spread (RTPS) \citep{whi12} method multiplies the particles after filtering and inflation together; the formulation is
\begin{equation}
\boldsymbol{z}_t^{c,(i)}=(\alpha\boldsymbol{\sigma}_t+(1-\alpha)\boldsymbol{\sigma}_t^f)\oslash\boldsymbol{\sigma}_t^f\odot\boldsymbol{z}_t^{f,(i)},
\end{equation}
where $\odot$ is element-wise multiplication, $\oslash$ is element-wise division, $\alpha\in[0,1]$ is the inflation factor, and $\boldsymbol{\sigma}_t\in\mathbb{R}^{d_z}$ and $\boldsymbol{\sigma}_t^f\in\mathbb{R}^{d_z}$ are the sample variance of particles before and after filtering, respectively.

\section{Ensemble Kalman Variational Objective (EnKO)}
\label{sec:enko}

\begin{algorithm}[ht]
\caption{Ensemble Kalman Variational Objectives}
\label{alg:enko}
\begin{multicols}{2}
\begin{algorithmic}[1]
\STATE {\bf EnKO}$(\boldsymbol{x}_{1:T},p_{\boldsymbol{\theta}},q_{\boldsymbol{\varphi}},N)$:
\FOR{$t\in\{1,\cdots,T\}$}
\FOR{$i\in\{1,\cdots,N\}$}
\IF{$t=1$}
\STATE $\boldsymbol{z}_1^{(i)}\sim q_{\boldsymbol{\varphi}}(\boldsymbol{z}_1|\boldsymbol{x}_{1:T})$
\ELSE
\STATE $\boldsymbol{z}_t^{(i)}\sim q_{\boldsymbol{\varphi}}(\boldsymbol{z}_t|\boldsymbol{x}_{1:T},\boldsymbol{z}_{1:t-1}^{f,(i)})$
\STATE $\boldsymbol{z}_{1:t}^{(i)}=${\bf CONCAT}$(\boldsymbol{z}_{1:t-1}^{(i)},\boldsymbol{z}_t^{(i)})$
\ENDIF
\STATE $w_t^{(i)}=\frac{p_{\boldsymbol{\theta}}(\boldsymbol{x}_t,\boldsymbol{z}_t^{(i)}|\boldsymbol{z}_{1:t-1}^{(i)})}{q_{\boldsymbol{\varphi}}(\boldsymbol{z}_t^{(i)}|\boldsymbol{x}_{1:T},\boldsymbol{z}_{1:t-1}^{f,(i)})}$
\ENDFOR
\STATE $\{\boldsymbol{z}_t^{f,(i)}\}_{i=1}^N=${\bf EnKF}$(\{\boldsymbol{z}_t^{(i)}\}_{i=1}^N,\boldsymbol{x}_t)$
\ENDFOR
\OUTPUT $\hat p_N(\boldsymbol{x}_{1:T})=\frac{1}{N}\sum_{i=1}^N\prod_{t=1}^Tw_t^{(i)}$
\newline
\newline
\newline
\STATE {\bf EnKF}$(\{\boldsymbol{z}_t^{(i)}\}_{i=1}^N,\boldsymbol{x}_t)$:
\FOR{$i\in\{1,\cdots,N\}$}
\STATE $\boldsymbol{x}_t^{(i)}\sim g_{\boldsymbol{\theta}}(\boldsymbol{x}_t|\boldsymbol{z}_t^{(i)})$
\STATE $\boldsymbol{\mu}_t^{x,(i)}=\mathbb{E}_g[\boldsymbol{x}_t^{(i)}]=h_{\boldsymbol{\theta}}(\boldsymbol{z}_t^{(i)})$
\ENDFOR
\STATE $\bar{\boldsymbol{x}}_t=\frac{1}{N}\sum_{i=1}^N\boldsymbol{x}_t^{(i)}$
\STATE $\bar{\boldsymbol{\mu}}_t^x=\frac{1}{N}\sum_{i=1}^N\boldsymbol{\mu}_t^{x,(i)}$
\STATE $\bar{\boldsymbol{z}}_t=\frac{1}{N}\sum_{i=1}^N\boldsymbol{z}_t^{(i)}$
\STATE $\Sigma_t^x=\frac{1}{N-1}\sum_{i=1}^N(\boldsymbol{x}_t^{(i)}-\bar{\boldsymbol{x}}_t)(\boldsymbol{x}_t^{(i)}-\bar{\boldsymbol{x}}_t)^T$
\STATE $\Sigma_t^{z\mu^x}=\frac{1}{N-1}\sum_{i=1}^N(\boldsymbol{z}_t^{(i)}-\bar{\boldsymbol{z}}_t)(\boldsymbol{\mu}_t^{x,(i)}-\bar{\boldsymbol{\mu}}_t^x)^T$
\STATE $K_t=\Sigma_t^{z\mu^x}(\Sigma_t^x)^{-1}$
\FOR{$i\in\{1,\cdots,N\}$}
\STATE $\boldsymbol{z}_t^{f,(i)}=\boldsymbol{z}_t^{(i)}+K_t(\boldsymbol{x}_t-\boldsymbol{x}_t^{(i)})$
\ENDFOR
\STATE $\{\boldsymbol{z}_t^{c,(i)}\}_{i=1}^N=${\bf CI}$(\{\boldsymbol{z}_t^{(i)}\}_{i=1}^N,\{\boldsymbol{z}_t^{f,(i)}\}_{i=1}^N)$
\OUTPUT $\{\boldsymbol{z}_t^{c,(i)}\}_{i=1}^N$
\end{algorithmic}
\end{multicols}
\end{algorithm}

\begin{table*}[t]
\caption{Comparison among ensemble systems}
\label{tab:comp}
\vskip 0.05in
\begin{center}
\begin{small}
\begin{tabular}{lccccc}
\toprule
System&\multicolumn{2}{c}{Unbiasedness}&Diversity&Low variance&Robustness\\
& Gradient & Objective &&\\
\midrule
IWAE& \multirow{2}{*}{$\surd$} & \multirow{2}{*}{$\surd$} & \multirow{2}{*}{$\surd$} & \multirow{2}{*}{$\times$} & \multirow{2}{*}{$\surd$} \\
\citep{bur15} &&&&\\
FIVO& \multirow{2}{*}{$\times$} & \multirow{2}{*}{$\surd$} & \multirow{2}{*}{$\times$} & \multirow{2}{*}{$\times$} & \multirow{2}{*}{$\surd$} \\
\citep{mad17} &&&&\\
PSVO& \multirow{2}{*}{$\times$} & \multirow{2}{*}{$\surd$} & \multirow{2}{*}{$\times$} & \multirow{2}{*}{$\times$} & \multirow{2}{*}{$\times$} \\
\citep{mor19p} &&&&\\
EnKO& \multirow{2}{*}{$\surd$} & \multirow{2}{*}{$\times$} & \multirow{2}{*}{$\surd$} & \multirow{2}{*}{$\surd$} & \multirow{2}{*}{$\surd$} \\
(current work) &&&&\\
\bottomrule
\end{tabular}
\end{small}
\end{center}
\vskip -0.1in
\end{table*}

{\it Ensemble Kalman Variational Objective} (EnKO) is a framework to infer the latent time-series model using the EnKF in the inference phase.
Although the overall framework is similar to the SMC-based VI methods such as FIVO \citep{mad17}, the only but crucial difference is the nonlinear filtering step.
Thanks to the property of the EnKF in EnKO, the proposed method has three advantages over the SMC-based methods as shown in Table \ref{tab:comp}:
\begin{enumerate}
\item Particle diversity is maintained because the EnKF is not associated with the particle degeneracy problem, unlike the FIVO.
\item The variance of the gradient estimator is less than that of the FIVO.
Although the FIVO uses the biased gradient estimator because the full gradient estimator has a resampling term with high variance, the variance of the biased gradient estimator is higher than the full gradient estimator of the EnKO.
This low variance is experimentally shown in Appendix \ref{sec:grad}.
\item Unbiasedness of the gradient estimator is guaranteed because of low variance of the gradient estimators.
A detail of the gradient estimator is shown in Appendix \ref{sec:grad}.
\end{enumerate}
The rest of this section is laid out as follows.
We first describe a detailed algorithm of the proposed method in the following subsection.
We then introduce our objective function and its property in subsection \ref{ssec:obj}.
Finally, subsection \ref{ssec:high} is devoted to a technique to apply the proposed method to high-dimensional data.

\subsection{Algorithm of the Proposed Method}
\label{ssec:alg}
\paragraph{Overall Algorithm}
We describe the overall algorithm of EnKO with Algorithm \ref{alg:enko}.
First, the method infers the initial latent state $\boldsymbol{z}_1$ from the observations $X=\boldsymbol{x}_{1:T}$ by neural networks, e.g., bidirectional LSTM (biLSTM) in SRNN \citep{fra16}, biLSTM plus Dense NN in SVO \citep{mor19s} ({\it line 5}).
At each time-step $t$, the latent state $\boldsymbol{z}_t^{(i)}$ is inferred from the observations $X$ and the latent states until the previous time-step $\boldsymbol{z}_{1:t-1}^{(i)}$ ({\it line 7}).
The importance weight $w_t^{(i)}$ is calculated by
\begin{align}
w_t^{(i)}&=\frac{p_{\boldsymbol{\theta}}(\boldsymbol{x}_t,\boldsymbol{z}_t^{(i)}|\boldsymbol{z}_{1:t-1}^{(i)})}{q_{\boldsymbol{\varphi}}(\boldsymbol{z}_t^{(i)}|\boldsymbol{x}_{1:T},\boldsymbol{z}_{1:t-1}^{(i)})}\\
&=\frac{g_{\boldsymbol{\theta}}(\boldsymbol{x}_t|\boldsymbol{z}_{t}^{(i)})f_{\boldsymbol{\theta}}(\boldsymbol{z}_t^{(i)}|\boldsymbol{z}_{1:t-1}^{(i)})}{q_{\boldsymbol{\varphi}}(\boldsymbol{z}_t^{(i)}|\boldsymbol{x}_{1:T},\boldsymbol{z}_{1:t-1}^{(i)})},\label{eq:weight2}
\end{align}
for the objective function ({\it line 10}).
The latent states $\{\boldsymbol{z}_t^{(i)}\}_{i=1}^N$ are corrected by the EnKF using their sample covariance and the observation at time $t$ ({\it line 12}).
Finally, the $\hat p_N(\boldsymbol{x}_{1:t})$ are computed ({\it output}).

\paragraph{Detailed Algorithm of the EnKF}
We describe the detailed algorithm of the EnKF in EnKO with Algorithm \ref{alg:enko}.
At each time-step $t$ and for each particle $i$, the mean estimator of the observation $\boldsymbol{\mu}_t^{x,(i)}$ and a sample $\boldsymbol{x}_t^{(i)}$ is generated from the emission network $g_{\boldsymbol{\theta}}(\boldsymbol{x}_t|\boldsymbol{z}_t^{(i)})$ ({\it lines 16-17}).
For example, in the Gaussian output distribution $g_{\boldsymbol{\theta}}(\boldsymbol{x}_t|\boldsymbol{z}_t^{(i)})=N(h_{\boldsymbol{\theta}}(\boldsymbol{z}_t^{(i)}),s_{\boldsymbol{\theta}}(\boldsymbol{z}_t^{(i)}))$, the mean estimator $\boldsymbol{\mu}_t^{x,(i)}=h_{\boldsymbol{\theta}}(\boldsymbol{z}_t^{(i)})$ is gained.
In this example, $\boldsymbol{w}_t^{(i)}=\boldsymbol{x}_t^{(i)}-\boldsymbol{\mu}_t^{x,(i)}$ follows the zero-mean Gaussian distribution $N(\boldsymbol{0},s_{\boldsymbol{\theta}}(\boldsymbol{z}_t^{(i)}))$ and this random variable corresponds to the noise at \eqref{eq:enkf3} in the EnKF update formula.
Covariance matrices $\Sigma^x_t$ and $\begin{pmatrix}\Sigma_t^{z\mu^x}&\Sigma_t^{\mu^x}\end{pmatrix}^T$ correspond to $H_t\Sigma^{\tilde z}_tH_t^T+\Sigma^w_t$ and $\Sigma^{\tilde z}_tH_t^T$, respectively, at \eqref{eq:enkf6} in the augmented system, where $H_t=\begin{pmatrix}O_{d_z}&I_{d_x}\end{pmatrix}$ ({\it lines 22-23}).
In the augmented system, because the first $d_z$ coordinates of $\tilde{\boldsymbol{z}}_t$ constitute the only update target, the augmented gain $\tilde K_t\in\mathbb{R}^{(d_z+d_x)\times d_z}$ is only needed at the first $d_z$ rows $K_t\in\mathbb{R}^{d_z\times d_x}$ ({\it line 24}).
The latent state $\boldsymbol{z}_t^{(i)}$ is updated to $\boldsymbol{z}_t^{f,(i)}$ by the gain explaining observation $\boldsymbol{x}_t$ well ({\it line 26}).
Finally, the updated latent states $\{\boldsymbol{z}_t^{f,(i)}\}_{i=1}^N$ are adjusted by covariance inflation methods such as RTPP and RTPS ({\it line 28}).

\subsection{Objective Function}
\label{ssec:obj}
We train the EnKO to maximize the objective function
\begin{align}
\mathcal{L}_{\mathrm{EnKO}}^N(\boldsymbol{\theta},\boldsymbol{\varphi},X):&=\mathbb{E}_{Q_{\mathrm{EnKO}}(\boldsymbol{z}_{1:T}^{(1:N)},\boldsymbol{x}_{1:T}^{(1:N)}|X)}[\log\hat p_N(\boldsymbol{x}_{1:T})],\\
\hat p_N(\boldsymbol{x}_{1:T})&=\frac{1}{N}\sum_{i=1}^N\prod_{t=1}^T\frac{f_{\boldsymbol{\theta}}(\boldsymbol{z}_t^{(i)}|\boldsymbol{z}^{(i)}_{1:t-1})g_{\boldsymbol{\theta}}(\boldsymbol{x}_t|\boldsymbol{z}_t^{(i)})}{q_{\boldsymbol{\varphi}}(\boldsymbol{z}_t^{(i)}|\boldsymbol{x}_{1:T},\boldsymbol{z}_{1:t-1}^{f,(i)})},\\
Q_{\mathrm{EnKO}}(\boldsymbol{z}_{1:T}^{(1:N)},\boldsymbol{x}_{1:T}^{(1:N)}|X)&=\prod_{t=1}^T\prod_{i=1}^Nq_{\boldsymbol{\varphi}}(\boldsymbol{z}_t^{(i)}|\boldsymbol{x}_{1:T},\boldsymbol{z}_{1:t-1}^{f,(i)})g_{\boldsymbol{\theta}}(\boldsymbol{x}_t^{(i)}|\boldsymbol{z}_t^{(i)}),
\end{align}
where $f_{\boldsymbol{\theta}}(\boldsymbol{z}_1|\boldsymbol{z}_{1:0})=f_{\boldsymbol{\theta}}(\boldsymbol{z}_1)$, $q_{\boldsymbol{\varphi}}(\boldsymbol{z}_1^{(i)}|\boldsymbol{x}_{1:T},\boldsymbol{z}_{1:0}^{f,(i)})=q_{\boldsymbol{\varphi}}(\boldsymbol{z}_1^{(i)}|\boldsymbol{x}_{1:T})$ and $\hat p_N(\boldsymbol{x}_{1:T})$ is the output of Algorithm \ref{alg:enko}.
This formulation is the same as the objective of sequential IWAE \citep{bur15} as shown in equation \eqref{eq:iwae} and slightly different from FIVO \citep{mad17}.

The following theorem and corollary are easily proved.
\begin{theorem}
The $\hat p_N(\boldsymbol{x}_{1:T})$ is an approximately unbiased estimator of the marginal likelihood $p(\boldsymbol{x}_{1:T})$.
If the emission distribution $g_{\boldsymbol{\theta}}(\boldsymbol{x}_t|\boldsymbol{z}_t)$ and the variational distribution $q_{\boldsymbol{\varphi}}(\boldsymbol{z}_t|\boldsymbol{x}_{1:T},\boldsymbol{z}_{1:t-1}^{f})$ are Gaussian, the $\hat p_N(\boldsymbol{x}_{1:T})$ is an unbiased estimator of the marginal likelihood.
\end{theorem}
\begin{corollary}
The objective function $\mathcal{L}_{\mathrm{EnKO}}^N(\boldsymbol{\theta},\boldsymbol{\varphi},X)$ is an approximately lower bound of the log marginal likelihood $\log p(\boldsymbol{x}_{1:T})$.
If the emission distribution $g_{\boldsymbol{\theta}}(\boldsymbol{x}_t|\boldsymbol{z}_t)$ and the variational distribution $q_{\boldsymbol{\varphi}}(\boldsymbol{z}_t|\boldsymbol{x}_{1:T},\boldsymbol{z}_{1:t-1}^{f})$ are Gaussian, the objective function is an unbiased estimator of the log marginal likelihood.
\end{corollary}
There is one point to be noted here.
Since the objective function is not an exact lower bound of the log marginal likelihood $\log p(\boldsymbol{x}_{1:T})$, but an approximate lower bound, it is not strictly an ELBO.
However, we describe it as ELBO because it is an approximate lower bound of the log marginal likelihood and an exact lower bound under appropriate constraints.
Our goal is to infer models efficiently with a small number of particles, and we are not concerned with the tightness of the theoretical lower bound.
In fact, in the following section, we show that the proposed method can experimentally infer more appropriate models than IWAE and FIVO.

\subsection{High-dimensional Application of the Proposed Method}
\label{ssec:high}
\figimage{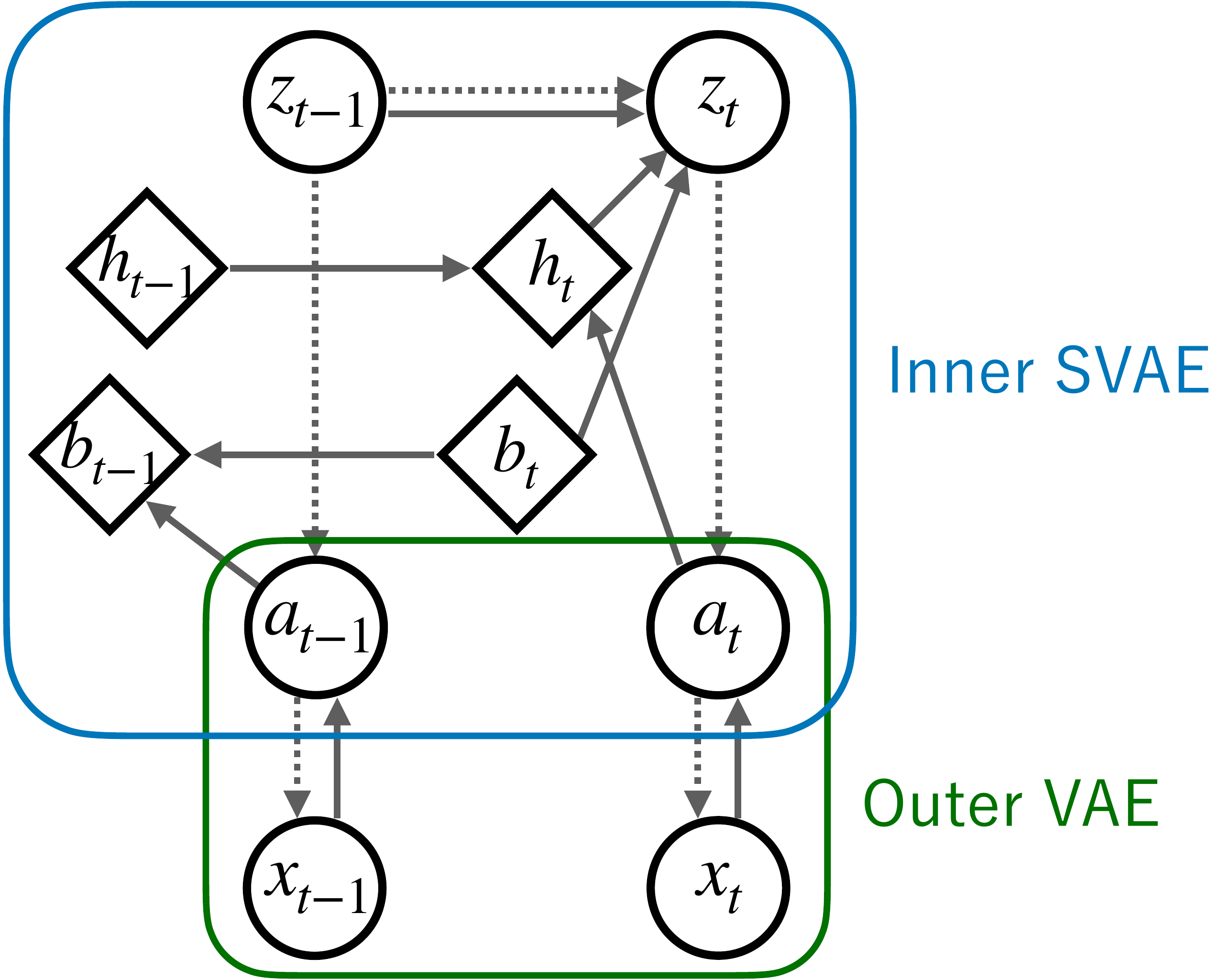}{Hierarchical structure for application to high-dimensional data. We first embed the observed variables $\boldsymbol{x}_t$ to auxiliary variables $\boldsymbol{a}_t$, and then applies SVAE with EnKO to latent variables $\boldsymbol{z}_t$ and $\boldsymbol{a}_t$. In this figure, while SVAE is assumed to SVO, other SVAEs can be applied.}{fig:hdsvae}{0.4}
While the computational complexity of FIVO scales linearly with the observed dimension $d_x$, the computational complexity of EnKO scales quadratically with the dimension $d_x$, making it difficult to apply directly to high-dimensional data.
To solve this problem, we use VAE to embed the observed variables $\boldsymbol{x}_t$ to auxiliary variables $\boldsymbol{a}_t\in\mathbb{R}^{d_a}$, and then propose an alternative that applies SVAE with EnKO to latent variables $\boldsymbol{z}_t$ and auxiliary variables $\boldsymbol{a}_t$ (Figure \ref{fig:hdsvae}).
In this structure, the computational complexity of EnKO scales quadratically with the dimension $d_a$ and linearly with the observed dimension $d_x$, which significantly reduces the computational cost.
In addition, because EnKO has particle diversity, the number of particles used in the calculation can be reduced compared to FIVO, resulting in lower computational costs than FIVO.

\section{Experiments}
\label{sec:exp}
\figimagethree{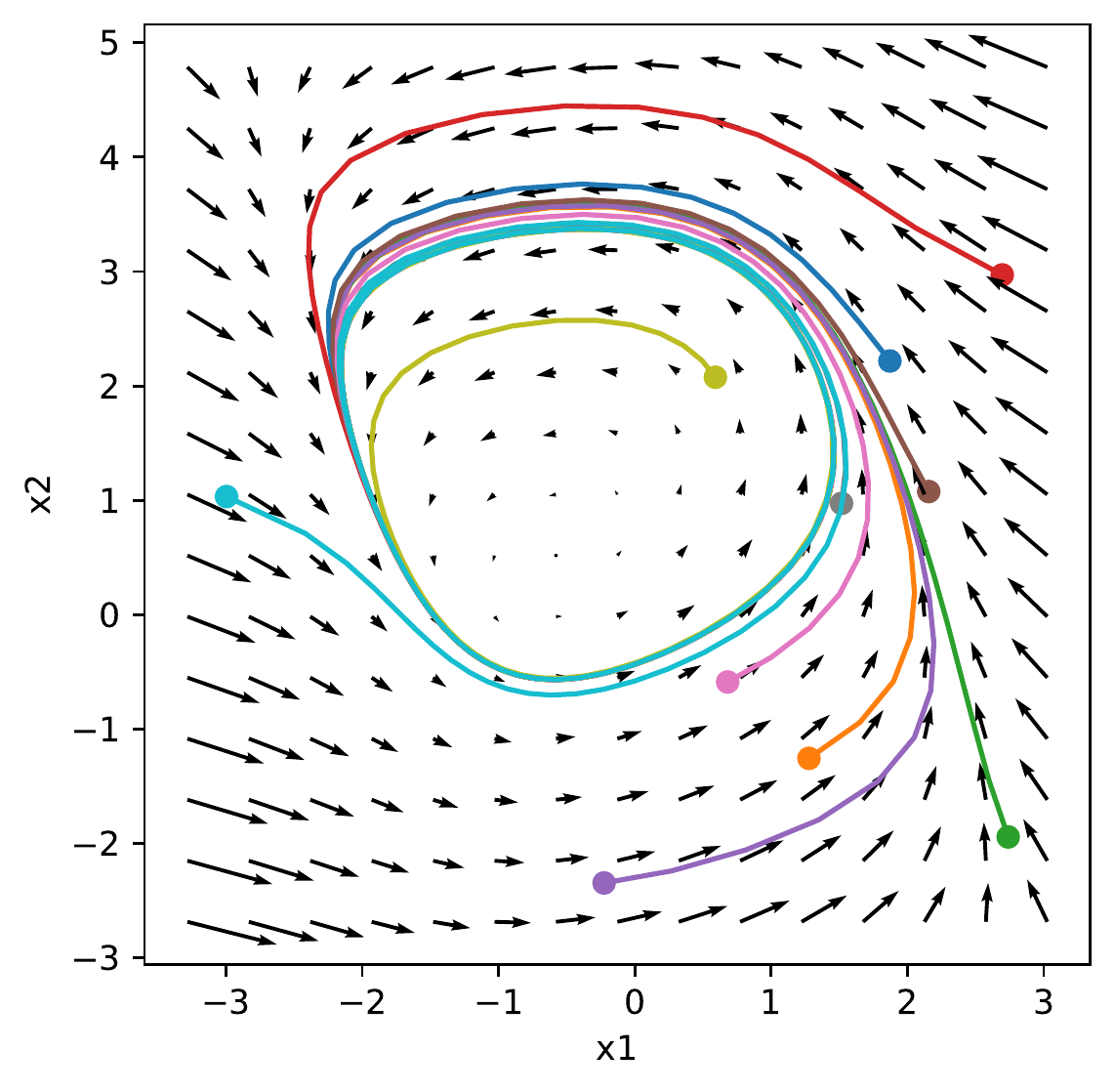}{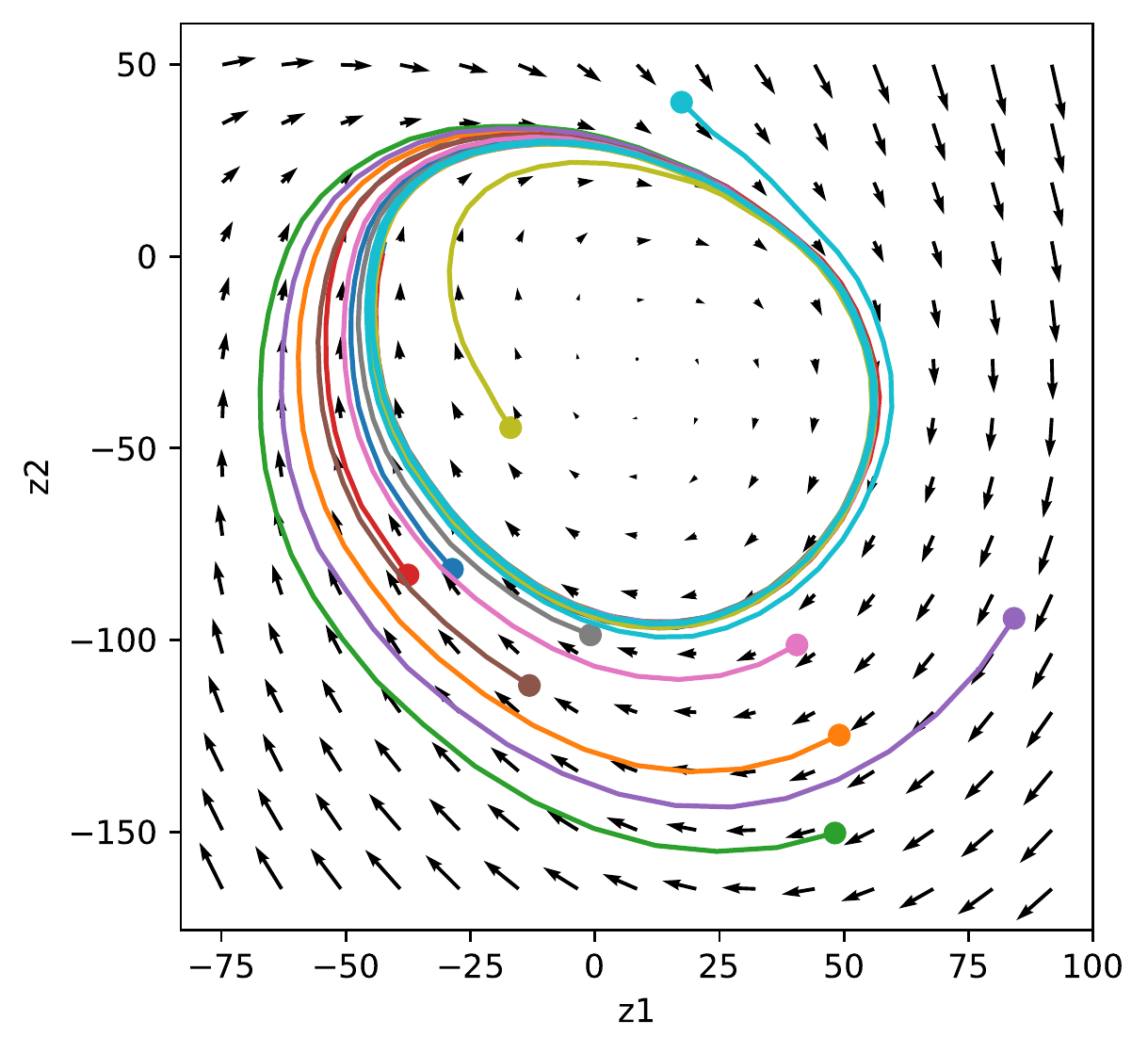}{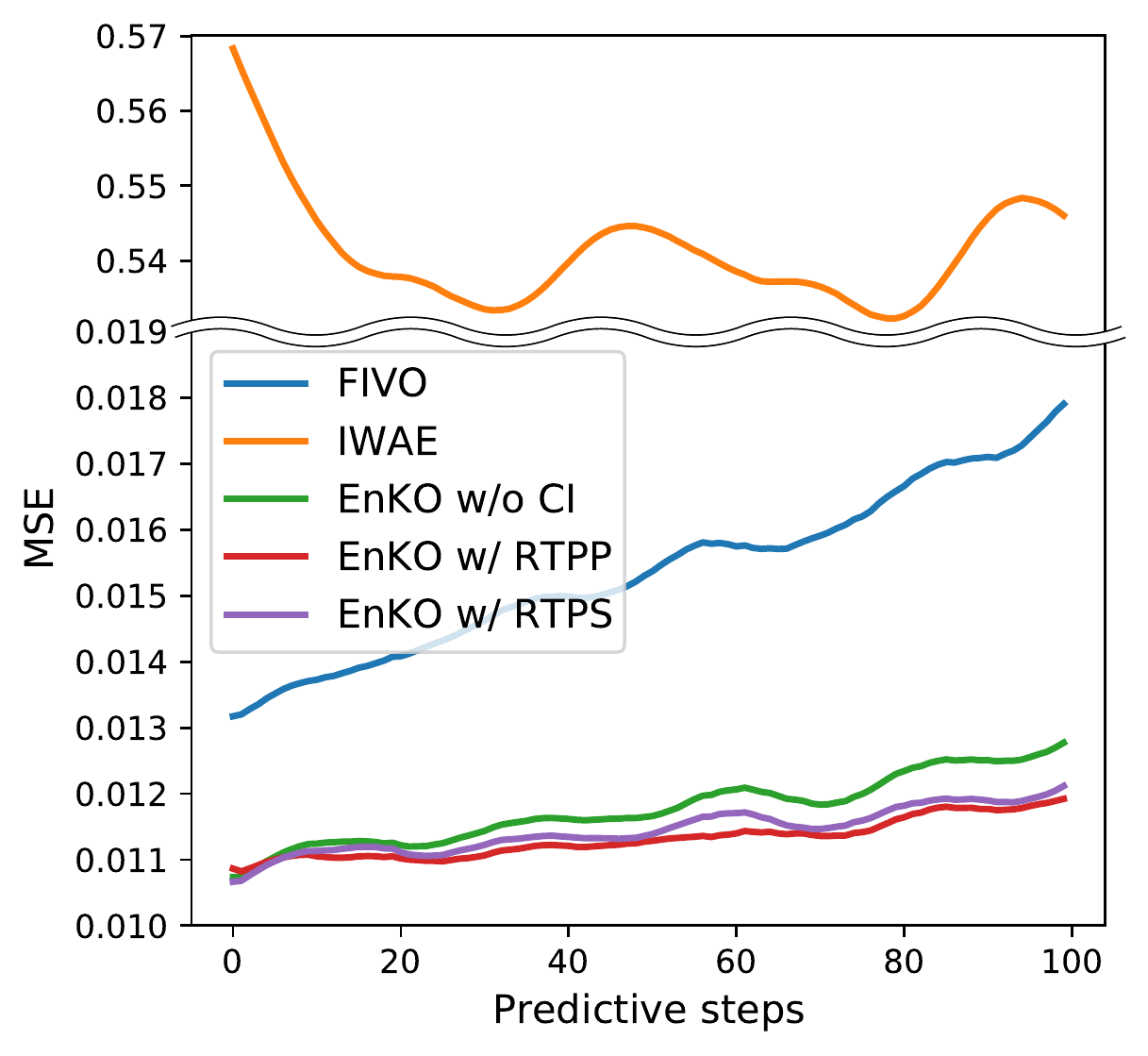}{Inferred latent dynamics and MSE for FitzHugh-Nagumo model. (left) True dynamics and trajectories on the first 10 test set. (center) Inferred latent dynamics and trajectories on the 10 test set using EnKO to perform dimensionality expansion. (right) MSE for various ensemble frameworks, including FIVO ({\it blue}), IWAE ({\it orange}), and EnKO without ({\it green}) and with covariance inflation (RTPP: {\it red}, RTPS: {\it purple}).}{fig:fhn}{0.3}{0.32}{0.32}
\figimagethree{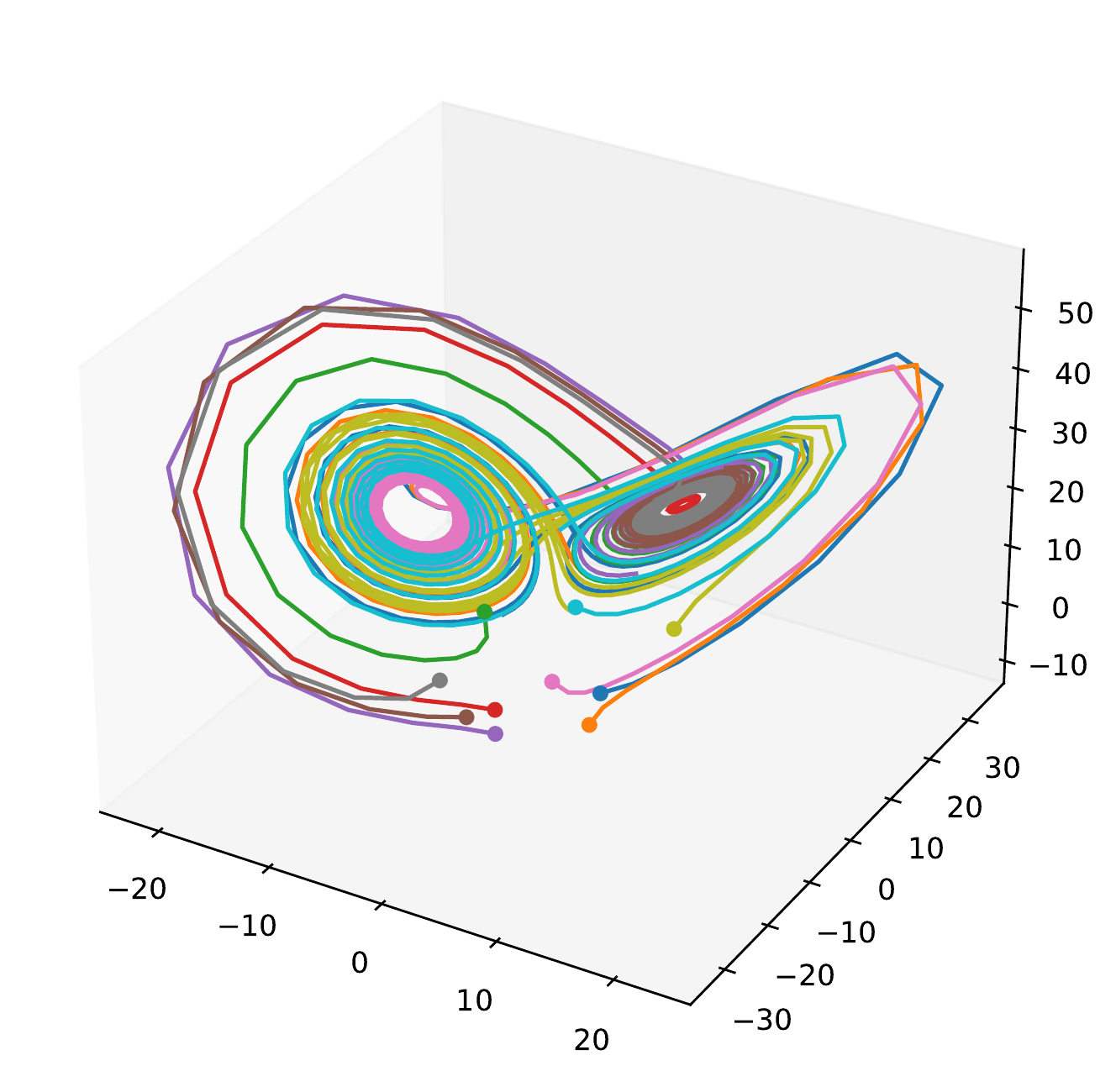}{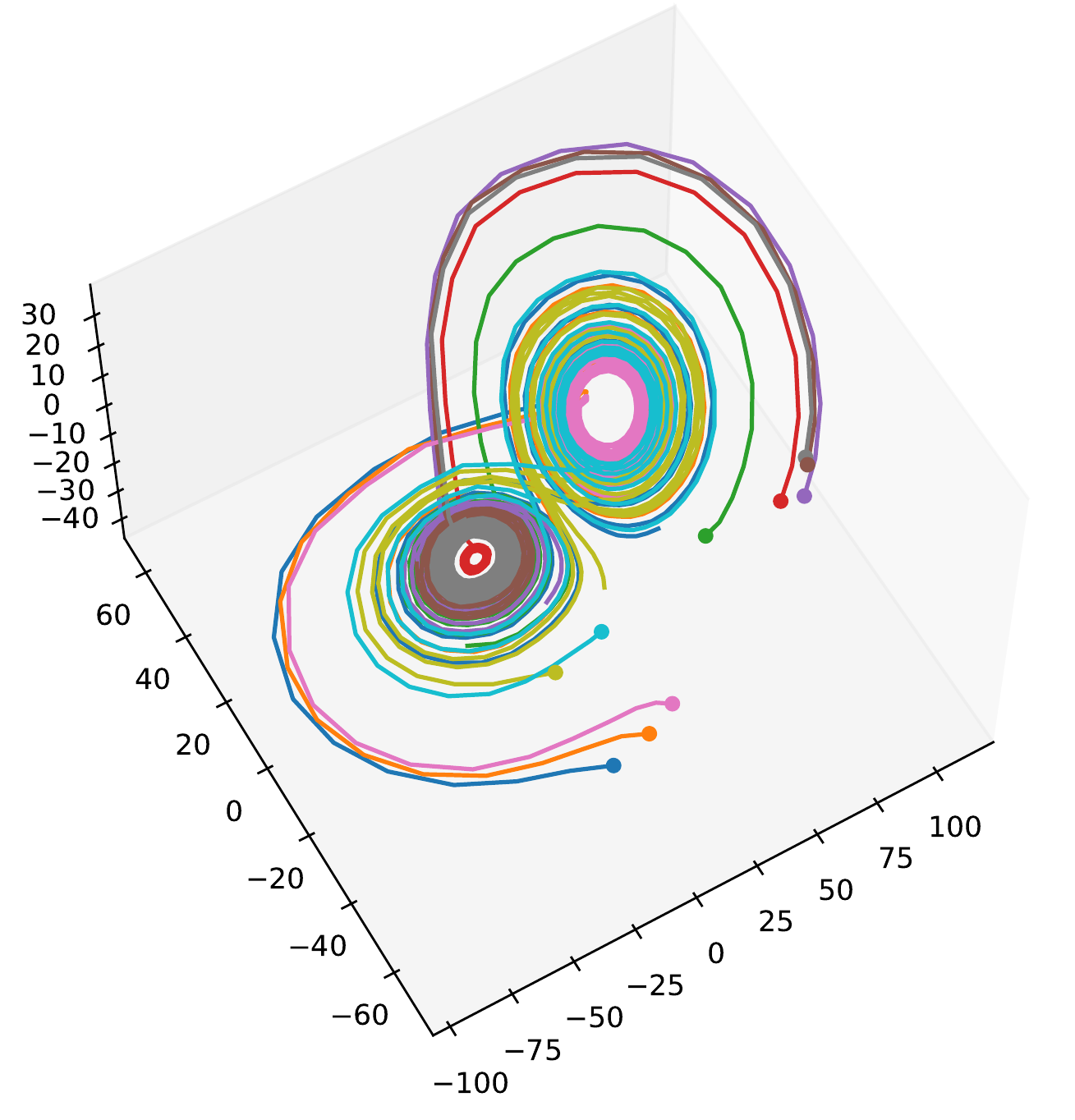}{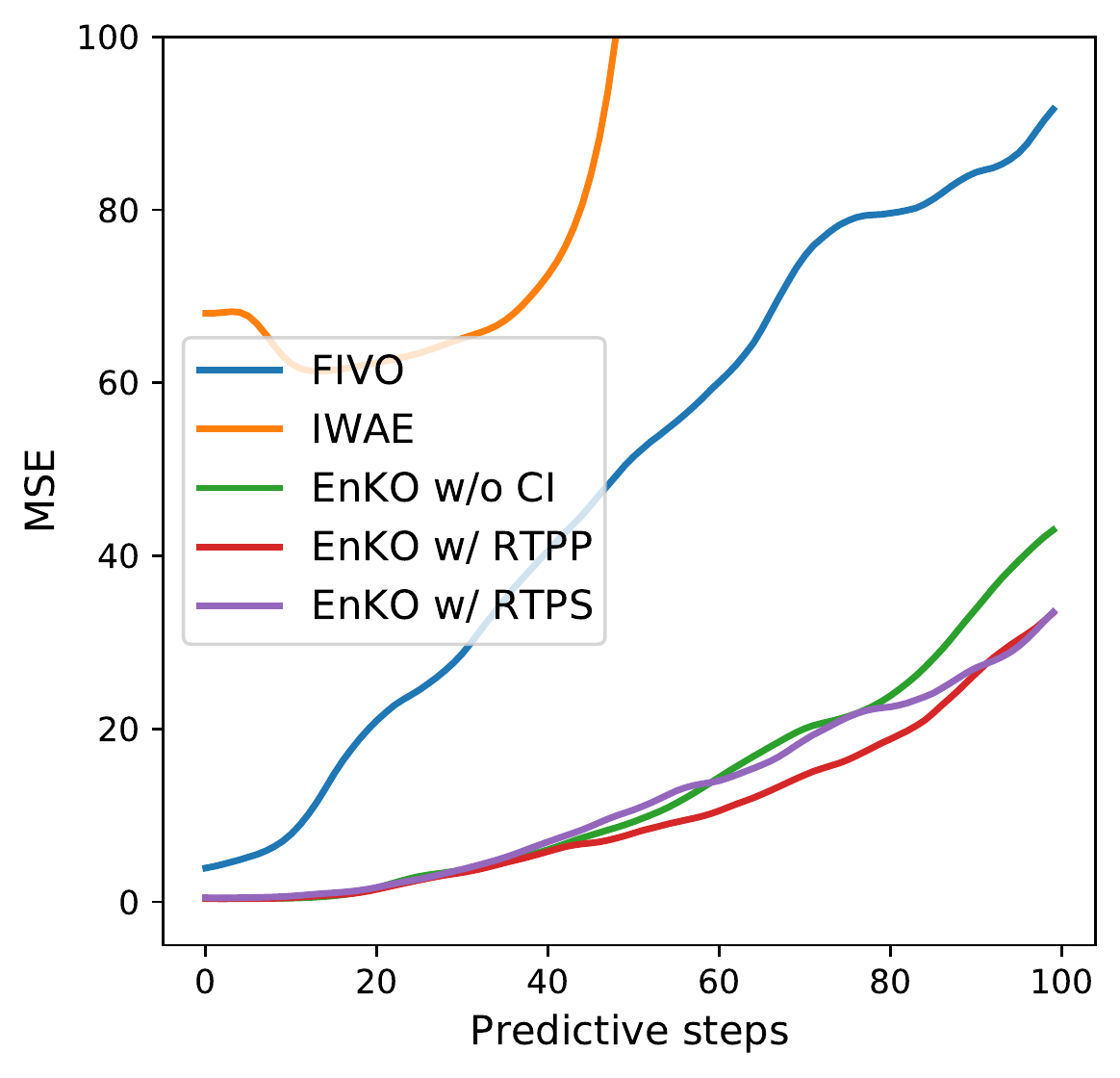}{Inferred latent trajectories and MSE for Lorenz model. (left) True trajectories on the first 10 test set. (center) Inferred latent trajectories on the 10 test set using EnKO. (right) MSE for various ensemble frameworks, including FIVO ({\it blue}), IWAE ({\it orange}), and EnKO without ({\it green}) and with covariance inflation (RTPP: {\it red}, RTPS: {\it purple}).}{fig:lorenz}{0.3}{0.3}{0.3}

We trained the SVO network \citep{mor19s} with EnKO, IWAE \citep{bur15}, and FIVO \citep{mad17} for two synthetic and two real-world benchmark datasets.
We used RTPP and RTPS as inflation methods in EnKO and set an inflation factor by grid search from \{0.1, 0.2, 0.3\}.
We implemented our experiments in PyTorch, and neural network parameters were optimized using the Adam optimizer \citep{kin14} with a learning rate of 0.001.
The settings of the other hyper-parameters and the details of the grid search are described in the supplementary material.
All experiments were performed on three random seeds, and performance metrics were averaged over the seeds.
The codes for the experiments are available at our GitHub page \url{https://github.com/ZoneMS/EnKO}.

\subsection{FitzHugh-Nagumo Model}
The FitzHugh-Nagumo model (FHN model) is a two-dimensional simplification of the Hodgkin-Huxley model, which models the spiking activity of neurons.
The model is represented by
\begin{align}
\dot V&=V-\frac{V^3}{3}-W+I_{\mathrm{ext}},\\
\dot W&=a(bV+d-cW),
\end{align}
where $V$ and $W$ represent the membrane potential and a recovery variable, respectively.
In our experiments, we set $I_{\mathrm{ext}}=0$, $a=0.7$, $b=0.8$, $c=0.08$, and $d=0$.
The initial states were uniformly sampled from $[-3,3]^2$ to generate 400 samples using 200 for training, 40 for validation, and 160 for testing.
A synthetic one-dimensional observation $x_t$ was sampled from $N(V_t,0.1^2)$.
We set the latent dimension of 2 and the number of particles 16.

The inferred latent dynamics and MSE are shown in Figure \ref{fig:fhn}.
The left panel displays the original dynamics and trajectories on the first 10 test set.
The center panel displays the inferred latent dynamics and inferred trajectories on the 10 test set using EnKO to perform dimensionality expansion.
The initial points located inside and outside the limit cycle and the order of the trajectories from the inside in the original system are equivalent to the reconstructed system.
This equivalence means that the reconstructed dynamics are topologically equivalent to the original dynamics.
The right panel shows the MSE comparison among ensemble frameworks.
The EnKO outperforms the previous methods, especially for the long predictive horizon.

\subsection{Lorenz Model}
The Lorenz model is a system of three ordinary differential equations, which was originally developed to simulate atmospheric convection.
The model is described by
\begin{align}
\frac{\diff x}{\diff t}&=\sigma (y-x),\\
\frac{\diff y}{\diff t}&=x(\rho-z)-y,\\
\frac{\diff z}{\diff t}&=xy-\beta z,
\end{align}
where $x$, $y$, and $z$ are the variables of this system.
In our experiments, we set $\sigma=10$, $\rho=28$, and $\beta=8/3$.
These parameter settings are known as chaos parameters, which lead to the Lorenz attractor.
The initial states were randomly chosen from $[-10,10]^3$ to generate 100 samples using 66 for training, 17 for validation, and 17 for testing.
A synthetic observation $\boldsymbol{x}_t$ was generated from $N\left(\begin{pmatrix}x_t&y_t&z_t\end{pmatrix}^T,0.1^2\times I_3\right)$.
We set the latent dimension of 3 and the number of particles 16.

The inferred latent dynamics and MSE are shown in Figure \ref{fig:lorenz}.
The left panel displays the original trajectories on the first 10 test set.
The center panel displays the inferred latent trajectories on the 10 test set using EnKO.
The order of the trajectories from inside the double-scroll attractor in the original system is equivalent to the reconstructed system.
This equivalence means that the reconstructed system is topologically equivalent to the original system.
The right panel shows the MSE comparison among ensemble frameworks.
The EnKO performs better prediction than the previous methods.

\subsection{CMU Walking Data}
\figimages{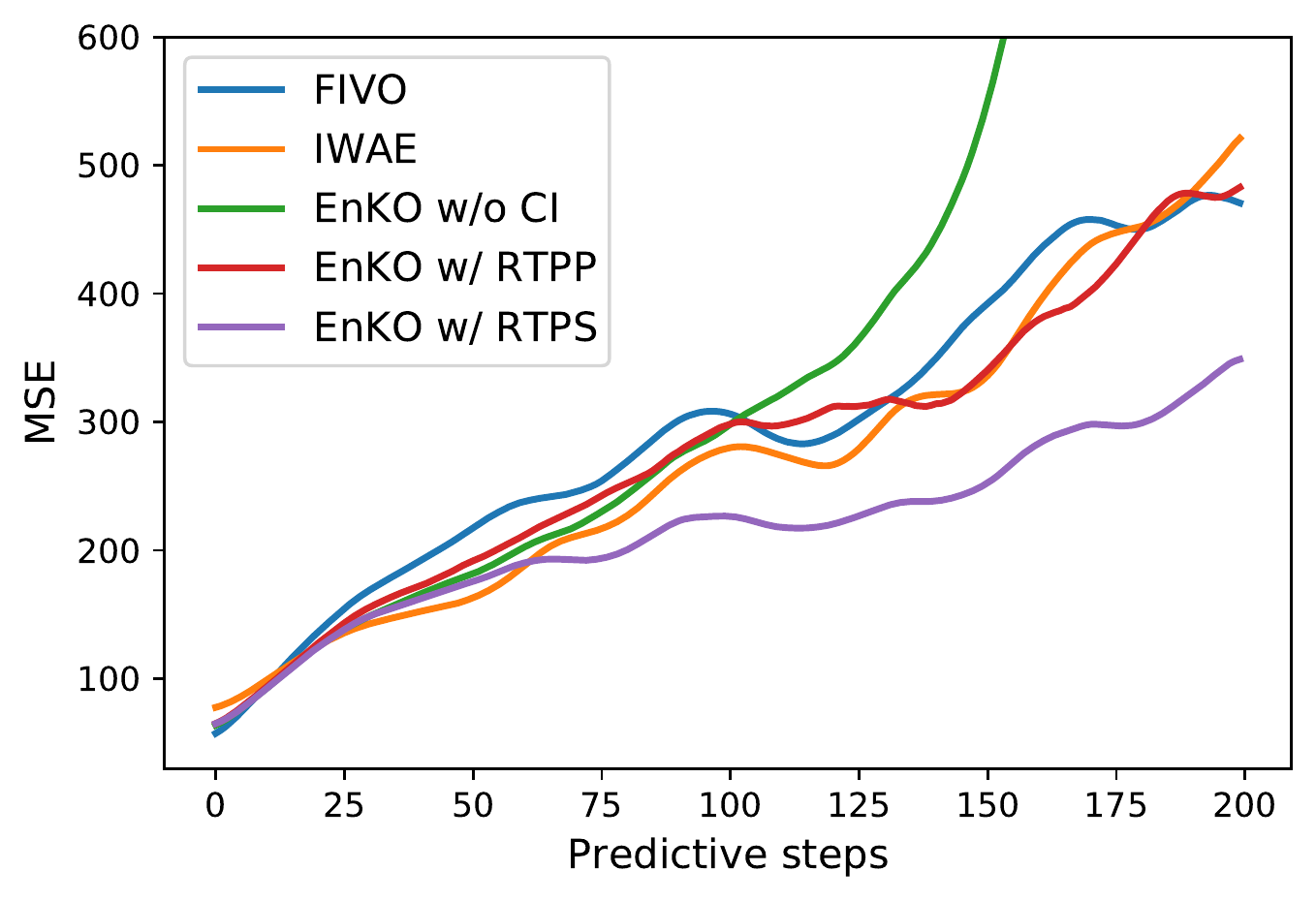}{MSE for walking dataset.}{fig:mocap_mse}{0.8}{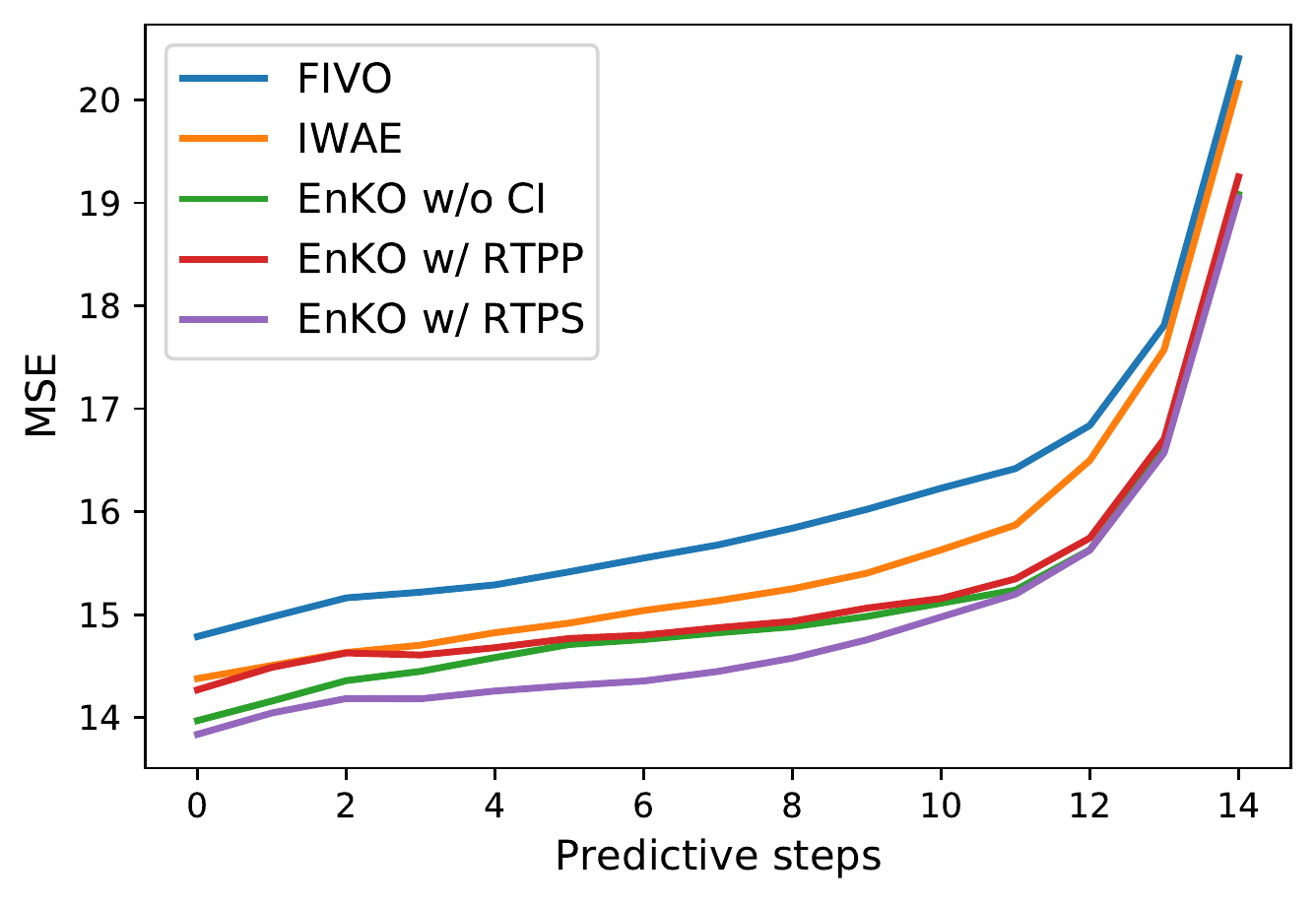}{MSE for rotating MNIST dataset.}{fig:rmnist_mse}{0.8}
To demonstrate that the proposed method can capture latent structures from real-world noisy observations, we experiment on a dataset extracting from CMU motion capture library.
We used the 23 walking sequences of subject 35 \citep{gan15}, which is partitioned into 16 for training, 3 for validation, and 4 for testing.
We followed the preprocessing procedure of \citep{wan08}, after which we were left with 47-dimensional joint angle measurements and 3-dimensional global velocities.
We set the latent dimension of 2 and the number of particles 128.

The predictied MSE are shown in Figure \ref{fig:mocap_mse}.
While the error of EnKO without inflation methods diverges, EnKO with RTPS outperforms the other methods.
This is probably because the EnKF tends to underestimate the state covariance when the ratio of the observed dimension to the number of particles is large.
The predicted reconstructions from the proposed method are shown in Appendix \ref{sec:aexp}.

\subsection{Rotating MNIST Dataset}
As a second showcase, we used a rotating MNIST dataset, consisting of rotated images of handwritten ``3" digits \citep{cas18}.
We used all of the rotation angles and divided the dataset into 360 for training, 40 for validation, and 600 for testing.
We used the hierarchical structure as presented in subsection \ref{ssec:high} and set the auxiliary dimension of 8, the latent dimension of 2, and the number of particles 32.

The predicted MSE are shown in Figure \ref{fig:rmnist_mse}.
The EnKO method provides lower errors than that of the FIVO and IWAE.
The predicted images from the methods are shown in Appendix \ref{sec:aexp}.

\section{Conclusion}
\label{sec:conclu}
We have introduced {\it Ensemble Kalman Variational Objective} (EnKO) to improve model inference by combining with sequential VAEs.
The proposed method uses the EnKF to infer latent variables and has three advantages over the previous SMC-based methods: particle diversity, the low variance of the gradient estimator, and the unbiased gradient estimators.

By these advantages, the proposed method outperforms the previous methods in terms of predictive ability for two synthetic and two real-world datasets.
The inferred 2-dimensional latent dynamics from 1-dimensional observations of the Fitz-Hugh Nagumo model are topologically equivalently to the original 2-dimensional dynamics.
The inferred latent trajectories for synthetic Lorenz data could reconstruct the double-scroll attractor.

It would be prudent and interesting for future work to focus on applications to an inference of stochastic differential equation (SDE).
Since the proposed framework is not constrained on a discrete-time state transition, we plan to start with ODE inference such as NeuralODE \citep{che18} and ODE2VAE \citep{yil19} and then extend to SDE.



\appendix
\section{Proof}
\label{sec:proof}

\setcounter{theorem}{0}
\begin{theorem}
A statistics
\begin{equation}
\hat p_N(\boldsymbol{x}_{1:T})=\frac{1}{N}\sum_{i=1}^N\prod_{t=1}^T\frac{f_{\boldsymbol{\theta}}(\boldsymbol{z}_t^{(i)}|\boldsymbol{z}^{(i)}_{1:t-1})g_{\boldsymbol{\theta}}(\boldsymbol{x}_t|\boldsymbol{z}_t^{(i)})}{q_{\boldsymbol{\varphi}}(\boldsymbol{z}_t^{(i)}|\boldsymbol{x}_{1:T},\boldsymbol{z}_{1:t-1}^{f,(i)})}
\end{equation}
is an approximately unbiased estimator of the marginal likelihood $p(\boldsymbol{x}_{1:T})$.
If the emission distribution $g_{\boldsymbol{\theta}}(\boldsymbol{x}_t|\boldsymbol{z}_t)$ and variational distribution $q_{\boldsymbol{\varphi}}(\boldsymbol{z}_t|\boldsymbol{z}_{1:t-1}^{f},\boldsymbol{x}_{1:T})$ are Gaussian, the $\hat p_N(\boldsymbol{x}_{1:T})$ is an unbiased estimator of the marginal likelihood.
\end{theorem}
EnKO applies the EnKF to the probabilistic time-series model (PTSM)
\begin{equation}
Q_{\boldsymbol{\theta},\boldsymbol{\varphi}}(\boldsymbol{x}_{1:T},\boldsymbol{z}_{1:T})=q_{\boldsymbol{\varphi}}(\boldsymbol{z}_1)\prod_{t=2}^Tq_{\boldsymbol{\varphi}}(\boldsymbol{z}_t|\boldsymbol{z}_{1:t-1},\boldsymbol{x}_{1:T})\prod_{t=1}^Tg_{\boldsymbol{\theta}}(\boldsymbol{x}_t|\boldsymbol{z}_t).
\end{equation}
With slightly abuse of notation, we denote the predictive distribution at time t by $Q(\boldsymbol{z}_t|\boldsymbol{x}_{1:t-1})$ and the filtered distribution by $Q(\boldsymbol{z}_t|\boldsymbol{x}_{1:t})$ in this PTSM.
Since the particle $\boldsymbol{z}_t^{f,(i)}$ can be approximated as sampling from the filtered distribution $Q(\boldsymbol{z}_t|\boldsymbol{x}_{1:t})$, the particle $\boldsymbol{z}_t^{(i)}$ can be considered as sampling from predictive distribution $Q(\boldsymbol{z}_t|\boldsymbol{x}_{1:t-1})$.
The expectation of the estimator is
\begin{align*}
&\mathbb{E}_{Q_{\mathrm{EnKO}}(\boldsymbol{z}_{1:T}^{(1:N)},\boldsymbol{x}_{1:T}^{(1:N)}|X)}[\hat p_N(\boldsymbol{x}_{1:T})]\\
&=\frac{1}{N}\sum_{i=1}^N\int\prod_{t=1}^T\frac{f_{\boldsymbol{\theta}}(\boldsymbol{z}_t^{(i)}|\boldsymbol{z}^{(i)}_{1:t-1})g_{\boldsymbol{\theta}}(\boldsymbol{x}_t|\boldsymbol{z}_t^{(i)})}{q_{\boldsymbol{\varphi}}(\boldsymbol{z}_t^{(i)}|\boldsymbol{x}_{1:T},\boldsymbol{z}_{1:t-1}^{f,(i)})}\prod_{j=1}^Nq_{\boldsymbol{\varphi}}(\boldsymbol{z}_t^{(j)}|\boldsymbol{x}_{1:T},\boldsymbol{z}_{1:t-1}^{f,(j)})\diff\boldsymbol{z}_{1:T}^{(1:N)}\\
&\simeq\frac{1}{N}\sum_{i=1}^N\int\prod_{t=1}^T\frac{f_{\boldsymbol{\theta}}(\boldsymbol{z}_t^{(i)}|\boldsymbol{z}^{(i)}_{1:t-1})g_{\boldsymbol{\theta}}(\boldsymbol{x}_t|\boldsymbol{z}_t^{(i)})}{Q_{\boldsymbol{\theta},\boldsymbol{\varphi}}(\boldsymbol{z}_t^{(i)}|\boldsymbol{x}_{1:t-1})}\prod_{j=1}^NQ_{\boldsymbol{\theta},\boldsymbol{\varphi}}(\boldsymbol{z}_t^{(j)}|\boldsymbol{x}_{1:t-1})\diff\boldsymbol{z}_{1:T}^{(1:N)}\\
&=\frac{1}{N}\sum_{i=1}^N\int\prod_{t=1}^T\frac{f_{\boldsymbol{\theta}}(\boldsymbol{z}_t^{(i)}|\boldsymbol{z}^{(i)}_{1:t-1})g_{\boldsymbol{\theta}}(\boldsymbol{x}_t|\boldsymbol{z}_t^{(i)})}{Q_{\boldsymbol{\theta},\boldsymbol{\varphi}}(\boldsymbol{z}_t^{(i)}|\boldsymbol{x}_{1:t-1})}Q_{\boldsymbol{\theta},\boldsymbol{\varphi}}(\boldsymbol{z}_t^{(i)}|\boldsymbol{x}_{1:t-1})\diff\boldsymbol{z}_{1:T}^{(i)}\\
&=\frac{1}{N}\sum_{i=1}^N\int\prod_{t=1}^Tf_{\boldsymbol{\theta}}(\boldsymbol{z}_t^{(i)}|\boldsymbol{z}^{(i)}_{1:t-1})g_{\boldsymbol{\theta}}(\boldsymbol{x}_t|\boldsymbol{z}_t^{(i)})\diff\boldsymbol{z}_{1:T}^{(i)}\\
&=p(\boldsymbol{x}_{1:T}).
\end{align*}
It is noteworthy that in the second line, each particle $\boldsymbol{z}_t^{(i)}$ depends on the other particles through $\boldsymbol{z}_t^{f,(i)}$, while in the third line, the dependence between particles are broken by approximating the predictive distribution $Q(\boldsymbol{z}_t|\boldsymbol{x}_{1:t-1})$.

The approximation equation in the second line is an equality if $g_{\boldsymbol{\theta}}(\boldsymbol{x}_t|\boldsymbol{z}_t)$ and $q_{\boldsymbol{\varphi}}(\boldsymbol{z}_t|\boldsymbol{z}_{1:t-1},\boldsymbol{x}_{1:T})$ are Gaussian due to the general fact of EnKF, and $\hat p_N(\boldsymbol{x}_{1:T})$ is an unbiased estimator of the marginal likelihood $p(\boldsymbol{x}_{1:T})$.

\begin{corollary}
An objective function 
\begin{equation}
\mathcal{L}_{\mathrm{EnKO}}^N(\boldsymbol{\theta},\boldsymbol{\varphi},X)=\mathbb{E}_{Q_{\mathrm{EnKO}}(\boldsymbol{z}_{1:T}^{(1:N)},\boldsymbol{x}_{1:T}^{(1:N)}|X)}[\log\hat p_N(\boldsymbol{x}_{1:T})]
\end{equation}
is an approximately lower bound of the log marginal likelihood $\log p(\boldsymbol{x}_{1:T})$.
If the emission distribution $g_{\boldsymbol{\theta}}(\boldsymbol{x}_t|\boldsymbol{z}_t)$ and the variational distribution $q_{\boldsymbol{\varphi}}(\boldsymbol{z}_t|\boldsymbol{x}_{1:T},\boldsymbol{z}_{1:t-1}^{f})$ are Gaussian, the objective function is an unbiased estimator of the log marginal likelihood.
\end{corollary}

This is easily proved by Jensen's inequality and Theorem 1, i.e.,
\begin{align*}
\mathcal{L}_{\mathrm{EnKO}}^N(\boldsymbol{\theta},\boldsymbol{\varphi},X)&=\mathbb{E}_{Q_{\mathrm{EnKO}}(\boldsymbol{z}_{1:T}^{(1:N)},\boldsymbol{x}_{1:T}^{(1:N)}|X)}[\log\hat p_N(\boldsymbol{x}_{1:T})]\\
&\le\log\mathbb{E}_{Q_{\mathrm{EnKO}}(\boldsymbol{z}_{1:T}^{(1:N)},\boldsymbol{x}_{1:T}^{(1:N)}|X)}[\hat p_N(\boldsymbol{x}_{1:T})]\\
&\simeq\log p(\boldsymbol{x}_{1:T}).
\end{align*}

\section{Gradient Estimator}
\label{sec:grad}
The gradient of the objective function is
\begin{align*}
\nabla_{\boldsymbol{\theta},\boldsymbol{\varphi}}&\mathcal{L}_{\mathrm{EnKO}}(\boldsymbol{\theta},\boldsymbol{\varphi},X)\notag\\
&=\nabla_{\boldsymbol{\theta},\boldsymbol{\varphi}}\int Q_{\mathrm{EnKO}}(\boldsymbol{z}_{1:T}^{(1:N)},\boldsymbol{x}_{1:T-1}^{(1:N)})\log Z_{\mathrm{EnKO}}(\boldsymbol{z}_{1:T}^{(1:N)},\boldsymbol{x}_{1:T-1}^{(1:N)})\ \diff\boldsymbol{z}_{1:T}^{(1:N)}\diff\boldsymbol{x}_{1:T-1}^{(1:N)}\\
&=\nabla_{\boldsymbol{\theta},\boldsymbol{\varphi}}\int\left(\prod_{i=1}^Nq_{\boldsymbol{\varphi}}(\boldsymbol{z}_1^{(i)}|\boldsymbol{x}_{1:T})\right)\cdot\\
&\qquad\left(\prod_{t=2}^T\prod_{i=1}^Nq_{\boldsymbol{\varphi}}(\boldsymbol{z}_t^{(i)}|\boldsymbol{x}_{1:T},\mathrm{EnKF}_{\boldsymbol{\theta}}(\boldsymbol{z}_{t-1}^{(i)},\boldsymbol{z}_{t-1}^{(1:N)},\boldsymbol{x}_{t-1}^{(i)},\boldsymbol{x}_{t-1}^{(1:N)},\boldsymbol{x}_t))g_{\boldsymbol{\theta}}(\boldsymbol{x}_{t-1}^{(i)}|\boldsymbol{z}_{t-1}^{(i)})\right)\cdot\\
&\qquad\log Z_{\mathrm{EnKO}}(\boldsymbol{z}_{1:T}^{(1:N)},\boldsymbol{x}_{1:T}^{(1:N)})\diff\boldsymbol{z}_{1:T}^{(1:N)}\diff\boldsymbol{x}_{1:T-1}^{(1:N)},
\end{align*}
where $Z_{\mathrm{EnKO}}(\boldsymbol{z}_{1:T}^{(1:N)},\boldsymbol{x}_{1:T-1}^{(1:N)}):=\hat p_N(\boldsymbol{x}_{1:T})$ and $\mathrm{EnKF}_{\boldsymbol{\theta}}(\boldsymbol{z}_{t-1}^{(i)},\boldsymbol{z}_{t-1}^{(1:N)},\boldsymbol{x}_{t-1}^{(i)},\boldsymbol{x}_{t-1}^{(1:N)},\boldsymbol{x}_t)$ denote the deterministic process of the EnKF as described in Algorithm \ref{alg:enko}.
When reparametrized sampling is applied, including the filtering process, we obtain
\begin{align*}
\nabla_{\boldsymbol{\theta},\boldsymbol{\varphi}}&\mathcal{L}_{\text{EnKO}}(\boldsymbol{\theta},\boldsymbol{\varphi},X)\notag\\
&=\int\left(\prod_{t=1}^T\prod_{i=1}^Ns^q_t(\boldsymbol{\varepsilon}_t^{q,(i)})\right)\left(\prod_{t=1}^{T-1}\prod_{i=1}^Ns^g_t(\boldsymbol{\varepsilon}_t^{g,(i)})\right)\cdot\notag\\
&\qquad\nabla_{\boldsymbol{\theta},\boldsymbol{\varphi}}\log Z_{\text{EnKO}}(r^q(\boldsymbol{\varepsilon}_{1:T}^{q,(1:N)}),r^g(\boldsymbol{\varepsilon}_{1:T-1}^{g,(1:N)}))\diff\boldsymbol{\varepsilon}_{1:T}^{q,(1:N)}\diff\boldsymbol{\varepsilon}_{1:T-1}^{g,(1:N)}\\
&=\mathbb{E}_{S_{\text{EnKO}}(\boldsymbol{\varepsilon}_{1:T}^{q,(1:N)},\boldsymbol{\varepsilon}_{1:T-1}^{g,(1:N)})}\left[\nabla_{\boldsymbol{\theta},\boldsymbol{\varphi}}\log Z_{\text{EnKO}}(r^q(\boldsymbol{\varepsilon}_{1:T}^{q,(1:N)}),r^g(\boldsymbol{\varepsilon}_{1:T-1}^{g,(1:N)}))\right],
\end{align*}
where $\boldsymbol{\varepsilon}_t^{q,(i)}$ and $\boldsymbol{\varepsilon}_t^{g,(i)}$ follows the distribution $s^q_t(\boldsymbol{\varepsilon}_t^{q,(i)})$ and $s^g_t(\boldsymbol{\varepsilon}_t^{g,(i)})$, respectively, and $r^q$ and $r^g$ represent the reparametrization transform from $\boldsymbol{\varepsilon}_{1:T}^{q,(1:N)}$ to $\boldsymbol{z}_{1:T}^{(1:N)}$ and from $\boldsymbol{\varepsilon}_{1:T}^{g,(1:N)}$ to $\boldsymbol{x}_{1:T}^{(1:N)}$, respectively.

\section{Experiment for Variance of Gradient Estimators}
The low variance of the gradient estimator means that stable learning is possible.
In this section, we compute the variance of the gradient estimators of IWAE, FIVO, and EnKO for two toy examples and compare the results among the methods.

\subsection{Linear Gaussian State Space Model}
\figimage{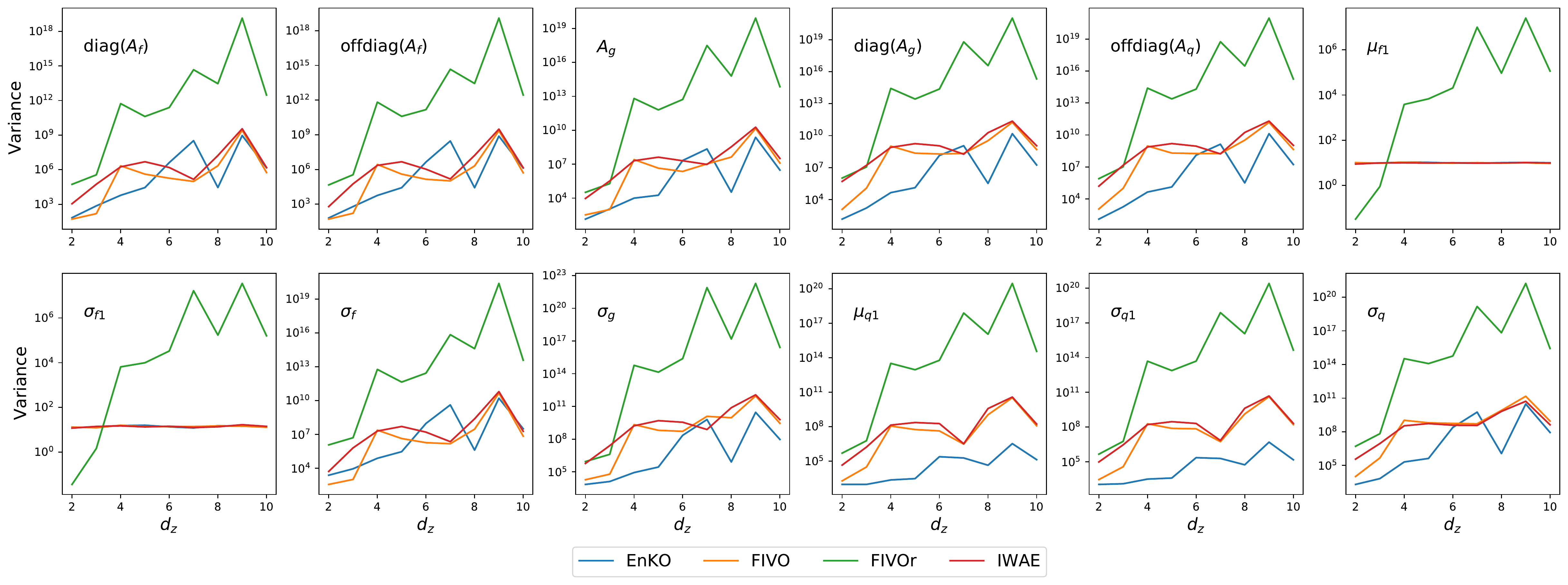}{Variance of the gradient estimates of EnKO, FIVO, FIVOr, and IWAE versus latent dimension $d_z$, where FIVOr means FIVO with resampling gradient term. For a vector parameter $\boldsymbol{v}$, the variance is computed for each elements and then averaged over the elements. For a matrix parameter $A$, $\mathrm{diag}$(A) and $\mathrm{offdiag}(A)$ represent average of the variance of the diagonal elements and the off-diagonal elements, respectively.}{fig:lgdz}{1}
\figimage{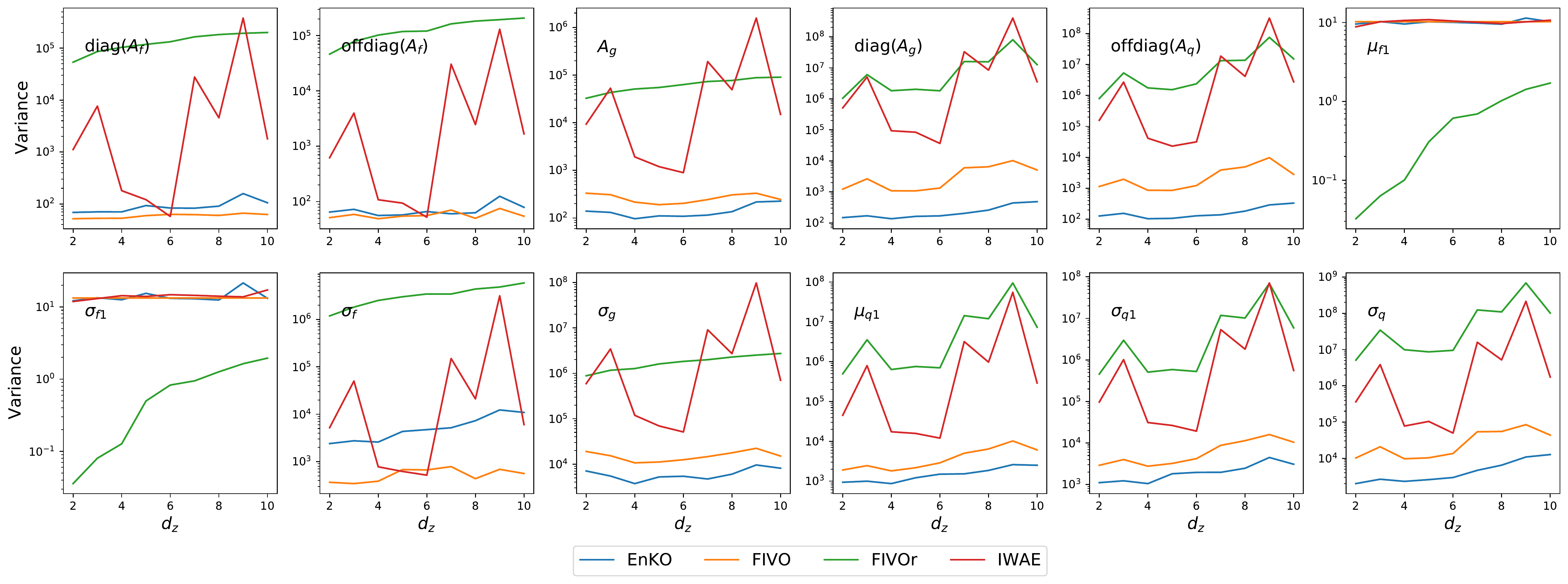}{Variance of the gradient estimates of EnKO, FIVO, FIVOr, and IWAE versus observed dimension $d_x$, where FIVOr means FIVO with resampling gradient term.}{fig:lgdx}{1}
\figimage{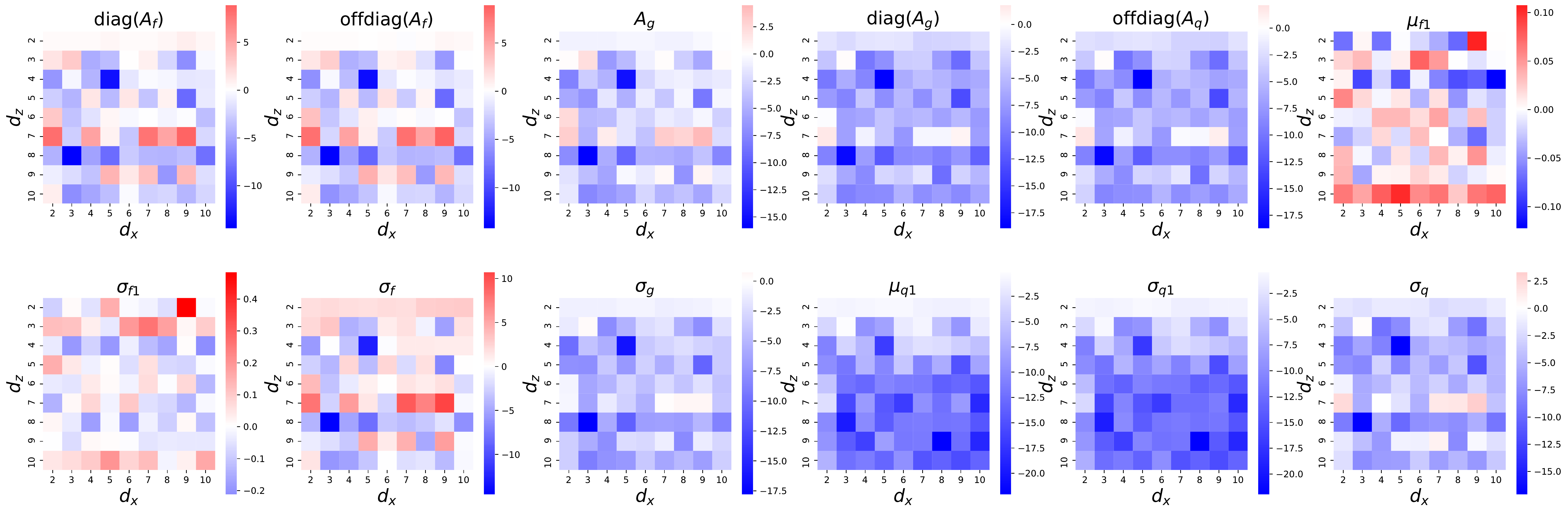}{Log relative variance of EnKO to FIVO $\log\{V_{\mathrm{EnKO}}[\nabla\mathcal{L}]/V_{\mathrm{FIVO}}[\nabla\mathcal{L}]\}$. Positive values ({\it red}) represent the variance of EnKO is greater than FIVO and the opposite is true for negative values ({\it blue}).}{fig:lgpm}{1}

The first example uses the following variational distribution and SSM
\begin{subequations}
\begin{align}
\ \boldsymbol{z}_1&\sim N(\boldsymbol{\mu}_{q1},\boldsymbol{\sigma}_{q1}^2),\\
\boldsymbol{z}_t&\sim q_{\boldsymbol{\varphi}}(\boldsymbol{z}_t|\boldsymbol{z}_{t-1})=N(A_q\boldsymbol{z}_{t-1},\boldsymbol{\sigma}_q^2),\\
p_{\boldsymbol{\theta}}(Z,X)&=f_{\boldsymbol{\theta}}(\boldsymbol{z}_1)\prod_{t=2}^Tf_{\boldsymbol{\theta}}(\boldsymbol{z}_t|\boldsymbol{z}_{t-1})\prod_{t=1}^Tg_{\boldsymbol{\theta}}(\boldsymbol{x}_t|\boldsymbol{z}_t),\\
f_{\boldsymbol{\theta}}(\boldsymbol{z}_1)&=N(\boldsymbol{\mu}_{f1},\boldsymbol{\sigma}_{f1}^2),\\
f_{\boldsymbol{\theta}}(\boldsymbol{z}_t|\boldsymbol{z}_{t-1})&=N(A_f\boldsymbol{z}_{t-1},\boldsymbol{\sigma}_f^2),\\
g_{\boldsymbol{\theta}}(\boldsymbol{x}_t|\boldsymbol{z}_t)&=N(A_g\boldsymbol{z}_t,\boldsymbol{\sigma}_g^2),
\end{align}
\end{subequations}
where $\boldsymbol{\theta}=\{\boldsymbol{\mu}_{f1},A_{f},A_{g},\boldsymbol{\sigma}_{f1},\boldsymbol{\sigma}_{f},\boldsymbol{\sigma}_{g}\}$ and $\boldsymbol{\varphi}=\{\boldsymbol{\mu}_{q1},A_{q},\boldsymbol{\sigma}_{q1},\boldsymbol{\sigma}_{q}\}$.
We set $T=100$, number of particles 16, batch size 10, number of simulations 100, and $d_x,d_z\in\{2,3,\cdots,10\}$.
We evaluated the gradient estimators at $\boldsymbol{\mu}_{f1}=\boldsymbol{\mu}_{q1}=\boldsymbol{0}$, $\boldsymbol{\sigma}_{q1}=\boldsymbol{\sigma}_{f1}=0.1\times\boldsymbol{1}$, $\boldsymbol{\sigma}_{q}=\boldsymbol{\sigma}_{f}=\boldsymbol{\sigma}_g=0.01\times\boldsymbol{1}$, $A_q=I_{d_z}+U_q$, $A_f=I_{d_z}+U_f$, $(U_q)_{ij},(U_f)_{ij}\sim U(-0.05,0.05)$, and $(A_g)_{ij}\sim U(-0.5,0.5)$.

Figure \ref{fig:lgdz} and \ref{fig:lgdx} show variance of the gradient estimates of EnKO, FIVO, FIVOr, and IWAE versus latent dimension $d_z$ and observed dimension $d_x$, respectively, where FIVOr means FIVO with resampling gradient term.
Figure \ref{fig:lgpm} shows log relative variance of EnKO to FIVO $\log\{V_{\mathrm{EnKO}}[\nabla\mathcal{L}]/V_{\mathrm{FIVO}}[\nabla\mathcal{L}]\}$.
For a vector parameter $\boldsymbol{v}\in\mathbb{R}^d$, the variance is computed for each element and then averaged over the elements
\begin{equation}
V[\boldsymbol{v}]=\frac{1}{d}\sum_{i=1}^dV\left[\frac{\partial\mathcal{L}}{\partial v_i}\right].
\end{equation}
For a square matrix parameter $A\in\mathbb{R}^{d\times d}$, $\mathrm{diag}$(A) and $\mathrm{offdiag}(A)$ represent average of the variance of the diagonal elements and the off-diagonal elements, respectively, i.e.,
\begin{align}
V[\mathrm{diag}(A)]&=\frac{1}{d}\sum_{i=1}^dV\left[\frac{\partial\mathcal{L}}{\partial A_{ii}}\right],\\
V[\mathrm{offdiag}(A)]&=\frac{1}{d(d-1)}\sum_{i\neq j, i,j\in\mathbb{N}_d}V\left[\frac{\partial\mathcal{L}}{\partial A_{ij}}\right],
\end{align}
For a rectangular matrix $A\in\mathbb{R}^{d_1\times d_2}$, the variance is computed same as a vector parameter
\begin{equation}
V[A]=\frac{1}{d^2}\sum_{i=1}^{d_1}\sum_{j=1}^{d_2}V\left[\frac{\partial\mathcal{L}}{\partial A_{ij}}\right].
\end{equation}

From Figure \ref{fig:lgdz} and \ref{fig:lgdx}, FIVOr (FIVO with resampling gradient term) has higher variance than the other methods, and EnKO has a relatively low variance for all parameters.
In particular, the variance of FIVOr is rapidly increased when $d_z$ is increased.
When $d_x=2$ is fixed and $d_z$ increases, FIVO and IWAE are competitive, but when $d_z=2$ is fixed and $d_x$ increases, the variance of IWAE shifts more than that of FIVO.
From Figure \ref{fig:lgpm}, the variance of gradient estimates of EnKO is generally lower than that of FIVO, especially for parameters regarding $g$ and $q$.

\subsection{Nonlinear Non-Gaussian State Space Model}
\figimage{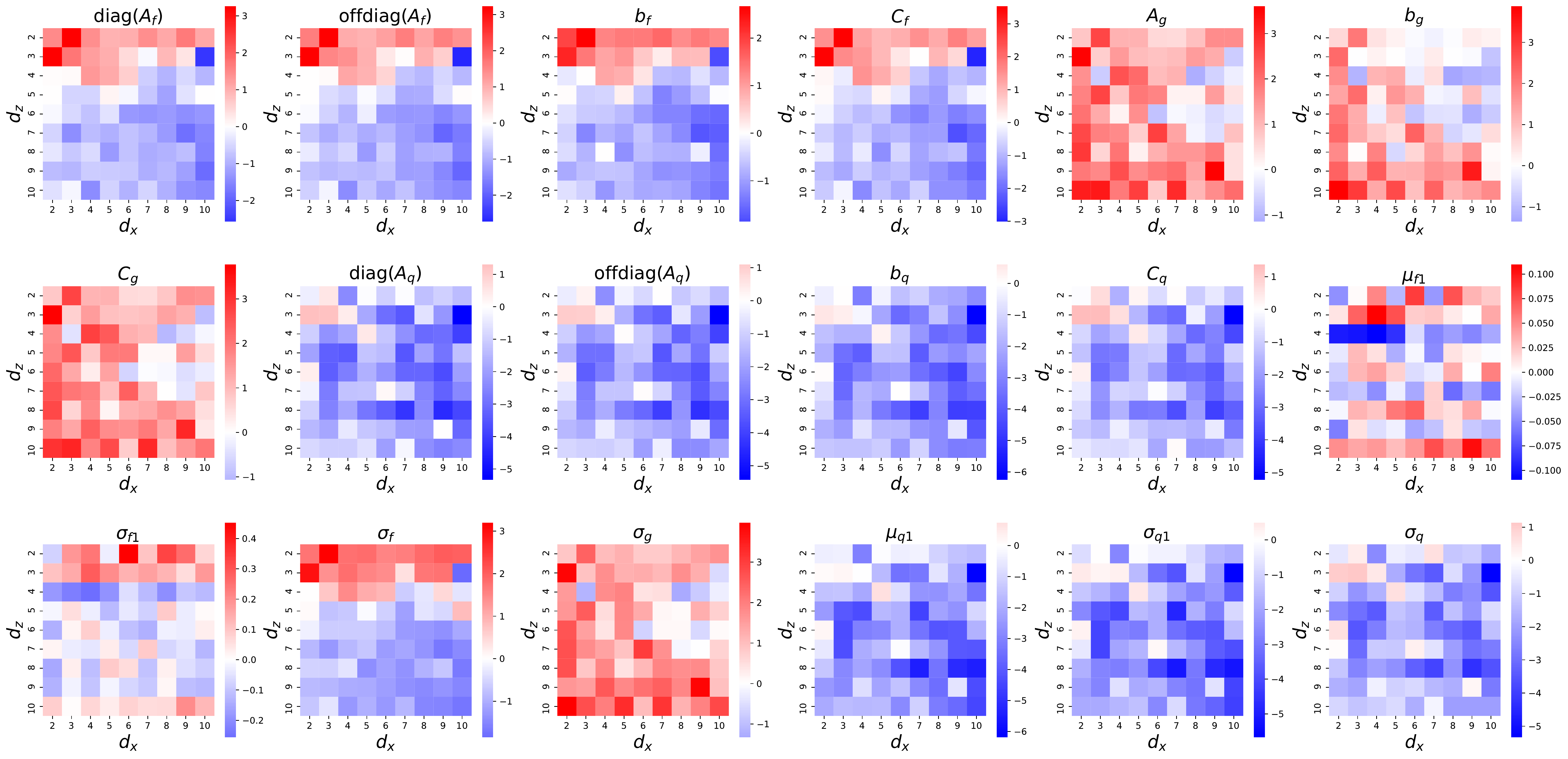}{Log relative variance of EnKO to FIVO $\log\{V_{\mathrm{EnKO}}[\nabla\mathcal{L}]/V_{\mathrm{FIVO}}[\nabla\mathcal{L}]\}$. Positive values ({\it red}) represent the variance of EnKO is greater than FIVO and the opposite is true for negative values ({\it blue}).}{fig:nlgpm}{1}

The second example uses the following variational distribution and SSM
\begin{subequations}
\begin{align}
\ \boldsymbol{z}_1&\sim N(\boldsymbol{\mu}_{q1},\boldsymbol{\sigma}_{q1}^2),\\
\boldsymbol{z}_t&\sim q_{\boldsymbol{\varphi}}(\boldsymbol{z}_t|\boldsymbol{z}_{t-1})=N(\tanh(C_q\mathrm{flatten}(\boldsymbol{z}_{t-1}\boldsymbol{z}_{t-1}^T)+A_q\boldsymbol{z}_{t-1}+\boldsymbol{b}_q),\boldsymbol{\sigma}_q^2),\\
p_{\boldsymbol{\theta}}(Z,X)&=f_{\boldsymbol{\theta}}(\boldsymbol{z}_1)\prod_{t=2}^Tf_{\boldsymbol{\theta}}(\boldsymbol{z}_t|\boldsymbol{z}_{t-1})\prod_{t=1}^Tg_{\boldsymbol{\theta}}(\boldsymbol{x}_t|\boldsymbol{z}_t),\\
f_{\boldsymbol{\theta}}(\boldsymbol{z}_1)&=N(\boldsymbol{\mu}_{f1},\boldsymbol{\sigma}_{f1}^2),\\
f_{\boldsymbol{\theta}}(\boldsymbol{z}_t|\boldsymbol{z}_{t-1})&=\mathrm{Student}_5(\tanh(C_f\mathrm{flatten}(\boldsymbol{z}_{t-1}\boldsymbol{z}_{t-1}^T)+A_f\boldsymbol{z}_{t-1}+\boldsymbol{b}_f),\boldsymbol{\sigma}_f),\\
g_{\boldsymbol{\theta}}(\boldsymbol{x}_t|\boldsymbol{z}_t)&=\mathrm{Student}_5(\tanh(C_g\mathrm{flatten}(\boldsymbol{z}_{t}\boldsymbol{z}_{t}^T)+A_g\boldsymbol{z}_{t}+\boldsymbol{b}_g),\boldsymbol{\sigma}_g),
\end{align}
\end{subequations}
where $\theta=\{\boldsymbol{\mu}_{f1},\sigma_{f1},C_f,A_f,\boldsymbol{b}_f,\boldsymbol{\sigma}_f,C_g,A_g,\boldsymbol{b}_g,\boldsymbol{\sigma}_g\}$, $\varphi=\{\boldsymbol{\mu}_{q1},\boldsymbol{\sigma}_{q1},C_q,A_q,\boldsymbol{b}_q,\boldsymbol{\sigma}_q\}$.
$\mathrm{Student}_\nu(\rho,\sigma)$ represent student-t distribution with degree of freedom $\nu$, location parameter $\rho$, and scale parameter $\sigma$.
For a matrix $A\in\mathbb{R}^{d\times d}$, $\mathrm{flatten}(A)=(a_{11},a_{12},\cdots,a_{dd})^T\in\mathbb{R}^{d^2}$.

Figure \ref{fig:nlgpm} shows log relative variance of EnKO to FIVO $\log\{V_{\mathrm{EnKO}}[\nabla\mathcal{L}]/V_{\mathrm{FIVO}}[\nabla\mathcal{L}]\}$.
EnKO's gradient estimates for the parameters regarding $f$ and $q$ have lower variance than FIVO, but the opposite is true for $g$.
This indicates that while EnKO can stabilize learning for variational and transition parameters, FIVO has the ability for generative parameters.
Since the learning of transition parameters is often unstable in time-series models, EnKO which can stabilize the learning transition parameters is a more effective method.

\section{Experiment Details}
\label{sec:expd}
\figimage{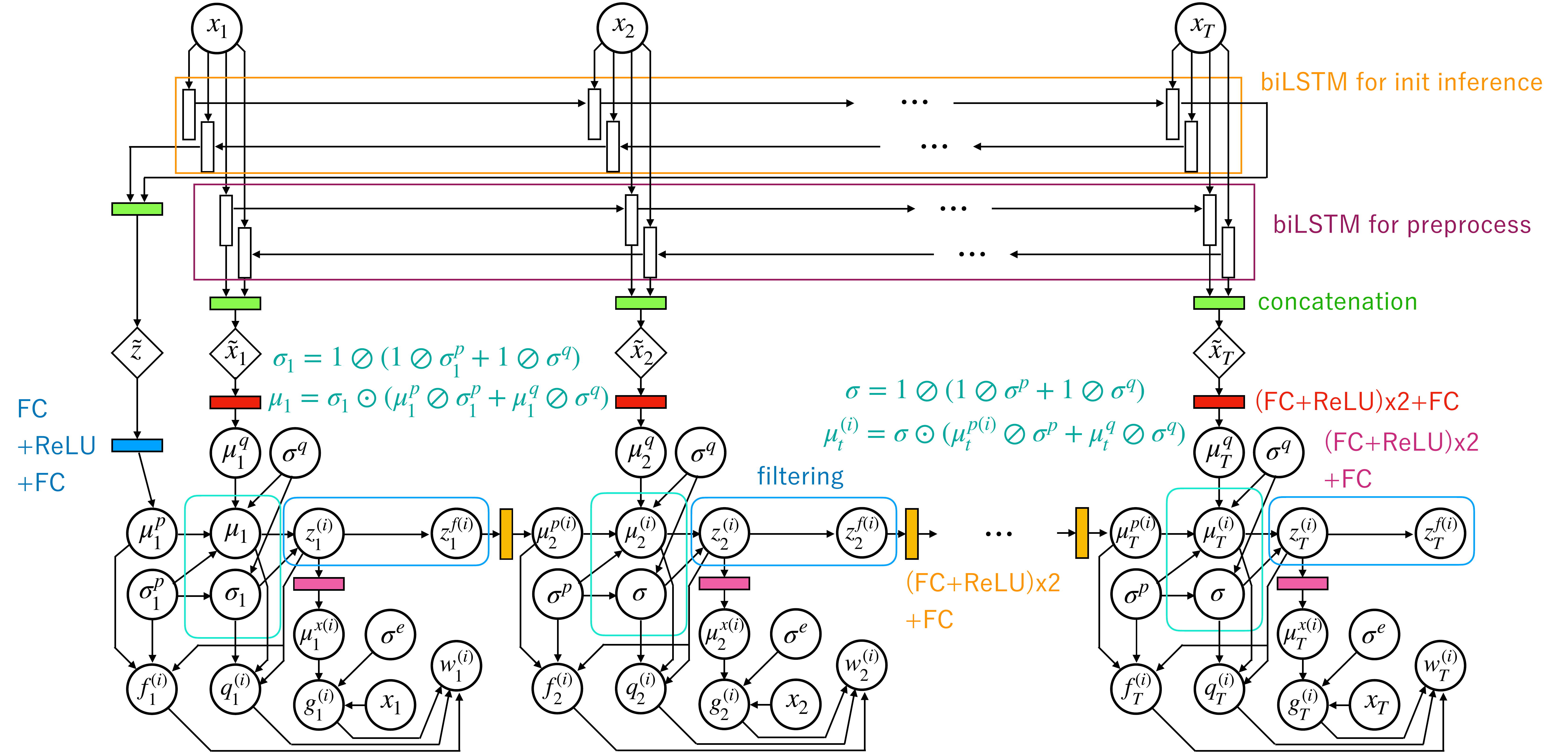}{The network architecture of SVO. FC and biLSTM stand for fully connected layer and bidirectional LSTM, respectively. A filtering step ({\it blue}) is different by ensemble systems.}{fig:svo}{0.95}
\figimage{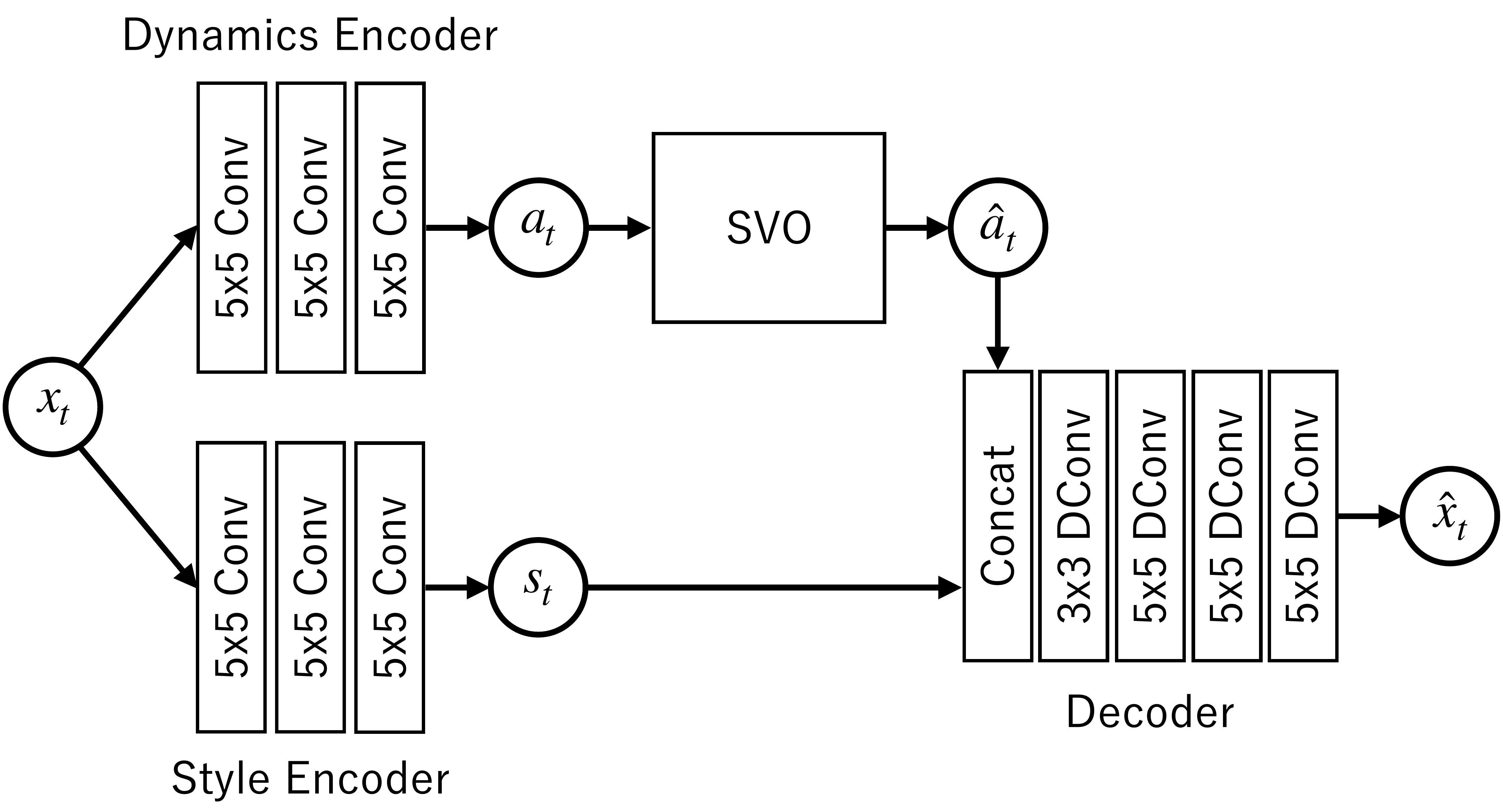}{The network architecture of outer VAE for rotating MNIST dataset. Conv and DConv stand for convolution layer and deconvolution layer, respectively. The encoder consists of style convolution block and dynamics convolution block. While the style convolution block extracts style (static) information such as shape and edge of the handwritten data, the dynamics convolution block extracts dynamics information such as rotation dynamics.}{fig:styleconv}{0.8}

\begin{table}[t]
\caption{Chosen hyperparamters and configurations}
\label{tab:hyper}
\vskip 0.05in
\begin{center}
\begin{small}
\begin{tabular}{p{0.25\textwidth}*{4}{|p{0.1\textwidth}}}
\toprule
\multirow{2}{*}{Parameters} & \multicolumn{4}{c}{Dataset} \\
& FHN & Lorenz & Walking & RMNIST \\
\midrule
\# Neural units per layer & \{16, {\bf 32}\} & \{16, {\bf 32}\} & \{50, {\bf 100}\} & \{25, {\bf 50}\} \\
Latent dimension & 2 & 3 & 6 & 2 \\
Batch size & 20 & 6 & 4 & 40 \\
Training epochs & 2000 & 2000 & 2000 & 3000 \\
Scaling & abs. div. & abs. div. & - & - \\
\bottomrule
\end{tabular}
\end{small}
\end{center}
\vskip -0.1in
\end{table}

Figure \ref{fig:svo} shows the network architecture of SVO \citep{mor19s} as coded in the authors' GitHub page \url{https://github.com/amoretti86/PSVO}.
Although they trained SVO with the FIVO system, we expanded this with EnKO and IWAE systems.
The chosen hyperparameters of SVO are the same as their experiments.
Figure \ref{fig:styleconv} shows the network architecture of outer VAE for rotating MNIST dataset.
The encoder consists of style convolution block and dynamics convolution block.
While the style convolution block extracts style (static) information such as shape and edge of the handwritten data, the dynamics convolution block extracts dynamics information such as rotation dynamics.
The chosen hyperparameters of convolution layers are the same as experiments for the dataset in \cite{yil19}.
The dimension of $\boldsymbol{a}_t$ and $\boldsymbol{s}_t$ were set to 2 and 6, respectively.
Table \ref{tab:hyper} summarize other hyperparameters and configurations used for generating the results reported in Section \ref{sec:exp}.
In this table, the boldfaces are the chosen hyperparameters through our experiments.
The more detailed conditions and our experiments are described in our GitHub page \url{https://github.com/ZoneMS/EnKO}.

We used NVIDIA Tesla V100 GPUs, CUDA 10.2, and PyTorch 1.6.0 for our experiments.
Each experiment for the FHN model, the Lorenz model, and the walking dataset spent around one day, and each experiment for the rotating MNIST dataset spent around two days.

\section{Additional Results}
\label{sec:aexp}
\figimagethree{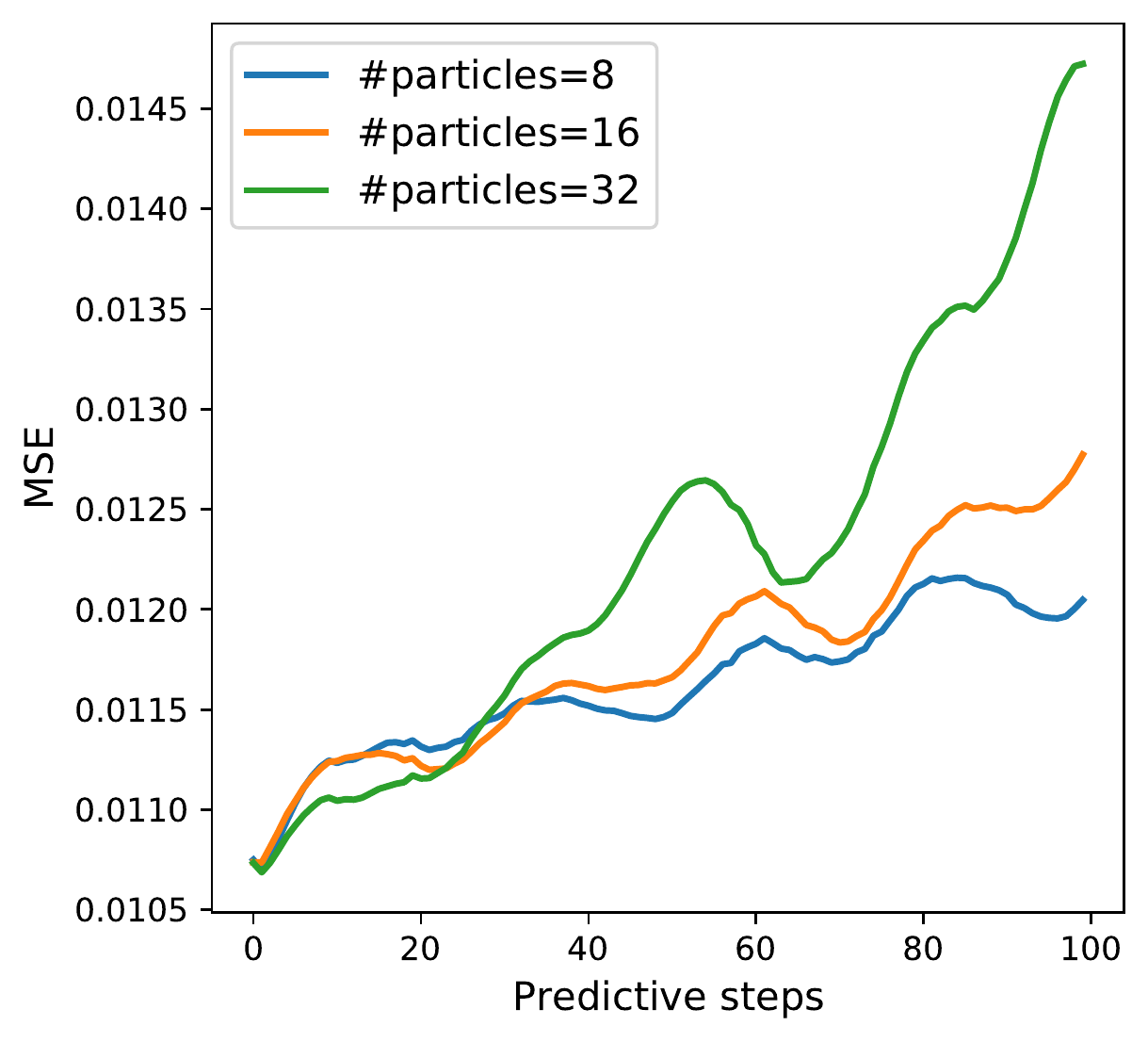}{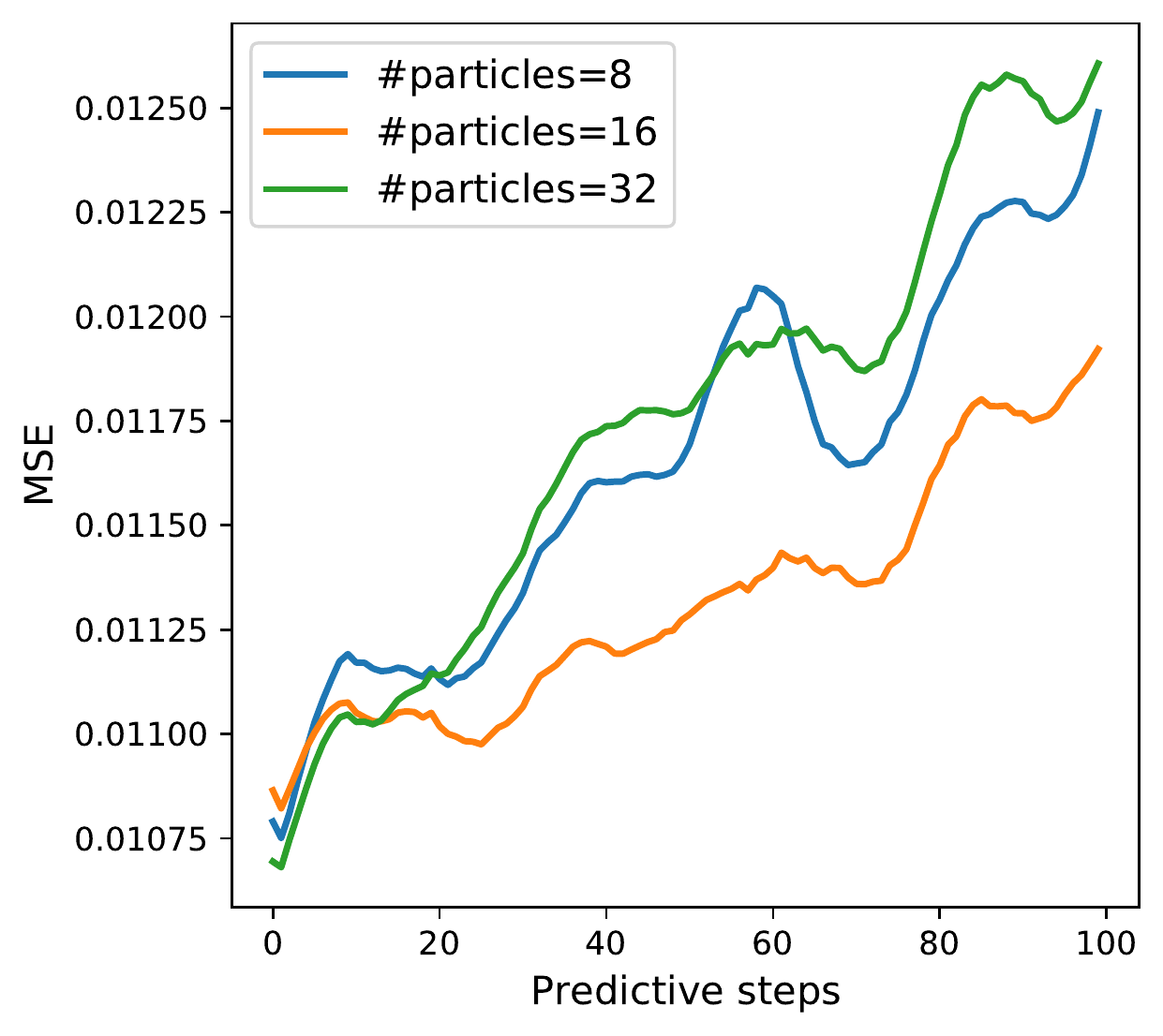}{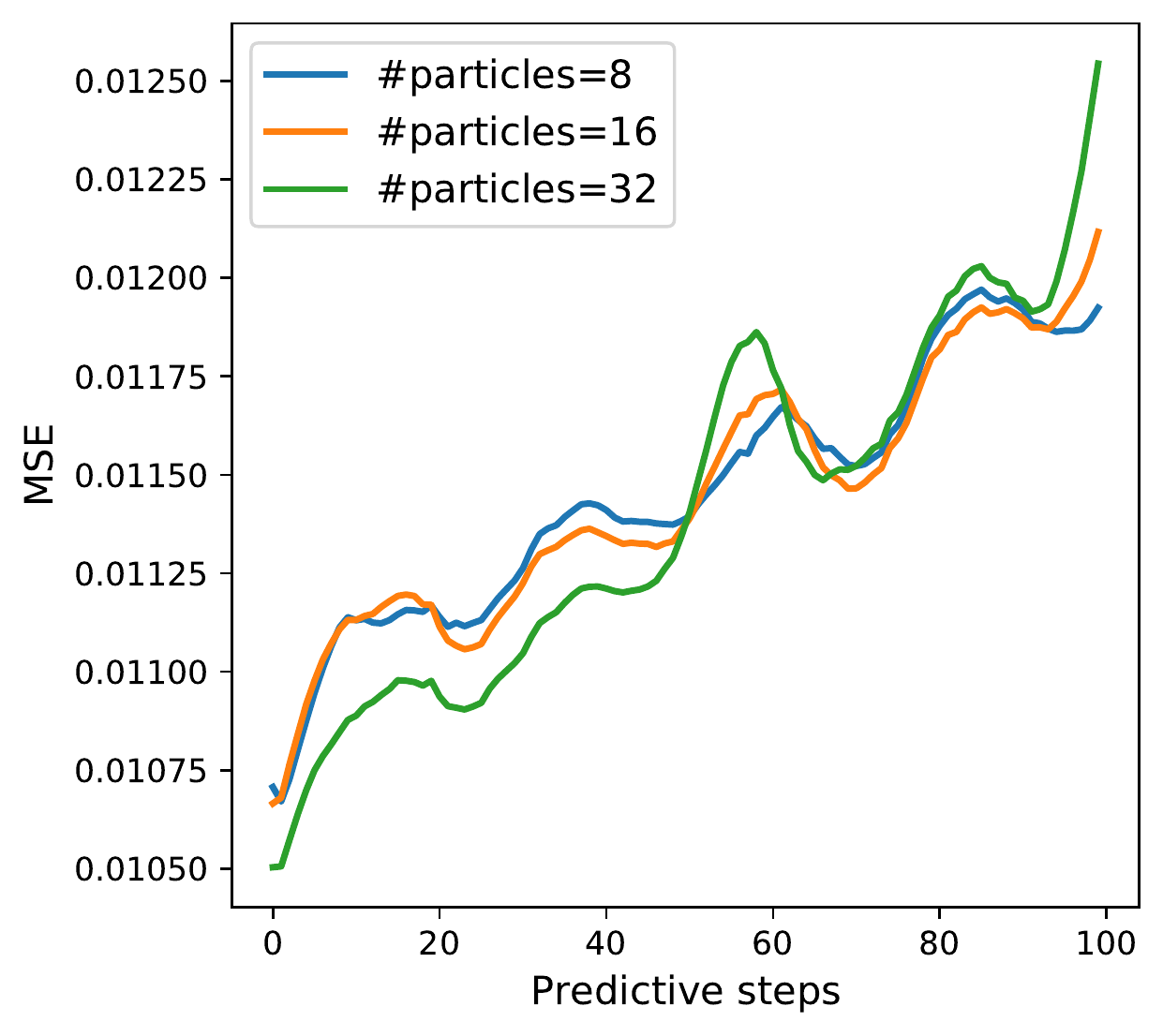}{MSE for synthetic Fitz-Hugh Nagumo data by EnKO without inflation methods (left) and with RTPP (center) and RTPS (right) for various ensemble sizes.}{fig:fhn_np}{0.32}{0.32}{0.32}
\figimagethree{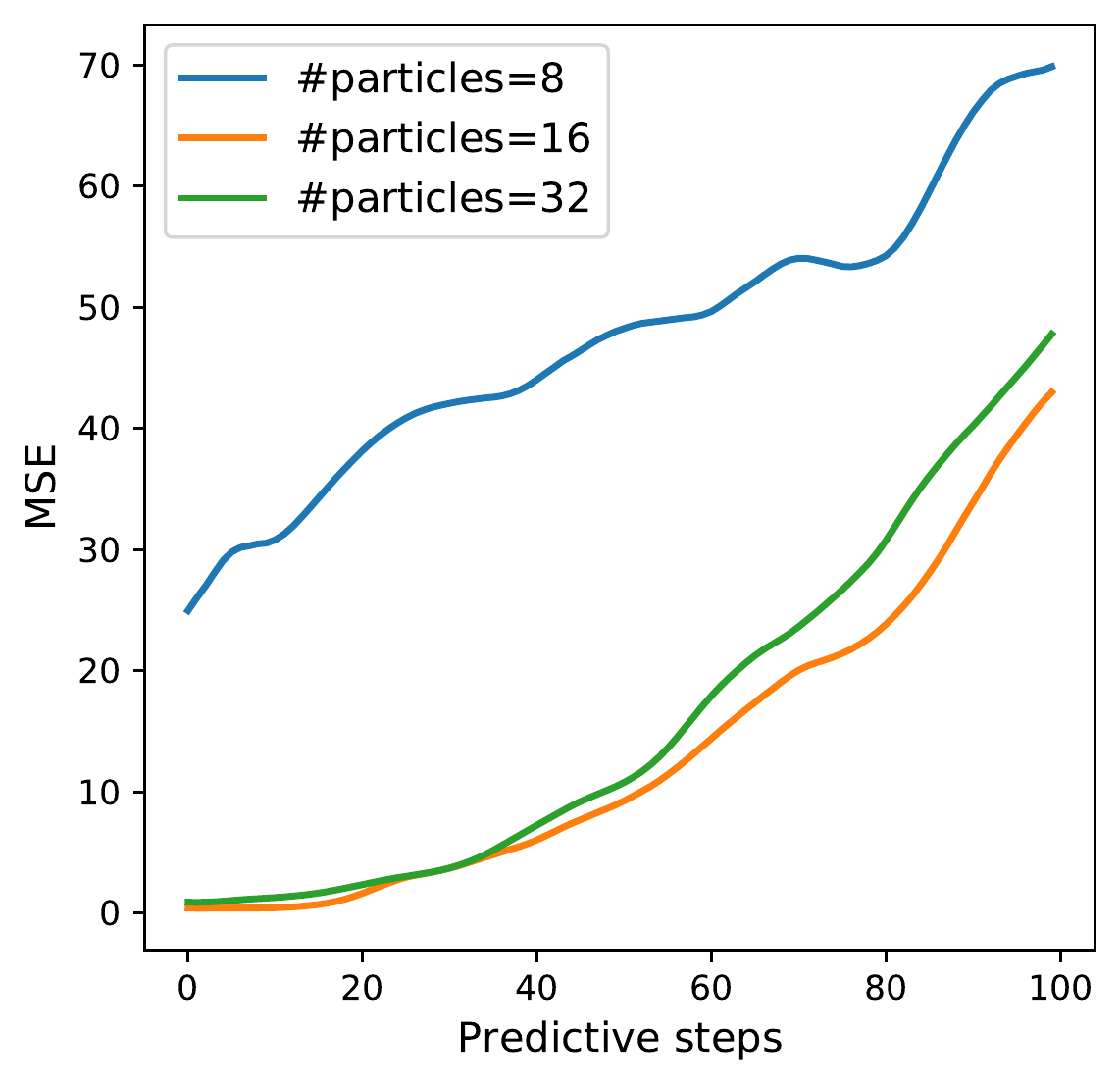}{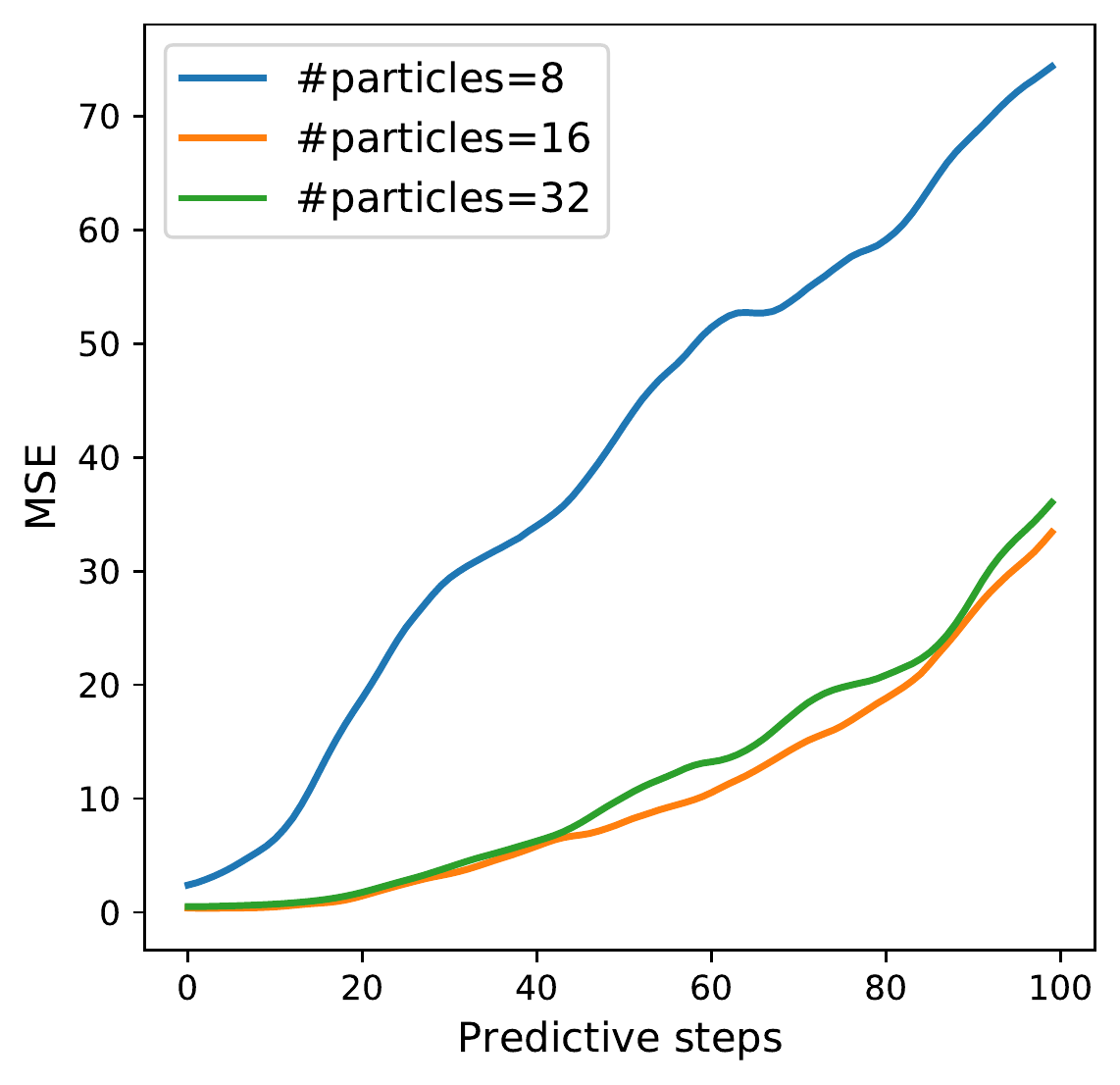}{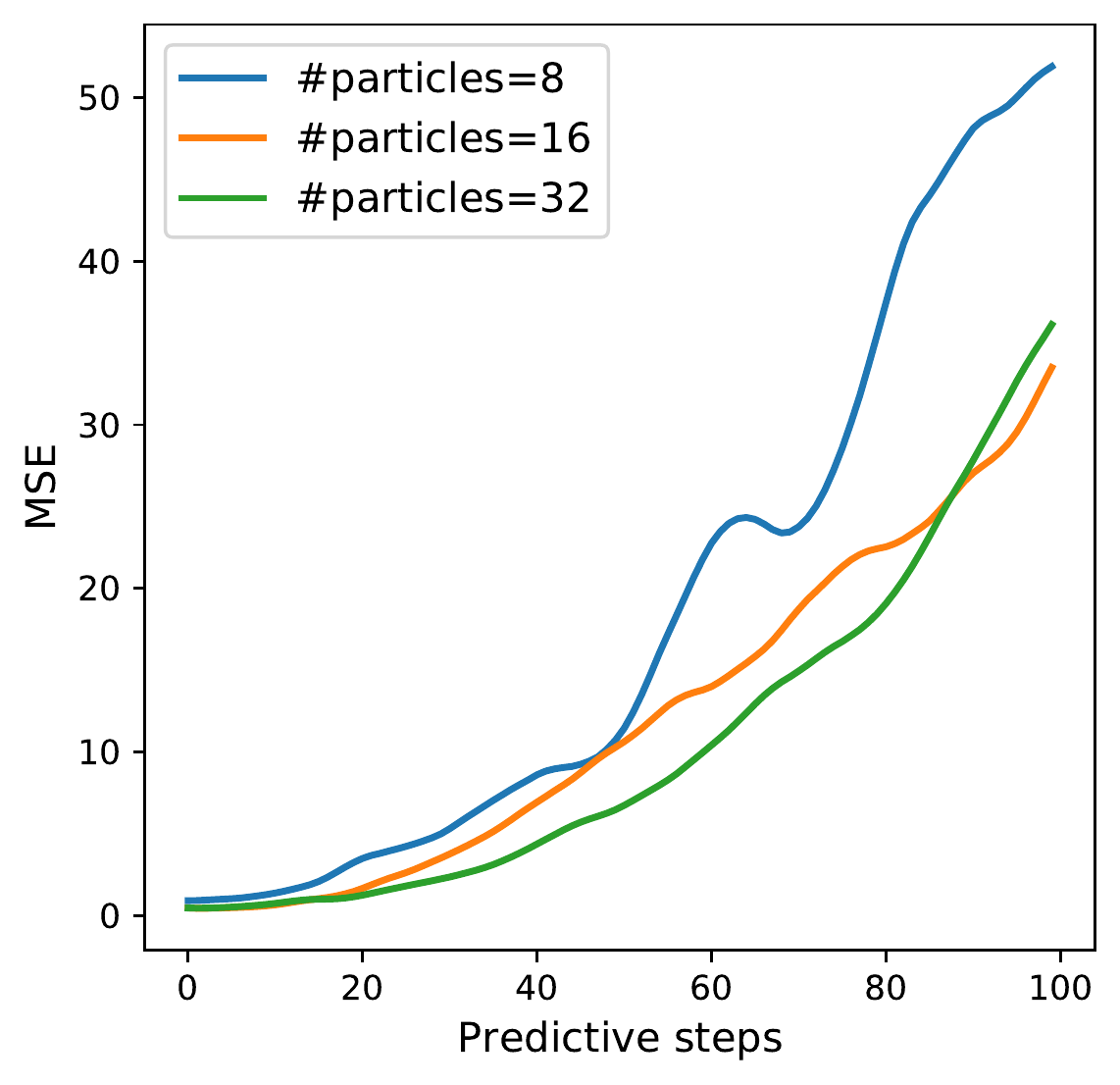}{MSE for synthetic Lorenz data by EnKO without inflation methods (left) and with RTPP (center) and RTPS (right) for various ensemble sizes.}{fig:lorenz_np}{0.32}{0.32}{0.32}
\begin{table}[t]
\caption{Chosen inflation factors}
\label{tab:factor}
\vskip 0.05in
\begin{center}
\begin{small}
\begin{tabular}{p{0.25\textwidth}*{4}{|p{0.1\textwidth}}}
\toprule
\multirow{2}{*}{Parameters} & \multicolumn{4}{c}{Dataset} \\
& FHN & Lorenz & Walking & RMNIST \\
\midrule
RTPP & 0.2 & 0.1 & 0.2 & 0.1 \\
RTPS & 0.1 & 0.1 & 0.2 & 0.3 \\
\bottomrule
\end{tabular}
\end{small}
\end{center}
\vskip -0.1in
\end{table}
\figimagetwo{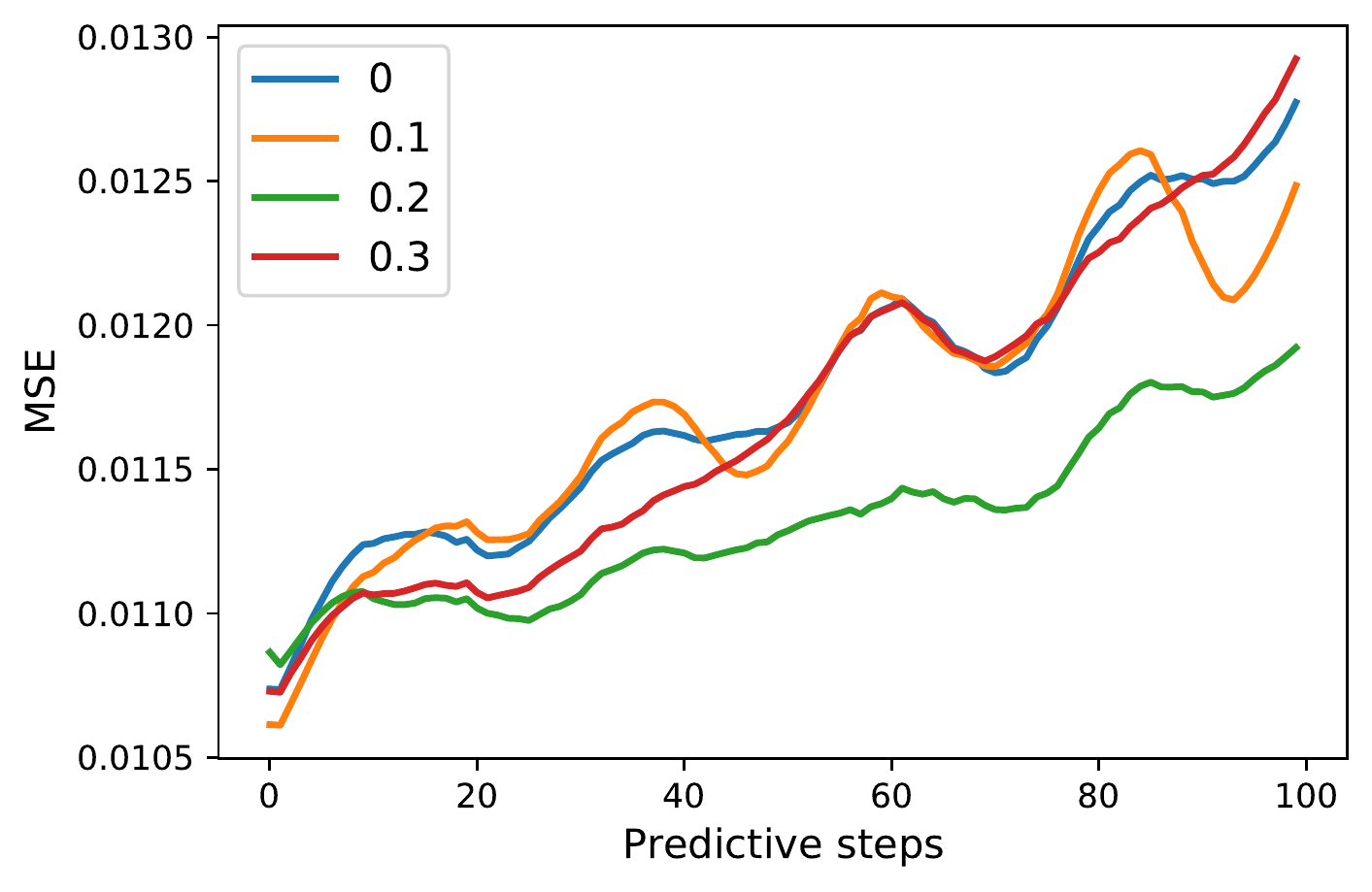}{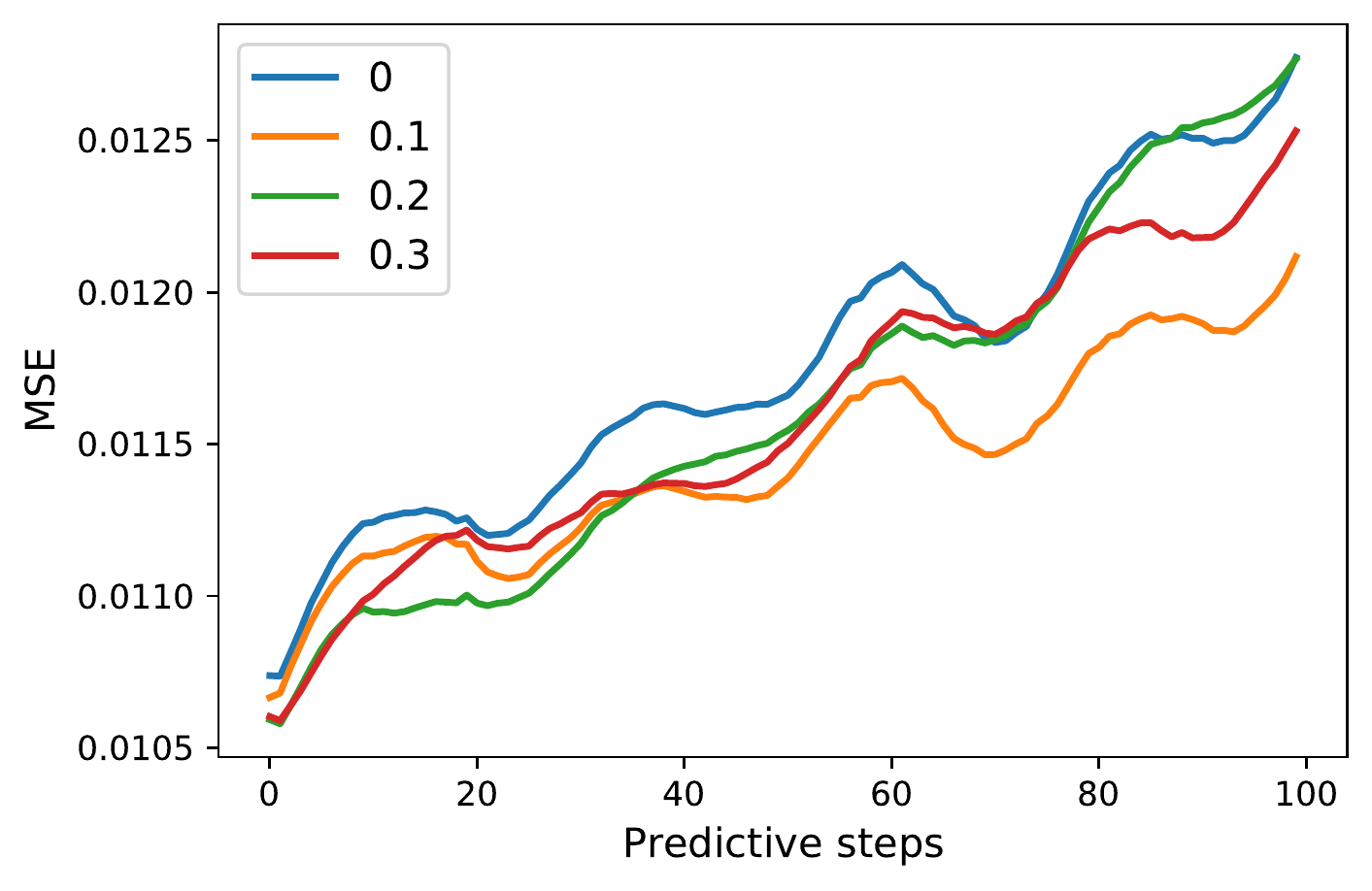}{MSE for synthetic Fitz-Hugh Nagumo data by EnKO with RTPP (left) and RTPS (right) for various inflation factors.}{fig:fhn_if}{0.42}{0.42}
\figimagetwo{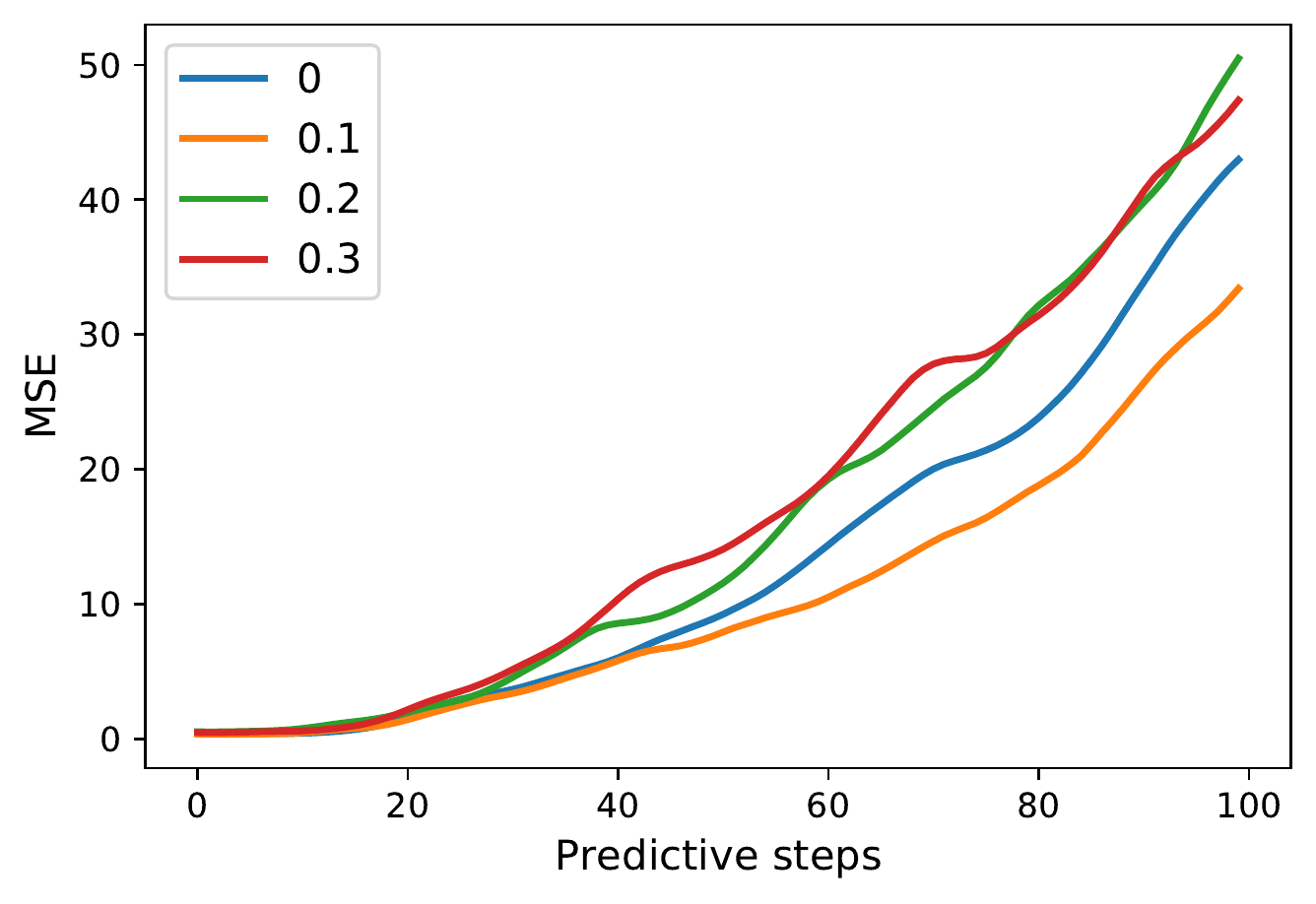}{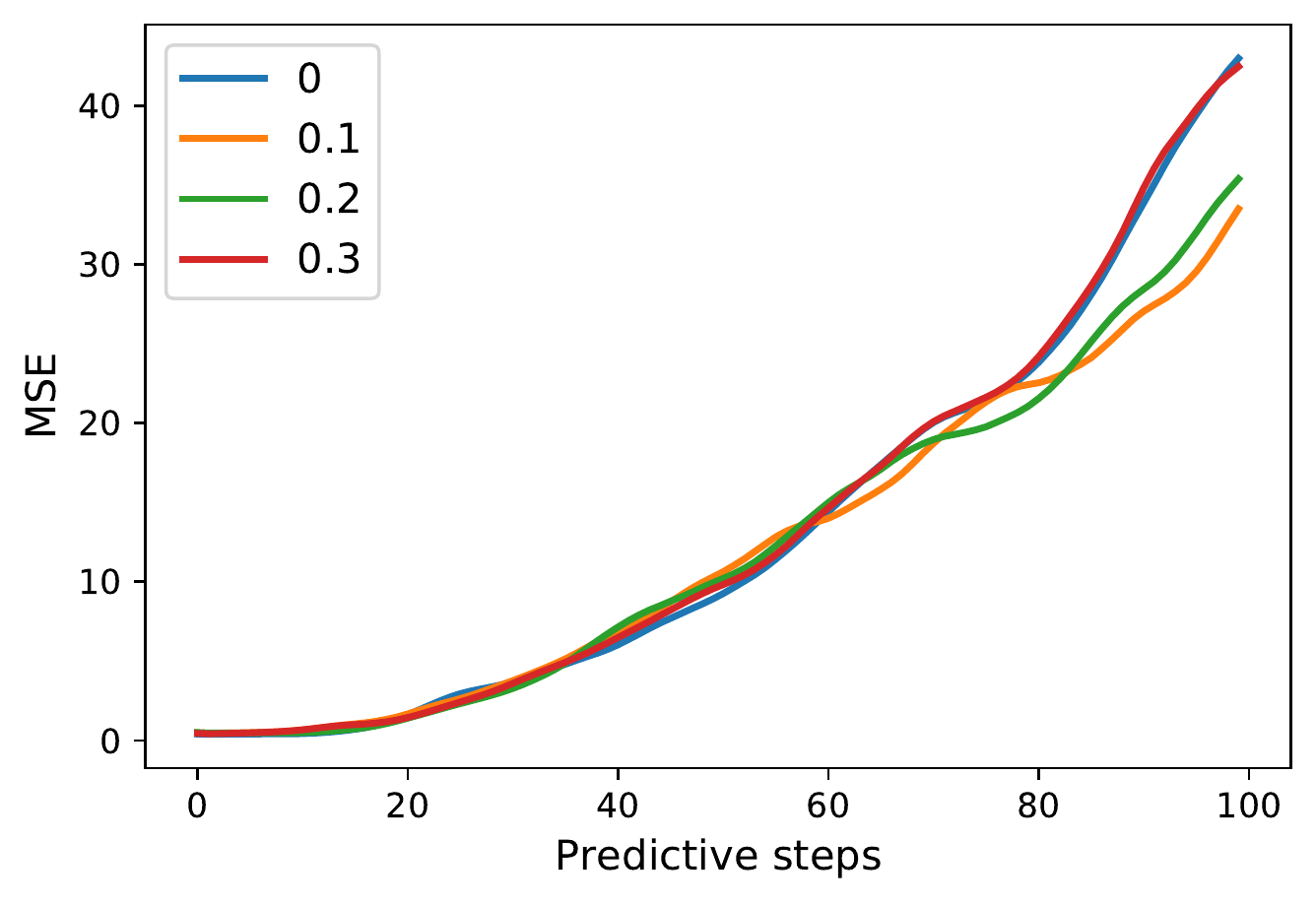}{MSE for synthetic Lorenz data by EnKO with RTPP (left) and RTPS (right) for various inflation factors.}{fig:lorenz_if}{0.42}{0.42}
\figimagetwo{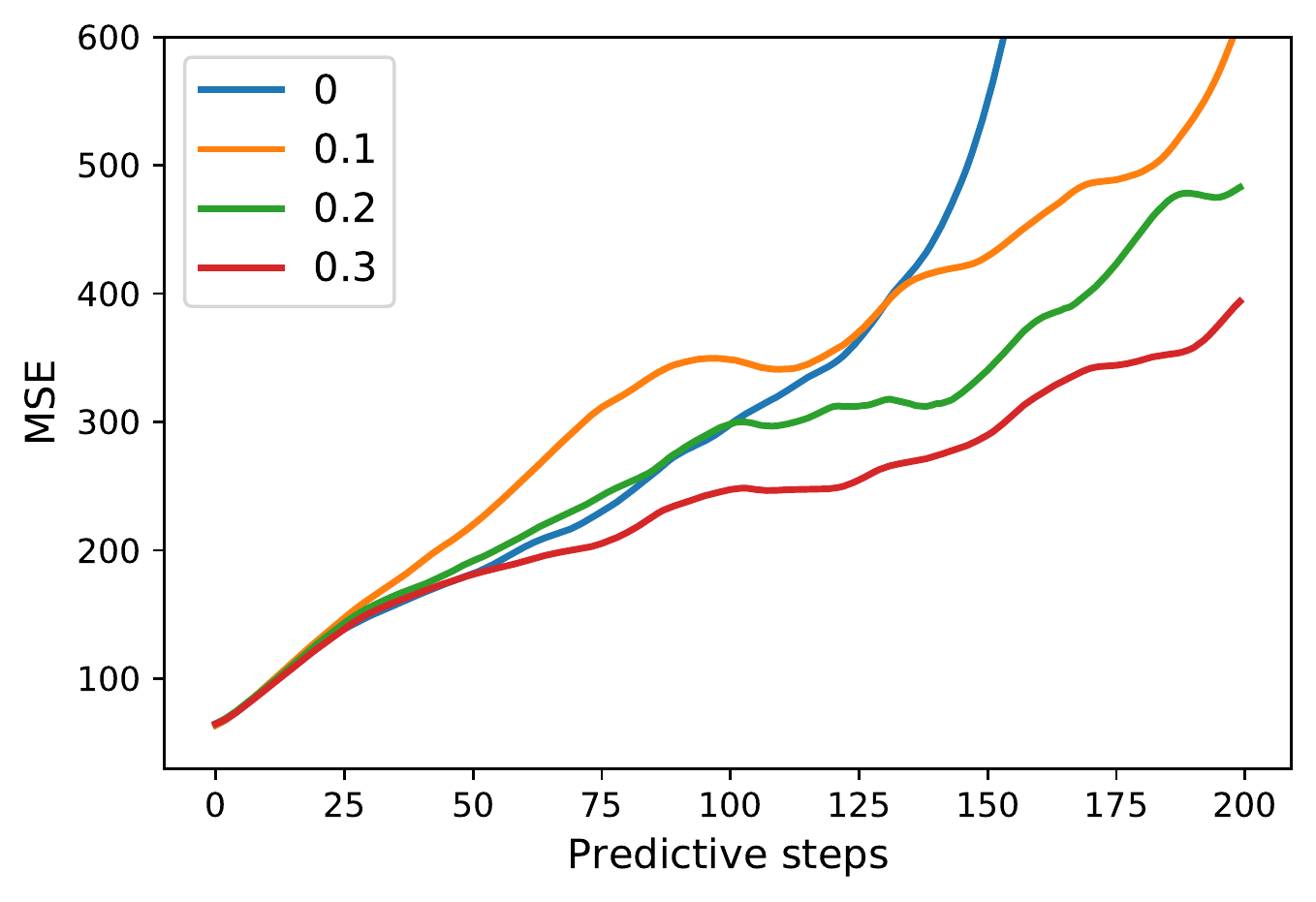}{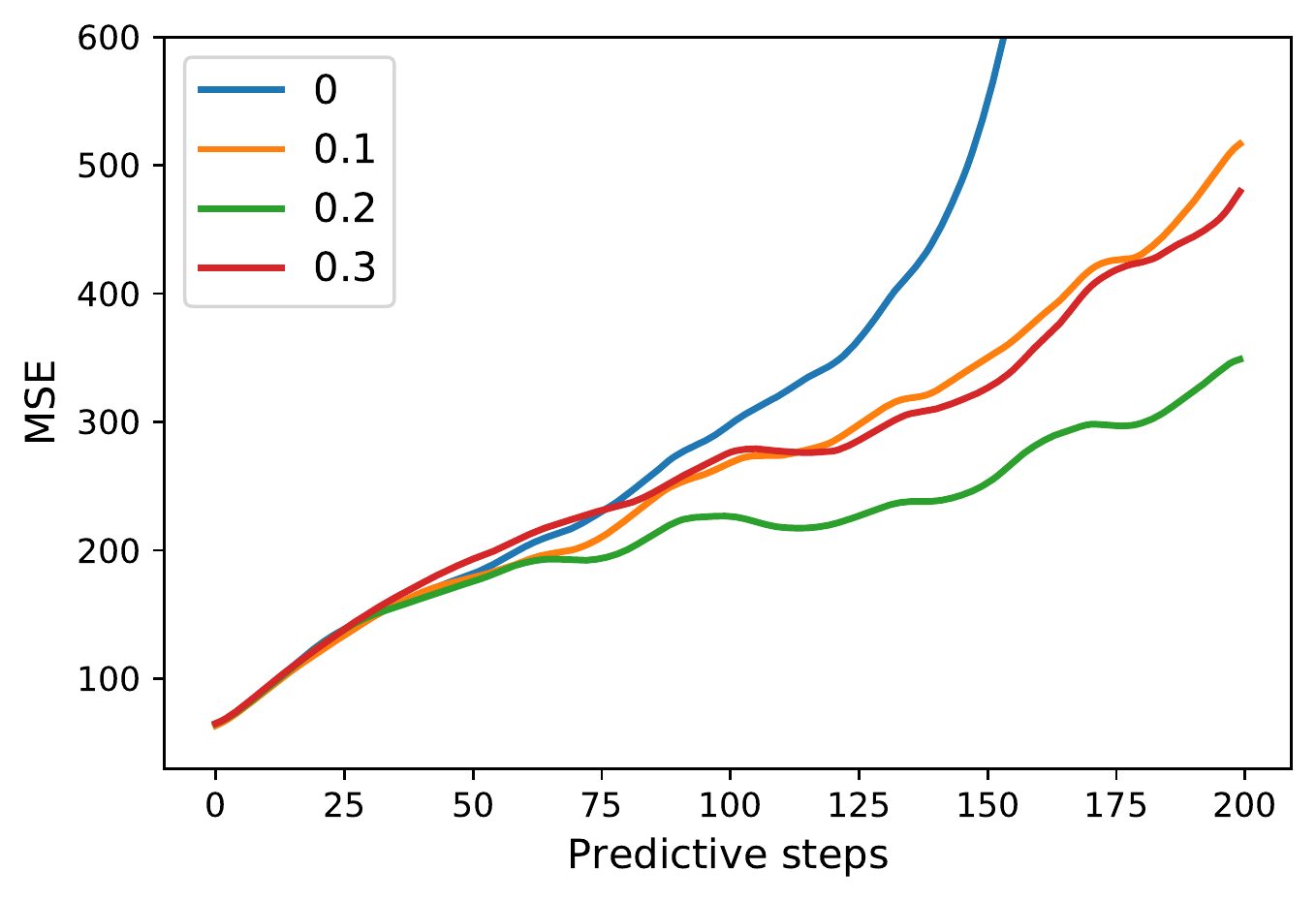}{MSE for the walking dataset by EnKO with RTPP (left) and RTPS (right) for various inflation factors.}{fig:mocap_if}{0.42}{0.42}
\figimagetwo{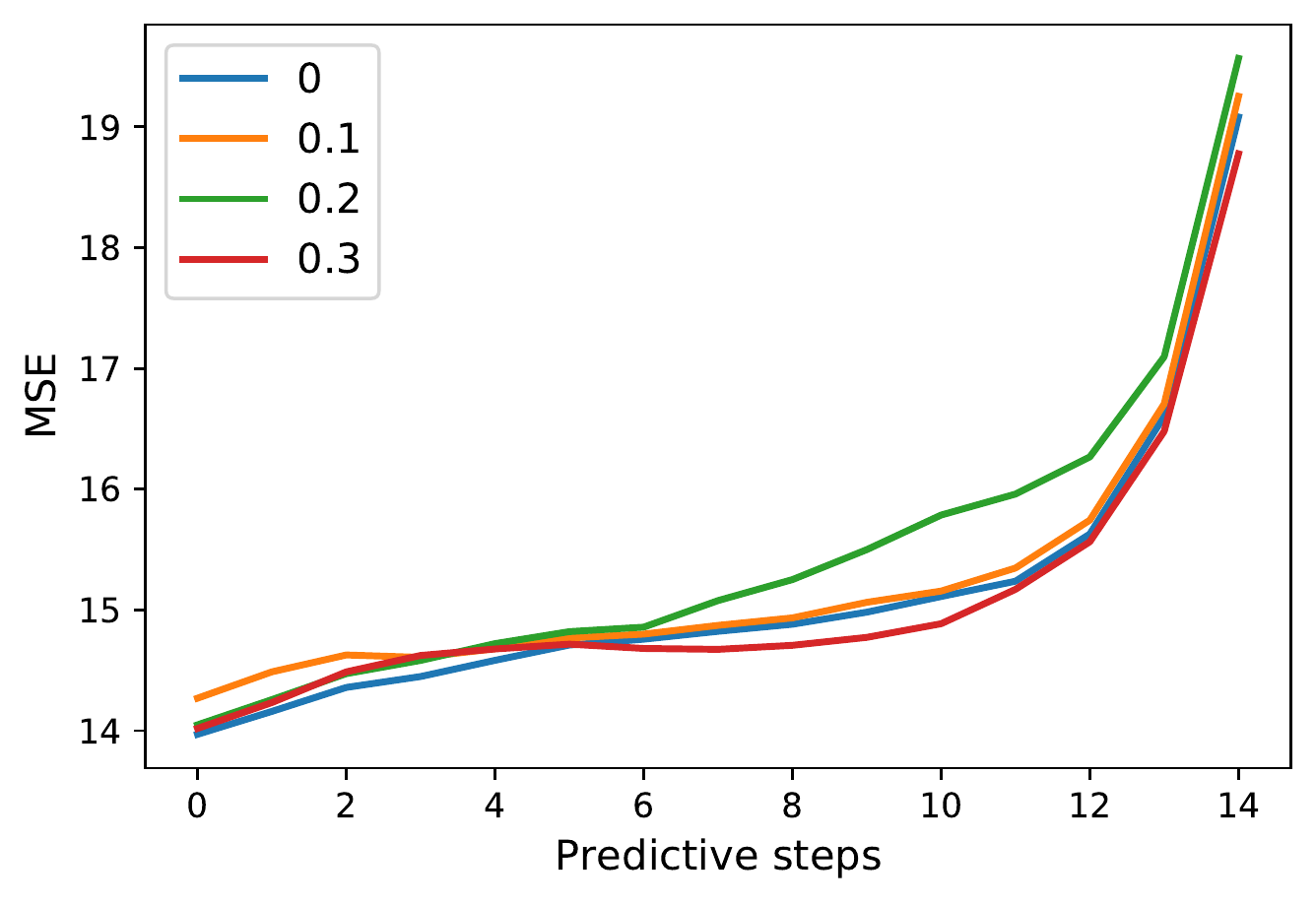}{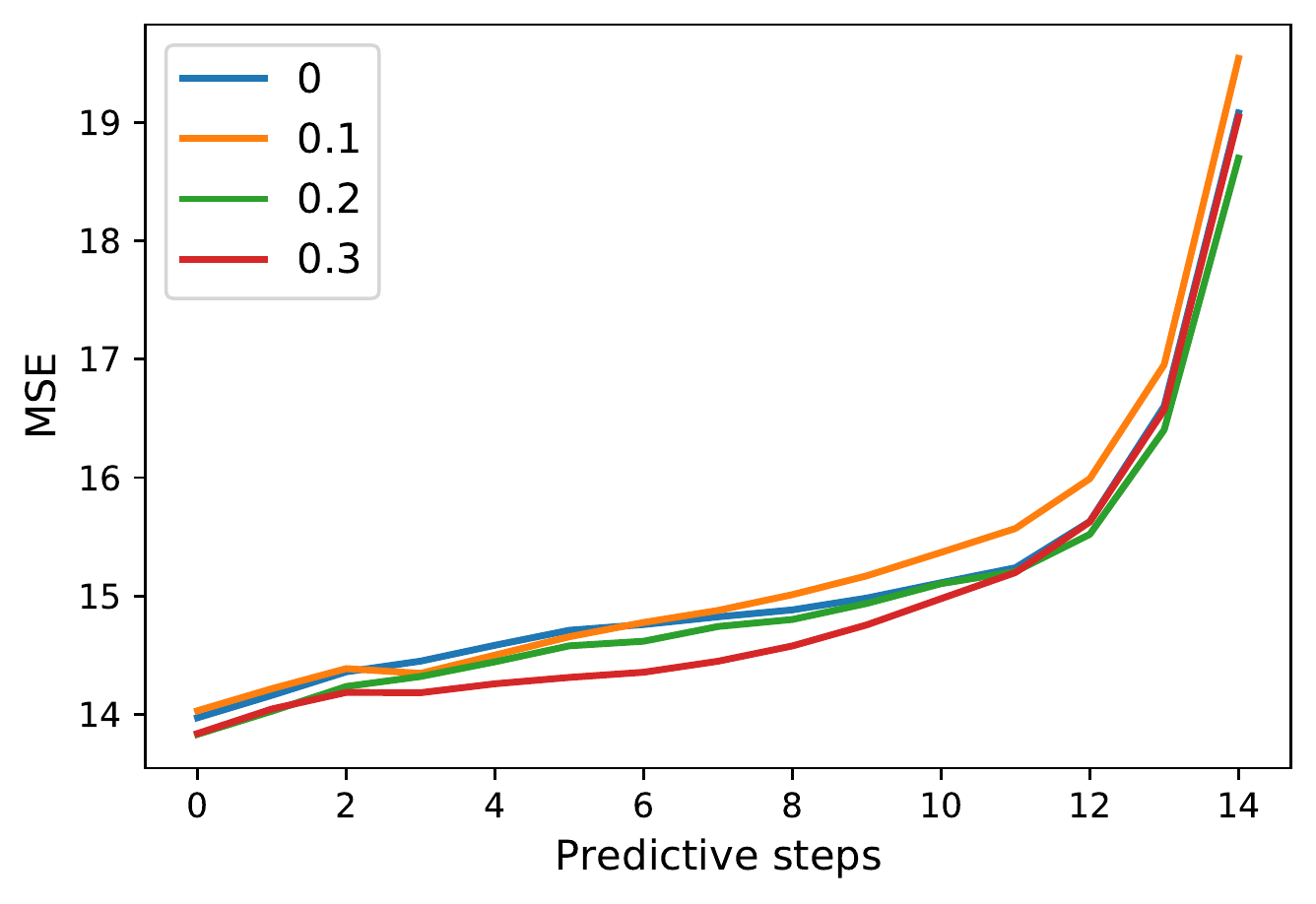}{MSE for the rotating MNIST dataset by EnKO with RTPP (left) and RTPS (right) for various inflation factors.}{fig:rmnist_if}{0.42}{0.42}
\figimage{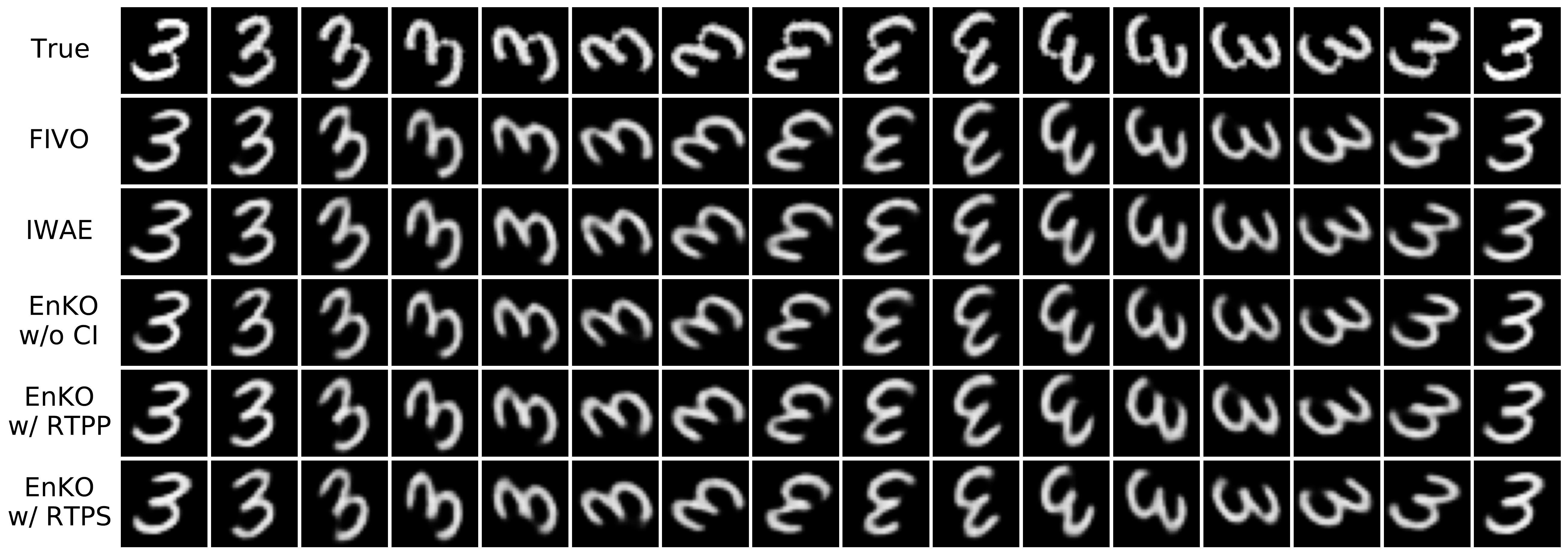}{True images and prediction results for rotating MNIST dataset.}{fig:rmnist_plot}{0.95}
\figimagep{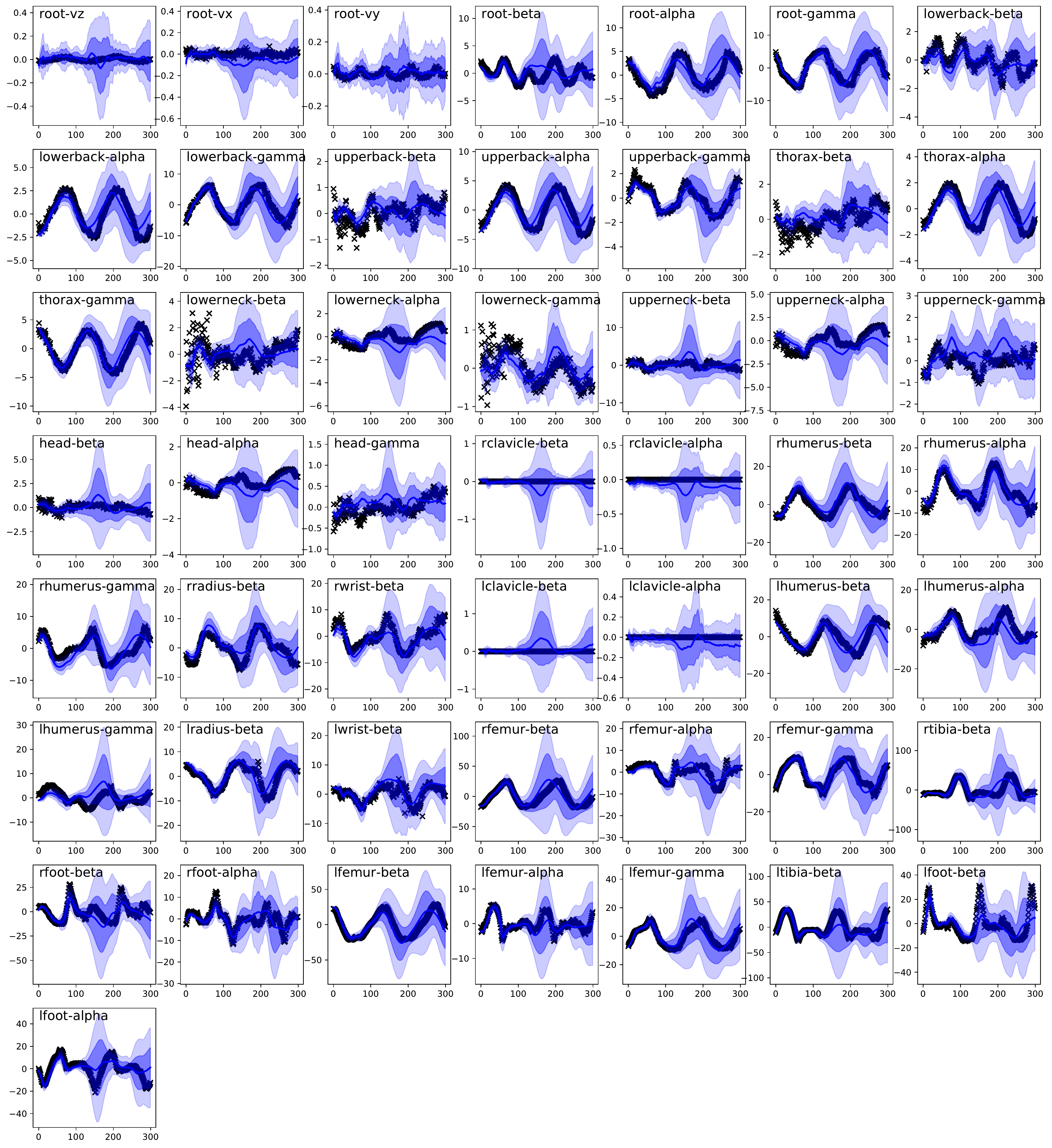}{Long prediction results for EnKO with RTPS. We inferred the initial latent state and predicted the values of the observations at all remaining time points according to the learned generative model. The black times represent the observed points, the solid blue line represents the predicted mean, and the dark and light blue widths represent the predicted mean plus or minus standard deviation and two standard deviations, respectively. The text in the figure shows the variable names. The vz, vx, and vy correspond to velocities, alpha, beta, and gamma correspond to Euler angles, and l and r correspond to left and right.}{fig:mocap_init}{0.95}

\subsection{Ensemble Size}
To verify that the number of particles changes the prediction ability, we performed experiments with varying numbers of particles.
Figure \ref{fig:fhn_np} and \ref{fig:lorenz_np} show predicted MSE for Fitz-Hugh Nagumo and Lorenz data, respectively, for various ensemble sizes.
For the Lorenz data, an increase in the number of particles improves the prediction accuracy.
In EnKO without inflation methods for FHN data, an increase in the number of particles hurts the prediction ability.
This result is contrary to the intuition that an increase in the number of particles results in better prediction accuracy; however, it is known that an increase in the number of particles does not necessarily have a positive effect on the results \citep{rai18}.
The Lorenz model is complex and requires a certain number of particles to learn well, while the FHN model is relatively simple and requires only eight particles.
It is a future task to calculate the intrinsic dimension of the data in advance and create a framework to appropriately determine the number of particles according to the complexity of the data.

\subsection{Inflation Factor}
To evaluate the effect of inflation factor on prediction and to enable appropriate factor selection, we conducted experiments with several factors \{0.1, 0.2, 0.3\}.
In Section \ref{sec:exp}, the factors that minimize the predicted MSE for the validation data are selected, and the results for the test data are shown, and the selected factors are summarized in Table \ref{tab:factor}.
Figure \ref{fig:fhn_if}-\ref{fig:rmnist_if} show predicted MSE for the Fitz-Hugh Nagumo, the Lorenz, the walking, and the rotating MNIST datasets, respectively, by EnKO with inflation methods for various inflation factors.
For the first three datasets, an inflation factor of 0.1 or 0.2 improves the prediction ability.
For the FHN and Lorenz data, since the errors when the inflation factor is 0.3 are higher, the appropriate factor to improve the inference accuracy is selected from 0.1 to 0.2.
For the walking data, a factor of 0.3 further improves the prediction accuracy of the RTPP, indicating that a relatively large factor is appropriate for inference.
This is because a small ratio of the number of particles to the observed dimension tends to underestimate the state covariance, as mentioned in Section 4.
For the rotating MNIST data, the inflation methods did not improve the prediction accuracy, and a factor between 0 and 0.1 is considered appropriate.
This is because the intrinsic dynamics of the dataset is a two-dimensional rotational motion, and if the outer VAEs are properly trained, the state covariance is rarely underestimated.

In light of the above, it is desirable to select the factor according to the observed dimension, the number of particles selected, and the complexity of a dataset.
Specifically, we propose that the factor should be selected from 0.01 to 0.2 for low-dimensional systems of 10 or less, 0.2 to 0.3 for medium-dimensional systems of 10 to 100, and according to the dimension of the space of auxiliary variables obtained using outer VAE for high-dimensional systems of 100 or more.

\subsection{CMU Walking Data}
Figure \ref{fig:mocap_init} shows long prediction results for EnKO with RTPS.
We inferred the initial latent state and predicted the values of the observations at all remaining time points according to the learned generative model.
Overall, the observation points are within the prediction interval of the mean plus or minus standard deviation, and the proposed method provides excellent predictions over a long period.
In particular, it is surprising that the proposed method predicts well even for variables with significant noise effects such as neck and thorax.
On the other hand, the proposed method overestimates the prediction interval for clavicles, which take almost zero values.

\subsection{Rotating MNIST Dataset}
Figure \ref{fig:rmnist_plot} shows true images and prediction results for FIVO, IWAE, and EnKO.
We inferred the initial latent state and predicted the values of the observations at all remaining time points according to the learned generative model.
It can be seen qualitatively that all the methods produce images close to the observations.
However, when we look at the predicted MSE (Figure \ref{fig:rmnist_mse}), EnKO outperforms IWAE and FIVO, indicating that EnKO can learn a more appropriate model at a level that is difficult to capture qualitatively.

\clearpage
\bibliography{enko}

\begin{thebibliography}{53}
\providecommand{\natexlab}[1]{#1}
\providecommand{\url}[1]{\texttt{#1}}
\expandafter\ifx\csname urlstyle\endcsname\relax
  \providecommand{\doi}[1]{doi: #1}\else
  \providecommand{\doi}{doi: \begingroup \urlstyle{rm}\Url}\fi

\bibitem[Anderson and Anderson(1999)]{and99}
Jeffrey~L. Anderson and Stephen~L. Anderson.
\newblock A {M}onte {C}arlo implementation of the nonlinear filtering problem
  to produce ensemble assimilations and forecasts.
\newblock \emph{Monthly Weather Review}, 127\penalty0 (12):\penalty0
  2741--2758, 1999.

\bibitem[Bayer and Osendorfer(2015)]{bay15}
Justin Bayer and Christian Osendorfer.
\newblock Learning stochastic recurrent networks.
\newblock \emph{arXiv preprint arXiv:1411.7610}, 2015.

\bibitem[Beal(2003)]{bea03}
Matthew~J. Beal.
\newblock Variational algorithms for approximate {B}ayesian inference.
\newblock Technical report, University of London, 2003.

\bibitem[Burda et~al.(2015)Burda, Grosse, and Salakhutdinov]{bur15}
Yuri Burda, Roger Grosse, and Ruslan Salakhutdinov.
\newblock Importance weighted autoencoders.
\newblock \emph{arXiv preprint arXiv:1509.00519}, 2015.

\bibitem[Casale et~al.(2018)Casale, Dalca, Saglietti, Listgarten, and
  Fusi]{cas18}
Francesco~Paolo Casale, Adrian Dalca, Luca Saglietti, Jennifer Listgarten, and
  Nicolo Fusi.
\newblock Gaussian process prior variational autoencoders.
\newblock In \emph{Neural Information Processing Systems}, pages 10369--10380,
  2018.

\bibitem[Chen et~al.(2019{\natexlab{a}})Chen, Lu, Wang, Trigoni, and
  Markham]{che19d}
Changhao Chen, Chris~X. Lu, Bing Wang, Niki Trigoni, and Andrew Markham.
\newblock {DynaNet}: Neural {K}alman dynamical model for motion estimation and
  prediction.
\newblock \emph{arXiv preprint arXiv:1908.03918}, 2019{\natexlab{a}}.

\bibitem[Chen et~al.(2018{\natexlab{a}})Chen, Lin, and Terejanu]{che18b}
Chao Chen, Xiao Lin, and Gabriel Terejanu.
\newblock An approximate {B}ayesian long short-term memory algorithm for
  outlier detection.
\newblock In \emph{International Conference on Pattern Recognition},
  2018{\natexlab{a}}.

\bibitem[Chen et~al.(2019{\natexlab{b}})Chen, Lin, Huang, and Terejanu]{che19b}
Chao Chen, Xiao Lin, Yuan Huang, and Gabriel Terejanu.
\newblock Approximate {B}ayesian neural network trained with ensemble {K}alman
  filter.
\newblock In \emph{International Joint Conference on Neural Networks},
  2019{\natexlab{b}}.

\bibitem[Chen et~al.(2018{\natexlab{b}})Chen, Rubanova, Bettencourt, and
  Duvenaud]{che18}
Ricky T.~Q. Chen, Yulia Rubanova, Jesse Bettencourt, and David~K. Duvenaud.
\newblock Neural ordinary differential equations.
\newblock In \emph{Advances in Neural Information Processing Systems},
  2018{\natexlab{b}}.

\bibitem[Cho et~al.(2014)Cho, van Merrienboer, Gulcehre, Bahdanau, Bougares,
  Schwenk, and Bengio]{cho14}
Kyunghyun Cho, Bart van Merrienboer, Caglar Gulcehre, Dzmitry Bahdanau, Fethi
  Bougares, Holger Schwenk, and Yoshua Bengio.
\newblock Learning phrase representations using {RNN} encoder-decoder for
  statistical machine translation.
\newblock In \emph{Conference on Empirical Methods in Natural Language Proc},
  2014.

\bibitem[Chung et~al.(2015)Chung, Kastner, Dinh, Goel, Courville, and
  Bengio]{chu15}
Junyoung Chung, Kyle Kastner, Laurent Dinh, Kratarth Goel, Aaron Courville, and
  Yoshua Bengio.
\newblock A recurrent latent variable model for sequential data.
\newblock In \emph{Neural Information Processing Systems}, 2015.

\bibitem[Corazza et~al.(2003)Corazza, Kalnay, Patil1, Yang, Morss, Cai,
  Szunyogh, Hunt, and Yorke]{cor03}
M.~Corazza, E.~Kalnay, D.~J. Patil1, S.-C. Yang, R.~Morss, M.~Cai, I.~Szunyogh,
  B.~R. Hunt, and J.~A. Yorke.
\newblock Use of the breeding technique to estimate the structure of the
  analysis "errors of the day".
\newblock \emph{Nonlinear Processes in Geophysics}, 10:\penalty0 233--243,
  2003.

\bibitem[Domke and Sheldon(2018)]{dom18}
Justin Domke and Daniel Sheldon.
\newblock Importance weighting and variational inference.
\newblock In \emph{Neural Information Processing System}, 2018.

\bibitem[Duceo and Johansen(2011)]{duc11}
Arnaud Duceo and Adam~M. Johansen.
\newblock A tutorial on particle filtering and smoothing: Fifteen years later.
\newblock \emph{Oxford Handbook of Nonlinear Filtering}, 12\penalty0
  (654--704):\penalty0 3, 2011.

\bibitem[Evensen(1994)]{eve94}
Geir Evensen.
\newblock Sequential data assimilation with a nonlinear quasi-geostrophic model
  using {M}onte-{C}arlo methods to forecast error statistics.
\newblock \emph{Journal of Geophysical Research: Oceans}, 99\penalty0
  (C5):\penalty0 10143--10162, 1994.

\bibitem[Evensen(2003)]{eve03}
Geir Evensen.
\newblock The ensemble {K}alman filter: Theoretical formulation and practical
  implementation.
\newblock \emph{Ocean dynamics}, 53\penalty0 (4):\penalty0 343--367, 2003.

\bibitem[Fox et~al.(2018)Fox, Hoar, Anderson, Arellano, Smith, Litvak, MacBean,
  Schimel, and Moore]{fox18}
Andrew~M. Fox, Timothy~J. Hoar, Jeffrey~L. Anderson, Avelino~F. Arellano,
  William~K. Smith, Marcy~E. Litvak, Natasha MacBean, David~S. Schimel, and
  David J.~P. Moore.
\newblock Evaluation of a data assimilation system for land surface models
  using {CLM}4.5.
\newblock \emph{Journal of Advances in Modeling Earth Systems}, 10\penalty0
  (10):\penalty0 2471--2494, 2018.

\bibitem[Fraccaro et~al.(2016)Fraccaro, S{\o}nderby, Paquet, and
  Winther]{fra16}
Marco Fraccaro, S{\o}ren~K. S{\o}nderby, Ulrich Paquet, and Ole Winther.
\newblock Sequential neural models with stochastic layers.
\newblock In \emph{Neural Information Processing System}, 2016.

\bibitem[Fraccaro et~al.(2017)Fraccaro, Kamronn, Paquet, and Winther]{fra17}
Marco Fraccaro, Simon Kamronn, Ulrich Paquet, and Ole Winther.
\newblock A disentangled recognition and nonlinear dynamics model for
  unsupervised learning.
\newblock In \emph{Neural Information Processing System}, 2017.

\bibitem[Gan et~al.(2015)Gan, Li, Henao, Carlson, and Carin]{gan15}
Zhe Gan, Chunyuan Li, Ricardo Henao, David~E Carlson, and Lawrence Carin.
\newblock Deep temporal sigmoid belief networks for sequence modeling.
\newblock In \emph{Neural Information Processing Systems}, pages 2467--2475,
  2015.

\bibitem[Godsill et~al.(2004)Godsill, Doucet, and West]{god04}
Simon~J Godsill, Arnaud Doucet, and Mike West.
\newblock Monte {C}arlo smoothing for nonlinear time series.
\newblock \emph{Journal of the american statistical association}, 99\penalty0
  (465):\penalty0 156--168, 2004.

\bibitem[Goyal et~al.(2017)Goyal, Sordoni, C{\^o}t{\'e}, Ke, and Bengio]{goy17}
Anirudh Goyal, Alessandro Sordoni, Marc-Alexandre C{\^o}t{\'e}, Nan~Rosemary
  Ke, and Yoshua Bengio.
\newblock {Z-Forcing}: Training stochastic recurrent networks.
\newblock In \emph{Neural Information Processing System}, 2017.

\bibitem[Hochreiter and Schmidhuber(1997)]{hoc97}
Sepp Hochreiter and J{\"u}rgen Schmidhuber.
\newblock Long short-term memory.
\newblock \emph{Neural computation}, 9\penalty0 (8):\penalty0 1735--1780, 1997.

\bibitem[Holmes et~al.(2012)Holmes, Lumley, Berkooz, and Rowley]{hol12}
Philip Holmes, John~L. Lumley, Gahl Berkooz, and Clarence~W. Rowley.
\newblock \emph{Turbulence, Coherent Structures, Dynamical Systems and
  Symmetry}.
\newblock Cambridge University Press, 2nd edition, 2012.

\bibitem[Jordan et~al.(1999)Jordan, Ghahramani, Jaakkola, and Saul]{jor99}
Michael~I. Jordan, Zoubin Ghahramani, Tommi~S. Jaakkola, and Lawrence~K. Saul.
\newblock An introduction to variational methods for graphical models.
\newblock \emph{Machine learning}, 37\penalty0 (2):\penalty0 183--233, 1999.

\bibitem[Kingma and Ba(2014)]{kin14}
Diederik~P. Kingma and Jimmy Ba.
\newblock Adam: A method for stochastic optimization.
\newblock In \emph{International Conference on Learning Representations}, 2014.

\bibitem[Kingma and Welling(2014)]{kin14v}
Diederik~P Kingma and Max Welling.
\newblock Auto-encoding variational {B}ayes.
\newblock In \emph{Proceedings of the 2nd International Conference on Learning
  Representations (ICLR)}, 2014.

\bibitem[Krishnan et~al.(2016{\natexlab{a}})Krishnan, Shalit, and
  Sontag]{kri16}
Rahul~G. Krishnan, Uri Shalit, and David Sontag.
\newblock Deep {K}alman filters.
\newblock \emph{arXiv preprint arXiv:1511.05121}, 2016{\natexlab{a}}.

\bibitem[Krishnan et~al.(2016{\natexlab{b}})Krishnan, Shalit, and
  Sontag]{kri16s}
Rahul~G Krishnan, Uri Shalit, and David Sontag.
\newblock Structured inference networks for nonlinear state space models.
\newblock \emph{arXiv preprint arXiv:1609.09869}, 2016{\natexlab{b}}.

\bibitem[Kutz et~al.(2016)Kutz, Brunton, Brunton, and Proctor]{kut16}
J.~Nathan Kutz, Steven~L. Brunton, Bingni~W. Brunton, and Joshua~L. Proctor.
\newblock \emph{Dynamic Mode Decomposition: Data-Driven Modeling of Complex
  Systems}.
\newblock Society for Industrial and Applied Mathematics, 2016.

\bibitem[Lawson et~al.(2018)Lawson, Tucker, Naesseth, Maddison, Adams, and
  Teh]{law18}
Dieterich Lawson, George Tucker, Christian~A Naesseth, Chris Maddison, Ryan~P
  Adams, and Yee~Whye Teh.
\newblock Twisted variational sequential {M}onte {C}arlo.
\newblock In \emph{Third workshop on Bayesian Deep Learning, NeurIPS}, 2018.

\bibitem[Le et~al.(2018)Le, Igl, Rainforth, Jin, and Wood]{le18}
Tuan~A. Le, Maximilian Igl, Tom Rainforth, Tom Jin, and Frank Wood.
\newblock Auto-encoding sequential {M}onte {C}arlo.
\newblock In \emph{International Conference on Learning Representations}, 2018.

\bibitem[Lindsten et~al.(2018)Lindsten, Helske, and Vihola]{lin18}
Fredrik Lindsten, Jouni Helske, and Matti Vihola.
\newblock Graphical model inference: Sequential {M}onte {C}arlo meets
  deterministic approximations.
\newblock In \emph{Neural Information Processing System}, volume~31, 2018.

\bibitem[Loh et~al.(2018)Loh, Omrani, and van~der Linden]{loh18}
Kelvin Loh, Pejman~Shoeibi Omrani, and Ruud van~der Linden.
\newblock Deep learning and data assimilation for real-time production
  prediction in natural gas wells.
\newblock \emph{arXiv preprint arXiv:1802.05141}, 2018.

\bibitem[Maddison et~al.(2017)Maddison, Lawson, Tucker, Heess, Norouzi, Mnih,
  Doucet, and Teh]{mad17}
Chris~J Maddison, John Lawson, George Tucker, Nicolas Heess, Mohammad Norouzi,
  Andriy Mnih, Arnaud Doucet, and Yee Teh.
\newblock Filtering variational objectives.
\newblock In \emph{Neural Information Processing System}, 2017.

\bibitem[Masrani et~al.(2019)Masrani, Le, and Wood]{mas19}
Vaden Masrani, Tuan~Anh Le, and Frank Wood.
\newblock The thermodynamic variational objective.
\newblock In \emph{Neural Information Processing Systems}, 2019.

\bibitem[Mitchell and Houtekamer(2000)]{mit00}
Herschel~L. Mitchell and P.~L. Houtekamer.
\newblock An adaptive ensemble {K}alman filter.
\newblock \emph{Monthly Weather Review}, 128\penalty0 (2):\penalty0 416--433,
  2000.

\bibitem[Molchanov et~al.(2019)Molchanov, Kharitonov, Sobolev, and
  Vetrov]{mol19}
Dmitry Molchanov, Valery Kharitonov, Artem Sobolev, and Dmitry Vetrov.
\newblock Doubly semi-implicit variational inference.
\newblock In \emph{International Conference on Artificial Intelligence and
  Statistics}, 2019.

\bibitem[Moretti et~al.(2019{\natexlab{a}})Moretti, Wang, Wu, and
  Pe'er]{mor19s}
Antonio Moretti, Zizhao Wang, Luhuan Wu, and Itsik Pe'er.
\newblock Smoothing nonlinear variational objectives with sequential {M}onte
  {C}arlo.
\newblock In \emph{International Conference on Learning Representations},
  2019{\natexlab{a}}.

\bibitem[Moretti et~al.(2019{\natexlab{b}})Moretti, Wang, Wu, Drori, and
  Pe'er]{mor19p2}
Antonio~Khalil Moretti, Zizhao Wang, Luhuan Wu, Iddo Drori, and Itsik Pe'er.
\newblock Particle smoothing variational objectives.
\newblock \emph{arXiv preprint arXiv:1909.09734}, 2019{\natexlab{b}}.

\bibitem[Moretti et~al.(2020)Moretti, Wang, Wu, Drori, and Pe'er]{mor19p}
Antonio~Khalil Moretti, Zizhao Wang, Luhuan Wu, Iddo Drori, and Itsik Pe'er.
\newblock Variational objectives for {M}arkovian dynamics with backwards
  simulation.
\newblock In \emph{European Conference on Artificial Intelligence}, 2020.

\bibitem[Naesseth et~al.(2018)Naesseth, Linderman, Ranganath, and Blei]{nae18}
Christian~A Naesseth, Scott~W Linderman, Rajesh Ranganath, and David~M Blei.
\newblock Variational sequential {M}onte {C}arlo.
\newblock In \emph{International Conference on Artificial Intelligence and
  Statistics}, 2018.

\bibitem[Rainforth et~al.(2018)Rainforth, Kosiorek, Le, Maddison, Igl, Wood,
  and Teh]{rai18}
Tom Rainforth, Adam~R Kosiorek, Tuan~Anh Le, Chris~J Maddison, Maximilian Igl,
  Frank Wood, and Yee~Whye Teh.
\newblock Tighter variational bounds are not necessarily better.
\newblock In \emph{International Conference on Machine Learning}, volume~80,
  pages 4274--4282, 2018.

\bibitem[Sakov et~al.(2012)Sakov, Counillon, Bertino, Lis{\ae}ter, Oke, and
  Korablev]{sak12}
Pavel Sakov, F~Counillon, L~Bertino, K~A Lis{\ae}ter, PR~Oke, and A~Korablev.
\newblock {TOPAZ4}: An ocean-sea ice data assimilation system for the north
  atlantic and arctic.
\newblock \emph{Ocean Science}, 8\penalty0 (4):\penalty0 633--656, 2012.

\bibitem[Shumway and Stoffer(2017)]{shu17}
Robert~H. Shumway and David~S. Stoffer.
\newblock \emph{Time Series Analysis and Its Applications With {R} Examples}.
\newblock Springer, 2017.

\bibitem[Titsias and Ruiz(2019)]{mic19}
Michalis~K. Titsias and Francisco J.~R. Ruiz.
\newblock Unbiased implicit variational inference.
\newblock In \emph{The 22nd International Conference on Artificial Intelligence
  and Statistics}, 2019.

\bibitem[Wang et~al.(2008)Wang, Fleet, and Hertzmann]{wan08}
Jack~M. Wang, David~J. Fleet, and Aaron Hertzmann.
\newblock Gaussian process dynamical models for human motion.
\newblock In \emph{IEEE Transactions on Pattern Analysis and Machine
  Intelligence}, volume~30, pages 283--298, 2008.

\bibitem[Watter et~al.(2015)Watter, Springenberg, Boedecker, and
  Riedmiller]{wat15}
Manuel Watter, Jost~Tobias Springenberg, Joschka Boedecker, and Martin
  Riedmiller.
\newblock Embed to control: A locally linear latent dynamics model for control
  from raw images.
\newblock In \emph{Neural Information Processing Systems}, 2015.

\bibitem[Whitaker and Hamill(2012)]{whi12}
Jeffrey~S. Whitaker and Thomas~M. Hamill.
\newblock Evaluating methods to account for system errors in ensemble data
  assimilation.
\newblock \emph{Monthly Weather Review}, 140\penalty0 (9):\penalty0 3078--3089,
  2012.

\bibitem[Yildiz et~al.(2019)Yildiz, Heinonen, and L{\"a}hdesm{\"a}ki]{yil19}
Cagatay Yildiz, Markus Heinonen, and Harri L{\"a}hdesm{\"a}ki.
\newblock {ODE2VAE}: Deep generative second order {ODEs} with {B}ayesian neural
  networks.
\newblock In \emph{Neural Information Processing Systems}, 2019.

\bibitem[Yin and Zhou(2018)]{yin18}
Mingzhang Yin and Mingyuan Zhou.
\newblock Semi-implicit variational inference.
\newblock In \emph{International Conference on Machine Learning}, 2018.

\bibitem[Zhang et~al.(2017)Zhang, Butepage, Kjellstrom, and Mandt]{zha17}
Cheng Zhang, Judith Butepage, Hedvig Kjellstrom, and Stephan Mandt.
\newblock Advances in variational inference.
\newblock 2017.

\bibitem[Zhang et~al.(2004)Zhang, Snyder, and Sun]{zha04}
F.~Zhang, Chris Snyder, and Juanzhen Sun.
\newblock Impacts of initial estimate and observation availability on
  convective-scale data assimilation with an ensemble {K}alman filter.
\newblock \emph{Monthly Weather Review}, 132\penalty0 (5):\penalty0 1238--1253,
  2004.

\end{thebibliography}
\bibliographystyle{plainnat}

\end{document}